\PassOptionsToPackage{hyphens}{url}
\documentclass[sigconf, nonacm]{acmart}

\renewcommand\footnotetextcopyrightpermission[1]{}
\usepackage{booktabs}
\usepackage{amsmath}

\usepackage{amssymb}
\usepackage{graphicx}
\usepackage{listings}
\usepackage{xcolor}
\usepackage{multirow}
\usepackage{makecell}
\usepackage{enumitem}
\usepackage{textcomp}
\usepackage{microtype}
\usepackage{tabularx}

\definecolor{codegray}{rgb}{0.95,0.95,0.95}
\definecolor{codeblue}{rgb}{0.2,0.2,0.6}
\definecolor{codegreen}{rgb}{0.0,0.5,0.0}
\lstdefinestyle{v}{
  backgroundcolor=\color{codegray},
  commentstyle=\color{codegreen}\itshape,
  keywordstyle=\color{codeblue}\bfseries,
  basicstyle=\ttfamily\footnotesize,
  breaklines=true, captionpos=b, keepspaces=true,
  numbers=none, showstringspaces=false, tabsize=2,
  frame=single, framerule=0.3pt
}
\usepackage{xspace}
\usepackage{hyperref}
\hypersetup{colorlinks=true, urlcolor=blue, linkcolor=blue, citecolor=blue}
\newcommand{\model}{\textsc{VectraYX-Vision-1B}\xspace}

\acmConference[Preprint]{Preprint}{2026}{}
\acmYear{2026}
\copyrightyear{2026}
\acmISBN{}
\acmDOI{}

\begin{document}

\title[VectraYX-Vision-1B]{VectraYX-Vision-1B: A Sub-2B Spanish/LATAM
Cybersecurity Vision--Language Model, and What Limits Its Visual
Grounding}

\author{Juan S. Santillana}
\affiliation{%
  \institution{Globant$^*$}
  \city{}
  \country{}}
\email{juan.salas@globant.com}
\thanks{$^*$The author is a DevOps engineer at Globant. This work was carried
out independently of that role; institutional affiliation approval is
pending, and the affiliation is listed here for disclosure. This is a
preprint and has not been peer reviewed.}

\begin{abstract}
We build \model, a sub-2B Spanish/LATAM cybersecurity vision--language model
for offline use, and measure what limits its visual grounding. A frozen
1.04B-parameter decoder is coupled to a frozen vision encoder through a
trainable projector: first SigLIP, then the Qwen2-VL-2B tower with a projector
shaped for \texttt{llama.cpp}'s \texttt{mmproj} export. With SigLIP, after five
fine-tuning defects were repaired, a nine-field extraction gate with a
shuffled-image control passes the same 2/9 fields under every configuration
that keeps the encoder and adds no text hint. A frozen-feature probe explains
why. Transplanting the Qwen2-VL tower with its own merger reads an 8-nibble
address SigLIP never read (0.00 to 0.81 exact match). On B8 (2,040 items, 34
fields, 16 templates, shuffled-image and best-constant controls), 9 fields pass
on a Qwen-tower checkpoint, all within trained template--field pairs.
Screenshot tool identification (B6) sat at exactly 0.000. That was not a
perception ceiling. Stock Qwen2-VL-2B transcribes the same images at word
recall 0.93 at our pixel budget (0.52 with our letterboxing). Training only the
projector on B6's task format and render family lifts B6 tool identification
to 0.96 and recall to 0.48--0.51. Perception of rendered text is therefore
present and trainable in this frozen-backbone design. The evidence is narrow.
B6 is nearly in-distribution for that projector, which also forgets part of B8
(mean field accuracy 0.53 to 0.29 on a retention check); runs are single-seed,
and no checkpoint yet has both. We document four harness defects, retract an
earlier B6 score, and release code, benchmarks and checkpoints.
\end{abstract}


\maketitle

\section{Introduction}
\label{sec:intro}

Security analysts work through images. A reverse engineer reads a disassembly
pane in IDA; a SOC analyst triages a Splunk search or a Wireshark capture; a
forensic examiner scans a Volatility process listing; a penetration tester
interprets Nmap output and Metasploit sessions. Yet the dominant tools for
machine assistance in these workflows are text-only large language models
(LLMs), which cannot see the screen the analyst is looking at, and general
vision--language models (VLMs) such as LLaVA~\cite{liu2023llava},
InternVL~\cite{chen2024internvl}, and Qwen-VL~\cite{bai2023qwenvl}, which are
neither specialized for security imagery nor deployable in the air-gapped,
data-sovereign environments where much security work happens. Two further gaps
compound the problem in Latin America: (i)~these models answer poorly in
Spanish on technical content, and (ii)~their 7B--70B footprints preclude local,
offline inference on the commodity hardware available to under-resourced
security teams.

This paper presents \model, a vision--language model that targets exactly this
niche: a 1.04B-parameter Spanish/LATAM cybersecurity decoder
(\textsc{VectraYX-1B}, the mid-tier of the VectraYX family~\cite{santillana2026vectrayx})
augmented with a frozen SigLIP-so400m~\cite{zhai2023siglip} encoder and an MLP
projector, trained to reason over security screenshots and invoke tools, and
exportable to \texttt{llama.cpp}~\cite{llamacpp} for on-premise use. A second
encoder pathway, introduced after the SigLIP results located the bottleneck in
the encoder, couples the same frozen decoder to the frozen visual tower of
Qwen2-VL-2B~\cite{wang2024qwen2vl} through a projector that takes the place of
Qwen's own merger and keeps its tensor layout, so that the pair is shaped for
export as an \texttt{mmproj}; in that pathway only the projector is ever
trained (\S\ref{sec:arch:qwenline}).

\subsection{Threat model and deployment setting}
\label{sec:intro:threat}
We assume a defensive or authorized-offensive analyst operating inside an
organization's security perimeter, frequently on \emph{air-gapped} or
egress-restricted hosts (malware-analysis sandboxes, classified forensics
workstations, incident-response ``war rooms''). Sending screenshots of
potentially sensitive artifacts---customer data in a packet capture, malware
strings, internal network topology from a scan---to a third-party cloud VLM is
often prohibited by policy or law. The model must therefore run entirely
locally. Two adversarial concerns follow directly. First, \emph{the input images
are attacker-controlled}: a malware sample rendered in a disassembler, or a
crafted screenshot, can carry adversarial visual content, so we treat visual
prompt injection~\cite{gong2025figstep} as in-scope for the safety discussion
(Section~\ref{sec:limitations}). Second, \emph{the model's tool-invocation
capability is a dual-use surface}: native \texttt{<|tool\_call|>} emission that
drives an MCP~\cite{mcp2024} server can execute real actions, so tool grounding
and guardrails are a first-class design concern rather than an afterthought.

\subsection{Contributions}
\label{sec:intro:contrib}
We make the following contributions. All training phases have run to completion;
the system, corpus, benchmarks, and ablation design are complete and released.
The vision result of the SigLIP pathway is negative, and we treat the
diagnosis---not the score---as the contribution: what was broken, what was
fixed, what did not move when it was, and why. Later, separately trained
checkpoints of the Qwen-tower pathway add two positive measurements to that
account without displacing any of it: grounding confirmed against two controls
on B8, scoped to the template--field combinations trained
(\S\ref{sec:results:b8}), and a screenshot benchmark that had sat at exactly
zero brought to 0.96 tool identification by projector-only training, scoped to
that benchmark's own render family and bought at a measured cost in retention
(\S\ref{sec:results:b6fix}).

\begin{enumerate}[leftmargin=1.4em]
  \item \textbf{A diagnosed negative result: an invariant grounding ceiling with
    a quantitative explanation.} We build a sub-2B cybersecurity VLM, repair
    five silent defects in its vision fine-tuning path---optimizer underflow in
    bf16 master weights (21\% of trainable tensors with a parameter delta of
    exactly zero after 300 steps), an incomplete adapter re-freeze, an
    order-of-operations defect between LoRA wrapping and checkpoint loading
    that left all 22 attention layers at random initialization under
    \texttt{strict=False} (a measured weight norm of 40.9668 against a
    theoretical 40.96), its mirror image on same-phase resumes, and a learning
    rate 20--60$\times$ too low for a rank-16 adapter---and find that grounding
    on a nine-field exact-match gate is 2/9 \emph{invariantly}: the same two
    fields pass in all four configurations that leave the encoder alone and
    add no out-of-band text, and the one that changes it loses a field rather
    than gaining one (\S\ref{sec:results:gate}). We explain the ceiling rather
    than attribute it to training volume. The field that is read almost
    perfectly is the highest-entropy one in the corpus (1{,}120 strings, 9.68
    bits), so resolution per se is not the binding constraint; what binds is
    information per glyph compounded by a metric exponential in glyph count,
    $p^n$. A frozen-feature probe measuring $p\approx0.61$ predicts $1.9\%$
    exact-match on an 8-nibble address against an observed $0.00$, and the
    best of four preprocessing conditions---tiling and an oracle crop-zoom
    included---reaches only $29.9\%$, placing that field structurally beyond
    the gate \emph{for this encoder} (\S\ref{sec:results:probe}). A
    classical-OCR hint injected as text raises the gate to 6/9; because it
    supplies externally the glyphs the encoder cannot carry, we report it as a
    confirmation of the bound, among the limitations
    (\S\ref{sec:limitations}).
  \item \textbf{The intervention that does move it, and what it costs.} We
    transplant an aspect-preserving visual tower trained at its operating
    resolution~\cite{wang2024qwen2vl} onto the same frozen decoder, with the
    same projector shape, corpus, step count and learning-rate schedule as the
    control, and the 8-nibble address goes from $0.00$ to $0.81$ exact
    ($\Delta$-NLL $+5.47$ against a shuffled-image control; per-digit accuracy
    $93.0\%$ on the four digits the renderer does not hold constant). It does
    so on a \emph{coarser} token budget than the tiling condition that
    recovered nothing---270 source pixels per visual token for tiling against
    787 for the transplant---which rules out any account of the ceiling
    monotone in pixels per glyph and relocates it from the preprocessing to the
    encoder's pretraining regime (\S\ref{sec:results:qwenswap}). This is a
    single run on a single synthetic generator, with the second pre-registered
    field 63\% contaminated and reported as secondary, and it ships nothing:
    the transplant keeps Qwen's own merger, and that configuration has no
    \texttt{llama.cpp} export path for this backbone
    (\S\ref{sec:discussion:encoder}). The export-shaped projector that
    replaces the merger (\S\ref{sec:arch:qwenline}) does not reproduce the
    gain on that field.
  \item \textbf{Grounding that is confirmed, and confined to what was trained.}
    On a later checkpoint of the Qwen-tower pathway,
    \texttt{domain\_expansion\_v1}---trained from scratch on the
    three-template instrumented corpus plus 2{,}802 examples over six newly
    instrumented templates, and not a member of the ablation matrix
    above---we measure B8: 2{,}040 items over 34 fields and 16 rendered-UI
    templates, ground truth taken from the renderers' own return values, and
    each field scored against the maximum of a shuffled-image control and a
    best-constant baseline, a control the nine-field gate did not have. Nine of
    34 fields clear the same pre-registered rule
    ($\mathrm{acc}_{\text{real}}\geq0.60$, $\Delta\geq0.40$), and the 25 that do
    not partition cleanly into templates never trained on, the ceilings already
    characterized above, and a single controlled case---%
    \texttt{log\_table.n\_rows} at $0.00$, below its own $0.35$ best-constant
    baseline, in the template where \texttt{log\_table.flavor} clears both
    controls at $0.93$---which motivates the hypothesis that the fine-tune
    teaches a lookup keyed to a (visual landmark, question) pair rather than an
    ability to read on-screen text (\S\ref{sec:results:b8},
    \S\ref{sec:discussion:landmark}). This revises none of the numbers above: it
    is a different checkpoint under a strictly stronger decision rule, and it
    licenses no statement about any (template, field) combination outside the
    ones trained.
  \item \textbf{Why the screenshot benchmark was exactly zero, and a scoped
    statement that the model sees.} Tool identification on B6 was $0.000$ on
    every checkpoint trained without B6's task format. We show that this was
    not a perception ceiling of the tower and not a wiring error: stock
    Qwen2-VL-2B, at exactly our pixel budget, transcribes the same 50 images at
    word recall $0.93$ ($0.52$ under our letterboxing) where our reference
    checkpoint scores $0.00$; the capture of pre-merge features is identical to
    Qwen's own first two operations; and three projector-only fixes that
    supervise transcription and open tool naming take B6 from $0.000$ to
    $0.311$ (on the 45 items of the five tools that run trained), then $0.200$,
    and finally $0.960$ on all 50 items, with word recall $0.48$--$0.51$ and
    28/28 on the items whose accepted aliases do not appear on screen
    (\S\ref{sec:results:b6fix}). We state the resulting claim at the strength
    the evidence supports---perception of rendered text is present and
    trainable through a frozen backbone, and the earlier zero was a task-format
    and renderer-exposure gap (\S\ref{sec:discussion:sees})---together with
    what it does not license: B6 is nearly in-distribution for the projector
    that scores $0.96$, that projector falls from $0.534$ to $0.289$ mean
    field accuracy on a B8 retention check against the reference checkpoint,
    tools never trained on
    were identified $0/5$ in the one run that held tools out, and no single
    checkpoint yet has both competences.
  \item \textbf{A methodological caution: probe recoverability does not predict
    end-to-end gain.} The same probe showed 2$\times$2 tiling lifting per-glyph
    recoverability from 60.8\%/62.4\% to 86.0\%/79.2\% on two independent
    fields. Implemented end-to-end, tiling made the model \emph{worse}: the gate
    fell to 1/9, the two fields the probe favored stayed at exactly $0.00$, and
    abstention collapsed from 0.74--0.82 to 0.14 as the hallucination rate rose
    to 0.86 (\S\ref{sec:results:tiling}). A linear probe on frozen features measures an
    optimal extractor under oracle supervision, not what a bounded adapter will
    learn to use---a distinction we did not find stated as an empirical result
    elsewhere, and one that cost us a full training cycle.
  \item \textbf{Four defects found in our own benchmark, and the hygiene point
    they make.} Auditing B6 revealed an unused loop variable that turned a
    nominal $n{=}50$ suite into an effective $n{=}10$ of byte-identical
    duplicates; the depicted product's name printed into the synthetic window's
    title bar, making tool identification partly an OCR task; substring
    alias matching that mis-scored 23 of 50 correct GPT-4o responses as
    hallucinations, with a bias favoring our own terse model over fluent Spanish
    baselines (\S\ref{sec:results:harness}); and, found only once a model
    scored above zero, accepted aliases that are printed on the screen or in
    the question itself---the bare word ``exploit'' for Metasploit, which every
    Metasploit screen shows three times---so that copying on-screen text can
    score as identification (\S\ref{sec:results:b6fix}). We retract the 0.08
    tool-identification score reported in an earlier draft of this work, and we
    report B6 tool identification split by whether an accepted alias is visible
    on screen.
  \item \textbf{A sub-2B cybersecurity VLM designed for native reasoning and
    tool tokens, as a deliverable system.} To our knowledge \model is the
    first sub-2B VLM specialized for security-tool imagery. It couples
    SigLIP-so400m to a 1.04B Spanish/LATAM security decoder whose vocabulary
    carries native \texttt{<|think|>}\ldots\texttt{</think>} and
    \texttt{<|tool\_call|>} tokens, and its curriculum \emph{trains}
    image-conditioned reasoning traces and MCP tool calls (phase 4c). We keep
    that apart from what is \emph{measured}: on B7 no checkpoint we evaluated
    opens a \texttt{<|think|>} block over the image (\texttt{think\_present}
    $0.0$), and tool calling is measured only in text, by B4
    (\S\ref{sec:arch:tokens}). The complete artifact---\texttt{model.gguf} plus a LLaVA-compatible
    \texttt{mmproj.gguf}---runs offline under \texttt{llama.cpp}
    (Section~\ref{sec:arch}). The system is delivered and its efficiency is
    measured, even though its visual-grounding quality is not yet established;
    the Qwen-tower checkpoints that carry the positive results above have not
    yet been exported to GGUF.
  \item \textbf{A four-phase curriculum extending the VectraYX line to vision.}
    We add a vision phase (4a alignment, 4b instruct, 4c think+tools) on top of
    the three-phase language curriculum, with an explicit token budget, replay
    ratios that preserve Spanish and tool competence, and a justified
    freeze/unfreeze schedule for 2$\times$A100-40GB
    (Section~\ref{sec:curriculum}). A post-hoc comparison of two vision runs,
    consistent with \texttt{tok\_emb} freezing contributing to the tool-use
    degradation of one of them but confounded with their learning rate and step
    count, is reported and labelled as such (\S\ref{sec:results}).
  \item \textbf{A domain-balanced multimodal cybersecurity corpus and its
    construction method.} 14{,}596 QA pairs over ten domains (reverse
    engineering, assembly, architecture, debugging, forensics, SOC, offense,
    crypto, competitive programming, LATAM-specific), generated from real source
    material and rendered as synthetic IDA/Ghidra/terminal images plus real
    tool screenshots, on a dual English/Spanish pipeline
    (Section~\ref{sec:dataset}).
  \item \textbf{Three new security-VLM benchmarks.} \textbf{B6\_vision} (tool
    identification and answer correctness over held-out screenshots),
    \textbf{B7\_think} (thinking-chain presence and quality), and
    \textbf{B8} (a wider, ground-truth-by-construction extraction gate over
    34 fields and 16 templates with a dual shuffled-image and best-constant
    control, \S\ref{sec:results:b8}), all released and reproducible
    end-to-end via a synthetic image renderer
    (Sections~\ref{sec:eval},~\ref{sec:eval:artifacts}).
\end{enumerate}

\noindent
One further question the architecture makes available is deliberately
\emph{not} in that list. Whether the backbone's periodic
NoPE~\cite{kazemnejad2023nope} layers help or hurt attention over the injected
729-token visual block is a combination present in no previously released VLM,
and we ran the ablation rather than deferring it: three backbone variants were
trained through alignment and gated, and the all-RoPE variant leads by a single
field of nine (\S\ref{sec:results:nope}). We report it as an open question
because that margin confounds the positional effect with an architecture--weight
mismatch we can size ($+0.135$ nats/token on held-out text) but not sign, and
because the statistic the ablation was designed around is a between-variant
difference in a metric that is at its floor for every variant. The variants, the
design, and the runner are released so the experiment does not depend on us
(Sections~\ref{sec:discussion},~\ref{sec:eval:artifacts}).

\noindent
We are deliberately conservative about claims. This preprint reports a
diagnosed negative result for the SigLIP pathway and two scoped positive
results for the Qwen-tower pathway; it does not claim a working general visual
model, and it withdraws a claim an earlier draft made. Where a question is not
estimable from the runs we could afford, we say so and release the design
rather than report a value a floor effect would make meaningless
(Section~\ref{sec:limitations}).

\section{Related Work}
\label{sec:related}

\paragraph{General-purpose VLMs.}
The modern open VLM recipe---a pretrained vision encoder, a lightweight
projector, and an LLM decoder, trained in an alignment-then-instruct
sequence---was crystallized by LLaVA~\cite{liu2023llava,liu2024llava15} and
BLIP-2~\cite{li2023blip2}, and scaled by InternVL~\cite{chen2024internvl},
Qwen-VL / Qwen2-VL~\cite{bai2023qwenvl,wang2024qwen2vl}, and
IDEFICS~\cite{laurencon2024idefics}. \model follows the LLaVA-style projector
recipe but differs in three ways: (i)~its LLM backbone is
trained from scratch for a narrow domain and language rather than being a
general 7B+ model; (ii)~the backbone uses periodic NoPE layers, which no prior
released VLM has, raising the visual-attention question we study; and
(iii)~it is trained to emit native reasoning and tool-call tokens rather than
relying on prompt-format conventions (whether it does so over images is
measured separately, \S\ref{sec:arch:tokens}).

\paragraph{Visual encoders.}
We use SigLIP~\cite{zhai2023siglip}, whose sigmoid contrastive objective
outperforms CLIP~\cite{radford2021clip} at matched compute and has become the
default encoder in recent strong VLMs. Both are ViTs~\cite{dosovitskiy2021vit}.
Our so400m/patch14/384 configuration yields 729 patch tokens with no CLS token,
which we inject directly into the decoder sequence.

\paragraph{Visual chain-of-thought.}
Textual chain-of-thought~\cite{wei2022cot} has a visual analogue: Visual
CoT~\cite{shi2024visualcot} shows that eliciting intermediate reasoning grounded
in image regions improves multimodal QA. Our \texttt{<|think|>} mechanism is in
this spirit but is \emph{native and trained} (the tokens are in the vocabulary
and the model is fine-tuned on think traces in phase 4c) rather than a
prompting strategy, and it targets security artifacts (addresses, opcodes,
packet fields, IOCs) rather than natural scenes. Trained is not the same as
demonstrated: on B7 no checkpoint we evaluated opens a think block over the
image (\S\ref{sec:arch:tokens}).

\paragraph{Domain-specialized VLMs.}
The closest analogue outside security is medical imaging, where
MedVersa~\cite{zhou2024medversa} and related generalist medical models show
that domain specialization of the VLM stack yields large gains over
zero-shot general VLMs on in-domain imagery. We argue security tooling imagery
is at least as specialized---synthetic-looking, text-dense, and semantically
dependent on exact tokens (a hex address, a CVE id, an opcode)---and thus a
strong candidate for the same treatment. To our knowledge no prior VLM targets
security-tool screenshots.

\paragraph{Text-dense and GUI imagery.}
Security-tool screenshots are closer to documents and graphical user
interfaces than to natural scenes. Donut~\cite{kim2022donut} and
Pix2Struct~\cite{lee2023pix2struct} established pixel-only reading of
text-dense documents and rendered UIs; ScreenAI~\cite{baechler2024screenai},
CogAgent~\cite{hong2024cogagent}, and Ferret-UI~\cite{you2024ferretui}
specialize VLMs for screenshot and GUI understanding, grounding, and agentic
interaction. This literature supports two of our design choices---retaining the
full high-resolution patch grid (text-dense frames punish token pooling) and
scoring on exact on-screen tokens---but none of these models targets security
\emph{semantics} (what an opcode sequence does, whether a decompiled function
is vulnerable), answers in Spanish, or fits a sub-2B offline envelope; and
none is trained with native structured-reasoning or tool-call tokens.

\paragraph{Text-only cybersecurity models and benchmarks.}
SecureBERT~\cite{aghaei2022securebert} and
VulBERTa~\cite{hanif2022vulberta} adapt encoder LMs to security text and source
code; CyberSecEval~\cite{bhatt2023cyberseceval} benchmarks the secure-coding and
offensive-capability behavior of generative LLMs. These are text-only and
English-centric. \model is complementary: it is multimodal, Spanish/LATAM-first,
and its text side descends from VectraYX-Nano~\cite{santillana2026vectrayx},
whose curriculum, tokenizer, and B1--B5 benchmark suite we extend here.

\paragraph{Compute-optimality and small models.}
Chinchilla~\cite{hoffmann2022chinchilla} scaling motivates over-training small
models on more tokens rather than growing parameters, which underlies the
VectraYX design choice of a 1B backbone trained on tens of billions of tokens
for edge deployment; Phi-3~\cite{abdin2024phi3} is a prominent demonstration
that small, data-curated models can be competitive. \model inherits this
philosophy and adds the multimodal dimension while staying sub-2B for offline
inference.

\paragraph{Positional encoding.}
NoPE~\cite{kazemnejad2023nope} shows decoder-only transformers can learn
position implicitly and generalize in length without explicit encodings; RoPE
~\cite{su2024roformer} is the standard alternative. VectraYX interleaves NoPE
every fourth layer with RoPE elsewhere. The interaction of NoPE with a
contiguous block of \emph{non-causal-order} visual tokens spliced into the
sequence is, to our knowledge, unstudied---we make it an explicit ablation, run
it, and report why its outcome does not yet settle the question
(Section~\ref{sec:discussion}).

\section{Architecture}
\label{sec:arch}

\begin{figure}[t]
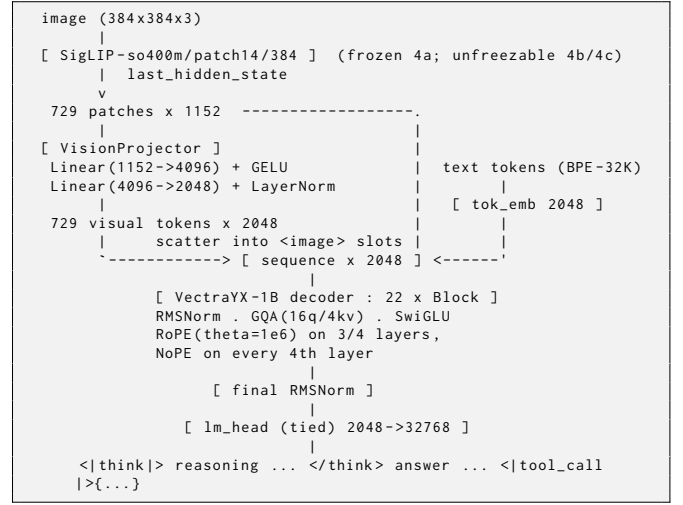

\centering
\begin{lstlisting}[basicstyle=\ttfamily\scriptsize]
  image (384x384x3)
        |
  [ SigLIP-so400m/patch14/384 ]  (frozen 4a; unfreezable 4b/4c)
        |  last_hidden_state
        v
   729 patches x 1152  ------------------.
        |                                |
  [ VisionProjector ]                    |
   Linear(1152->4096) + GELU             |  text tokens (BPE-32K)
   Linear(4096->2048) + LayerNorm        |        |
        |                                |   [ tok_emb 2048 ]
   729 visual tokens x 2048              |        |
        |     scatter into <image> slots |        |
        `------------> [ sequence x 2048 ] <------'
                              |
              [ VectraYX-1B decoder : 22 x Block ]
              RMSNorm . GQA(16q/4kv) . SwiGLU
              RoPE(theta=1e6) on 3/4 layers,
              NoPE on every 4th layer
                              |
                    [ final RMSNorm ]
                              |
                 [ lm_head (tied) 2048->32768 ]
                              |
      <|think|> reasoning ... </think> answer ... <|tool_call|>{...}
\end{lstlisting}
\caption{\model data flow. The projector maps 729 SigLIP patch tokens to the
2048-dim decoder space; the projected tokens replace the \texttt{<image>}
placeholder slot in the token sequence and are consumed by the decoder exactly
like text tokens.}
\label{fig:arch}
\end{figure}

\model is a three-stage LLaVA-style stack: a frozen visual encoder, a trainable
projector, and a from-scratch language decoder. Figure~\ref{fig:arch} (ASCII)
summarizes the data flow.

\subsection{Visual encoder}
We use \texttt{google/siglip-so400m-patch14-384}~\cite{zhai2023siglip}: a
so400m Vision Transformer producing a \texttt{last\_hidden\_state} of
$729{\times}1152$ (27$\times$27 patches, no CLS token). We take the full patch
grid; text-dense security imagery (disassembly, packet lists) has information
spread across the whole frame, so we do not pool. The encoder is frozen in
phase~4a and optionally unfrozen thereafter (Section~\ref{sec:curriculum}); we
default to keeping it frozen throughout because the projector plus a trainable
backbone already provide sufficient adaptation capacity at our data scale, and
freezing halves activation memory on 2$\times$A100.

\subsection{Projector}
The projector is a two-layer MLP: $\mathrm{Linear}(1152{\to}4096)$, GELU,
$\mathrm{Linear}(4096{\to}2048)$, followed by LayerNorm. The output LayerNorm is
a deliberate choice: the decoder was pretrained purely on text-token embeddings,
whose per-dimension scale is set by the tied embedding matrix; normalizing the
visual tokens to a comparable scale prevents them from dominating or vanishing
in the first attention layer, without a learned per-token gate. The projector is
$\approx$13.1M parameters and is the \emph{only} trainable component in
phase~4a.

\subsection{Language backbone}
The backbone is \textsc{VectraYX-1B}: 22 layers, $d_\mathrm{model}{=}2048$,
$d_\mathrm{ffn}{=}5504$, GQA~\cite{ainslie2023gqa} with 16 query / 4 KV heads,
SwiGLU~\cite{shazeer2020glu}, RMSNorm~\cite{zhang2019rmsnorm} pre-norm, tied
embeddings, a 32{,}768-entry BPE~\cite{sennrich2016bpe} vocabulary, and a
$z$-loss auxiliary. RoPE
with $\theta{=}10^6$ is applied on three of every four layers; every fourth
layer applies \emph{no} positional encoding (NoPE). The measured parameter count
of the instantiated model is 1{,}041.9M (``1B''). Adding the 13.1M projector
gives 1{,}055M ($\approx$1.05B) parameters in the full vision model; we report
the 1.04B decoder-only count in the abstract (the standalone artifact) and the
$\approx$1.05B full-vision-model count in the curriculum
(Table~\ref{tab:curriculum}). How many of those are \emph{trainable} in phases
4b/4c depends on the path taken: the configured recipe trains projector and
backbone jointly, whereas every phase-4b run reported in
Section~\ref{sec:results} uses the LoRA path and trains 17.8--25.8M
(\S\ref{sec:curriculum:asrun}). The backbone is warm-started
from the phase-3 language checkpoint (9.2B tokens phase~1 $+$ $\approx$50B
tokens phase~2 $+$ $\approx$6B tokens phase~3 tooling); the vision phases
continue from there.

\subsection{Visual token injection and positional handling}
\label{sec:arch:inject}
We adopt LLaVA-style substitution: the tokenizer reserves an \texttt{<image>}
special token; at data-build time the single placeholder is expanded to 729
placeholder positions, and at forward time the projected visual tokens are
scattered into those positions (in-place, constant length). The visual tokens
therefore occupy \emph{ordinary contiguous sequence positions}. This has a
direct consequence for the NoPE layers: on RoPE layers each visual token
receives a rotary phase according to its absolute position in the sequence,
imposing a left-to-right order on a patch grid that is intrinsically 2D; on NoPE
layers no such order is imposed and the visual block is permutation-equivariant
with respect to attention. Whether this asymmetry helps (NoPE layers ``see'' the
patch set without a spurious 1D order) or hurts (RoPE layers give conflicting
1D order signals) is an open question we make ablatable
(Section~\ref{sec:discussion}). We extend the precomputed RoPE tables to cover
$729 + \text{max\_seq\_len}$ positions so the visual prefix does not exhaust the
positional budget.

\subsection{Native reasoning and tool tokens: what is trained and what is measured}
\label{sec:arch:tokens}
The tokenizer reserves \texttt{<|think|>}/\texttt{</think>} (IDs 14/15),
\texttt{<|step|>}/\texttt{</step>} (16/17), \texttt{<|tool\_call|>} (8), and the
turn delimiter \texttt{<|end|>} (7). Phase~3 trains both behaviors in text. In
phase~4c the model is \emph{trained} on traces that, given a security
screenshot, open a \texttt{<|think|>} block (e.g., ``the pane header says
\emph{IDA}; the prologue pushes callee-saved registers; this is a context
switch''), then emit an answer, and where appropriate a
\texttt{<|tool\_call|>} with a JSON payload consumed by an MCP server.

What is \emph{measured} is narrower, and we keep the two apart throughout. On
B7 the released checkpoint, and every checkpoint of the SigLIP pathway, opens a
\texttt{<|think|>} block on none of the 50 items (\texttt{think\_present}
$0.0$, \S\ref{sec:results:b6}). In the Qwen-tower pathway
(\S\ref{sec:arch:qwenline}) the training answers carry a short trace,
\texttt{<|think|>}\ldots\texttt{</think>} followed by the answer, built
deterministically from the verified field value with no teacher model, and the
model reproduces that templated trace, closed by \texttt{</think>}, in its
outputs (\S\ref{sec:results:b6fix}); but on the one checkpoint of that pathway
evaluated on B7 it did not open a \texttt{<|think|>} block when asked to, again
$0.0$. We therefore claim trained reasoning tokens and a reproduced answer
format, not image-grounded reasoning. Tool calling is measured only in text,
by B4 before and after the vision phase (Table~\ref{tab:b1b5}); no
image-conditioned \texttt{<|tool\_call|>} is evaluated in this paper.

\subsection{Export for offline inference}
The stack is designed to split cleanly into the two GGUF~\cite{ggml2024gguf}
artifacts \texttt{llama.cpp} expects for LLaVA: \texttt{mmproj.gguf} (SigLIP tower $+$
projector) and \texttt{model.gguf} (the decoder). Our implementation exposes
\texttt{export\_llm\_state\_dict()} and \texttt{export\_mmproj\_state\_dict()}
with no cross-tensor tying between them, so the decoder checkpoint contains no
visual weights and vice versa. The decoder already matches the Llama tensor
convention (RMSNorm, SwiGLU, RoPE, GQA, no biases) that \texttt{llama.cpp}
supports natively; NoPE layers export as RoPE layers with the rotary application
disabled per layer.

\subsection{Second encoder pathway: the Qwen2-VL tower with an export-shaped projector}
\label{sec:arch:qwenline}
The encoder transplant of \S\ref{sec:results:qwenswap} keeps Qwen2-VL's
pretrained merger and feeds its 1{,}536-dimensional output to a second MLP, a
layout with no \texttt{llama.cpp} export path for this backbone
(\S\ref{sec:discussion:encoder}). The checkpoints behind B8
(\S\ref{sec:results:b8}) and the B6 study (\S\ref{sec:results:b6fix}) use a
different arrangement, built so that the vision side keeps the shape of an
\texttt{mmproj}. The frozen Qwen2-VL-2B~\cite{wang2024qwen2vl} vision tower
runs unchanged; we capture the input to its merger---each group of
2$\times$2 adjacent patch features, $4\times1280=5120$ dimensions---apply
Qwen's own pretrained \texttt{ln\_q}, and replace the merger's MLP with a
trainable projector of the same layout except for the output width:
$\mathrm{Linear}(5120{\to}5120)$, GELU, $\mathrm{Linear}(5120{\to}2048)$. The
first layer is initialized from Qwen's \texttt{merger.mlp[0]}; the second has
no shape-compatible counterpart (Qwen's maps to 1{,}536) and starts fresh. We
verified by reading the \texttt{transformers} implementation that the capture
plus \texttt{ln\_q} reproduces the first two operations of Qwen's own
\texttt{PatchMerger.forward} exactly, so the only difference from the stock
model on the vision side is the weights of this MLP. The decoder is the same
frozen phase-3 backbone as everywhere else in the paper, and in this pathway
it is never unfrozen: only the projector trains. Every image is letterboxed
top-left onto a 1280$\times$768 canvas and resized by Qwen's processor to a
fixed budget of $392^2$ pixels, about 180 visual tokens.

Bypassing the pretrained merger is not free. On the three-template pilot corpus
under the 1{,}496-step recipe of \S\ref{sec:results:qwenswap}, this projector
reads \texttt{hexdump.base\_addr} at $0.00$ exact ($n{=}40$, shuffled control
$0.00$), with 0.425 accuracy per free nibble against a chance rate of 0.0625,
where the transplant that keeps the merger reads $0.81$; coarse fields pass
through it cleanly (\texttt{siem.events} and \texttt{siem.alerts} $1.00$,
\texttt{siem.hosts} $0.97$, each against a shuffled control of at most
$0.03$). That is a single run, and it means the export-shaped pathway does not
inherit the transplant's result on the field that motivated it. Three
checkpoints of this pathway appear below: \texttt{domain\_expansion\_v1}
(\S\ref{sec:results:b8}); \texttt{combined\_v2}, the reference checkpoint of
the pathway, trained from scratch for 3{,}000 steps on 11{,}863 records over all
sixteen B8 templates; and three research projectors trained for B6
(\S\ref{sec:results:b6fix}). An earlier projector of exactly this shape,
aligned on the released vision dataset, has been converted by substituting its
two tensors into a stock Qwen2-VL \texttt{mmproj.gguf} and served through
\texttt{llama.cpp}'s multimodal command-line tool with the image pinned to 180
tokens; it is published in the experimental repository of
\S\ref{sec:eval:artifacts}. None of the checkpoints above has yet been
converted, so for the weights this paper reports on, the export is designed
for and not demonstrated; their raw projector weights are released in the same
repository (\S\ref{sec:eval:artifacts}).

\section{Training Curriculum}
\label{sec:curriculum}

\model is the vision extension of the VectraYX three-phase language curriculum.
We summarize the inherited language phases and then detail the new vision phase.

\subsection{Inherited language phases (backbone)}
The 1B backbone is produced by:
\emph{Phase~1} (9.2B tokens) general Spanish pretraining;
\emph{Phase~2} ($\approx$50B tokens) a three-block curriculum---BlkA factual,
BlkB code$+$math, BlkC domain$+$decay;
\emph{Phase~3} ($\approx$6B tokens) tooling specialization with a 30\% tool-SFT / 25\%
curated-cyber / 20\% reasoning / 15\% Spanish-replay / 6\% code-replay /
4\% math-replay mixture. Phase~3 establishes the native \texttt{<|tool\_call|>}
and \texttt{<|think|>} behavior in text before any image is seen. Vision training
starts from the phase-3 checkpoint.

\subsection{Vision phase (4a/4b/4c)}
\label{sec:curriculum:vision}
Table~\ref{tab:curriculum} gives the hyperparameters (from
\texttt{vision\_p4.json}). The design rationale:

\paragraph{Phase 4a --- alignment (projector only).}
Only the 13.1M-parameter projector trains; SigLIP and the 1B backbone are
frozen. The goal is narrow: learn a mapping from SigLIP's visual space into the
backbone's token-embedding space, so that visual tokens are ``legible'' to a
decoder that has never seen them. A high learning rate ($10^{-3}$) and a single
epoch suffice because the projector is small and the target (embedding-space
alignment) is a comparatively easy regression. Freezing the backbone here
prevents the still-random visual signal from corrupting the language and tool
competence built in phases 1--3.

\paragraph{Phase 4b --- instruct (projector + backbone).}
The projector and the full backbone train (SigLIP still frozen) on a 60\%
vision / 40\% text-replay mixture. The 40\% replay (conversational $+$ CVE Q\&A
$+$ tool traces, drawn from the phase-3 language mixture) is the anti-forgetting
mechanism: without it, the backbone over-fits the visual QA distribution and
regresses on B4 (tool use) and B5 (conversational)---the same catastrophic
forgetting dynamic documented for the language phases~\cite{ibrahim2024simple,french1999catastrophic}.
The low learning rate ($2{\times}10^{-5}$) and three epochs reflect that we are
adapting a converged 1B model, not training from scratch.

\paragraph{Phase 4c --- think + tools (projector + backbone).}
Same trainable set as 4b, on a 50\% vision / 25\% tool-replay / 25\%
text-replay mixture, where the vision portion now contains \texttt{<|think|>}
reasoning traces and \texttt{<|tool\_call|>} completions conditioned on images.
The even lower learning rate ($10^{-5}$) and two epochs are intended to sharpen
the reasoning and tool behavior without destabilizing the visual grounding from
4b; whether they do is measured in \S\ref{sec:arch:tokens}, and over images
they do not yet.

\begin{table}[t]
\centering
\caption{Vision-phase hyperparameters as \emph{configured} in
\texttt{vision\_p4.json} (2$\times$A100-40GB, DDP, BF16). ``Replay'' = fraction
of text-only anti-forgetting data. The runs that produced the grounding results
of Section~\ref{sec:results} deviate from this recipe in three respects---LoRA
rather than full fine-tuning, two epochs rather than three, and a phase-4b
learning rate corrected to $1{\times}10^{-4}$---itemized in
\S\ref{sec:curriculum:asrun}.}
\label{tab:curriculum}
\footnotesize
\begin{tabular}{@{}lccccc@{}}
\toprule
Phase & Trainable & LR & Ep. & Vis:Replay & Seq \\
\midrule
4a align   & proj (13M)      & $1{\times}10^{-3}$ & 1 & 100:0   & 1280 \\
4b instruct& proj$+$LLM (1.05B)& $2{\times}10^{-5}$ & 3 & 60:40   & 1280 \\
4c think   & proj$+$LLM (1.05B)& $1{\times}10^{-5}$ & 2 & 50:50$^\dagger$ & 1280 \\
\bottomrule
\end{tabular}
\\[2pt]
{\footnotesize $^\dagger$4c replay is 25\% tool $+$ 25\% text.}
\end{table}

\subsection{Compute and memory budget}
On 2$\times$A100-40GB, full fine-tuning of the 1.05B backbone in BF16 under DDP
fits comfortably with the frozen SigLIP encoder at sequence length~$\approx$1280
(729 visual $+$ text) and batch size~4. We provide a self-contained
LoRA~\cite{hu2022lora} path (rank-configurable, on the backbone's q/k/v/o
projections and optionally its SwiGLU FFN) for the memory-constrained
single-GPU case; full fine-tuning is the configured default because the
backbone is small enough and full FT avoids an adapter-merge step before GGUF
export. In practice the single-GPU constraint bound throughout, and every
phase-4b run we report took the LoRA path
(\S\ref{sec:curriculum:asrun}). Token budget for the vision phase is modest relative to the language
phases: the theoretical upper bound is 14{,}596 QA pairs $\times$ (1 + 3 + 2)
effective epochs across sub-phases, on the order of $10^{8}$ forward-pass tokens
(sequence length~1280); in the actually-run schedule (below) the optimizer
sees a subset of these because replay draw is without replacement and several
sub-stage continuations start mid-stream, so the measured total is lower.
Concretely, in the first full pass through the curriculum---the run whose
text-competence numbers appear in Table~\ref{tab:b1b5}---on
2$\times$A100-40GB\,DDP with batch size~4 and sequence
length~1280, phase~4a ran 228 steps (24\,min, $\approx$2.3\,M tokens), phase~4b
ran 819 steps (43\,min, $\approx$8.4\,M tokens), and phase~4c ran 546 steps
(63\,min, $\approx$5.6\,M tokens)---$\approx$16\,M tokens total, two orders of
magnitude below phase~2, consuming $\approx$2.2 hours of 2$\times$A100-40GB wall
time for all three sub-stages, consistent with the alignment-then-light-instruct
VLM recipe~\cite{liu2024llava15}.

\subsection{The runs behind the grounding results}
\label{sec:curriculum:asrun}
The grounding gate, probe, and ablation results of Section~\ref{sec:results}
come from a later and separately budgeted set of runs, on single rented GPUs
rather than the 2$\times$A100 node, and we state their recipe here so no number
in Section~\ref{sec:results} has to be traced back to a configuration file. All
of them train on the instrumented extraction corpus described in
\S\ref{sec:results:gate}---or, in two of the configurations, on that corpus
extended with the transfer or prose domains of \S\ref{sec:results:gate}---rather
than on the 14{,}596-pair instruct corpus of Section~\ref{sec:dataset}.
\emph{Phase 4a} trains the 13.1M projector alone with the backbone frozen, at
$10^{-3}$ for 8 epochs (1{,}496 steps); the three NoPE$\times$vision variants of
\S\ref{sec:results:nope} differ only in the backbone's positional schedule and
share this recipe exactly.
\emph{Phase 4b} resumes from that phase-4a checkpoint and takes the LoRA path at
rank 16, $\alpha{=}16$: 4.7M adapter parameters on the 88 attention projections,
or 12.7M when the SwiGLU FFN is included, on top of the trainable 13.1M
projector---17.8M or 25.8M trainable in total, against 1{,}055M in the model.
Each run is two epochs over a corpus of between 9{,}600 and 19{,}980 records
(600 steps for the smallest, 1{,}124 for the 18{,}000-record transfer corpus)
at the corrected learning rate of
$1{\times}10^{-4}$, except the one labelled ``original LR'' in
Table~\ref{tab:gate}, which is the same recipe at the inherited
$5{\times}10^{-6}$ (\S\ref{sec:results:fixes}). The tiling configuration
additionally widens the projector input from 1152 to 4608 and so re-initializes
it, raising the trainable count to 32.0M and forcing a fresh phase-4a alignment
of 562 steps (\S\ref{sec:results:tiling}).
This substitution of a bounded adapter for the configured full fine-tune is a
limitation of the reported evidence, not an incidental detail, and we treat it
as one in \S\ref{sec:limitations}.
The checkpoint behind B8 (\S\ref{sec:results:b8}) is later again and is not one
of these runs: \texttt{domain\_expansion\_v1} trains from scratch rather than
resuming, for 2{,}200 steps on 8{,}802 examples---the same instrumented
extraction corpus plus 2{,}802 examples over six further instrumented
templates---and its recipe is stated with its result rather than here, so that
the configurations compared in Table~\ref{tab:gate} are not confused with it.
It belongs to the Qwen-tower pathway of \S\ref{sec:arch:qwenline}, as do
\texttt{combined\_v2} and the three B6 projectors of
\S\ref{sec:results:b6fix}; none of these phases 4a--4c apply to that pathway,
in which the decoder stays frozen and only the projector is trained.

\subsection{Loss masking}
Throughout 4a--4c the loss is computed on assistant tokens only (the span from
after \texttt{<|assistant|>} through \texttt{<|end|>}), identical to the
VectraYX language SFT. Visual placeholder positions are always masked
(label $-100$); the model is never asked to ``predict'' a visual token.

\section{Multimodal Corpus}
\label{sec:dataset}

\subsection{Overview and statistics}
The vision corpus comprises 14{,}596 multimodal QA pairs across ten
cybersecurity domains. Table~\ref{tab:corpus} gives the per-domain counts (from
the released \texttt{qa/*.jsonl} manifests). Each record has the schema
\texttt{\{messages, image\_path, image\_desc, template\_hint, domain,
source\_file\}}; \texttt{image\_path=null} marks a text-only replay record.

\begin{table}[t]
\centering
\caption{Multimodal QA corpus by domain (14{,}596 pairs total).}
\label{tab:corpus}
\small
\begin{tabular}{lr|lr}
\toprule
Domain & Pairs & Domain & Pairs \\
\midrule
offense (offensive)        & 4{,}282 & debugging   & 927 \\
soc (blue-team)            & 2{,}254 & crypto      & 516 \\
arch (architecture)        & 1{,}973 & re (rev.\ eng.) & 543 \\
forense (forensics)        & 1{,}347 & latam       & 120 \\
asm (assembly)             & 1{,}332 & icpc (comp.\ prog.) & 1{,}302 \\
\midrule
\multicolumn{3}{l}{\textbf{Total}} & \textbf{14{,}596} \\
\bottomrule
\end{tabular}
\end{table}

\subsection{Source material and dual-track pipeline}
Corpus construction uses a dual English/Spanish pipeline: a \texttt{raw\_en/}
track (English source material---kernel assembly references, exploit writeups,
CTF/ICPC problems, forensics documentation) and a \texttt{raw\_es/} track
(Spanish-native security content plus machine translations of the English
track). QA pairs are generated per source chunk; each pair carries the
\texttt{source\_file} for provenance. A representative \texttt{asm} record is
generated from a Linux kernel entry-point (\texttt{arch/x86/entry/entry\_64.S})
and asks the model to interpret a \texttt{\_\_switch\_to\_asm} disassembly pane.

\subsection{Image generation}
Two image sources feed the corpus. \emph{(1)~Synthetic renders}: security-tool
panes are rendered programmatically from the QA \texttt{image\_desc}/\texttt{template\_hint}
fields---IDA/Ghidra disassembly and decompiler panes, terminal sessions (Nmap,
Volatility, hashcat, msfconsole), and packet lists (Wireshark)---using
deterministic templating so that the exact on-screen tokens (addresses, opcodes,
CVE ids, ports) are known ground truth. This is essential for scoring: unlike
natural-image VQA~\cite{marino2019okvqa}, and more stringently than in
read-the-text VQA~\cite{singh2019textvqa}, correctness here hinges on
reproducing exact technical tokens.
\emph{(2)~Real screenshots}: a smaller set of real tool captures grounds the
synthetic distribution. The synthetic renderer used by our benchmark
(Section~\ref{sec:eval}) is released so the image side is reproducible without
the private screenshot set.

\subsection{Synthetic QA generation method}
QA pairs are produced by an LLM-in-the-loop pipeline over source chunks: each
chunk yields questions that require reading specific on-screen entities and an
answer grounded in the source, with an \texttt{image\_desc} that specifies what
the accompanying image must show. The recorded generation cost for the
factual/technical subset (arch, asm, debugging, re, icpc) was
$\approx$\$3.74 USD for 6{,}077 pairs over 2{,}029 chunks
($\approx$2.77M input / 1.52M output tokens), illustrating the low cost of the
approach; the offense/soc/forense/crypto/latam subsets were generated by the
same method. \textbf{Caveat (marked honestly):} LLM-generated QA can contain
factual errors and can leak the answer into the question; we describe the
validation gap and the required human audit in Section~\ref{sec:limitations}.

\subsection{Domain balance and its rationale}
The corpus is intentionally offense/soc-heavy (45\% combined) because those are
the highest-volume real analyst workflows, but retains a long tail (crypto,
latam) so the benchmark can measure whether low-resource domains collapse. The
\texttt{latam} domain (120 pairs) is small and is explicitly flagged as
under-powered for per-domain claims; it exists to probe, not to certify, LATAM
regional grounding. Text-only replay records (drawn from the phase-3 language
mixture) are interleaved per the phase ratios of
Section~\ref{sec:curriculum}; they carry \texttt{image\_path=null} and are used
verbatim to preserve tool and conversational competence.

\paragraph{Scope note.}
This corpus is the instruct/think corpus of the vision phase, and it is the one
the first full curriculum pass trained on. It is \emph{not} the corpus behind
the grounding gate of \S\ref{sec:results:gate}: measuring extraction against a
literal on-screen string requires a renderer that returns its own ground truth,
which this corpus's generator does not, so the runs reported in
Section~\ref{sec:results} use a separate instrumented corpus built for that
purpose (\S\ref{sec:curriculum:asrun}, \S\ref{sec:results:gate}), extended with
six further instrumented templates for the checkpoint B8 is measured on
(\S\ref{sec:results:b8}).

\section{Evaluation Design}
\label{sec:eval}

We evaluate along the inherited text axes B1--B5 (to verify the vision phases do
not regress language and tool competence) and two new vision axes, B6\_vision
and B7\_think. All benchmarks are released and runnable against a
\texttt{llama.cpp}/Ollama-served GGUF export.

\subsection{Inherited text benchmarks (regression guard)}
B1--B5 are the VectraYX suite~\cite{santillana2026vectrayx}: B1 CVE
Q\&A keyword recall, B2 threat classification, B3 security command completion,
B4 tool selection (the project's hard mandate: a model is not released for
production use below B4~$>$~0.10), B5 conversational quality. We report B1--B5
\emph{before} and \emph{after} the vision phase to quantify forgetting. The
acceptance criterion is no regression on B4/B5 relative to the phase-3 backbone;
since every run is single-seed (\S\ref{sec:eval:artifacts}), we apply it as a
descriptive comparison against the step-to-step band of the phase-3 trajectory
and not as a significance test, which these data cannot support.

\subsection{B6\_vision}
\label{sec:eval:b6}
B6\_vision is 50 held-out image$+$question pairs over security-tool screenshots
spanning re, asm, soc, offense, forense, and crypto. The count is 50
\emph{distinct} items only after the harness fix of
\S\ref{sec:results:harness}: the generator originally emitted five templates
nine times each byte-identically, so the suite as first reported had an
effective $n{=}10$, and any interval computed over it treated duplicates as
independent draws. All B6 numbers in this paper come from the corrected
generator (45 distinct variants, 50/50 unique \texttt{screen\_text}). Two
metrics per item:
\begin{itemize}[leftmargin=1.4em]
  \item \textbf{tool\_identification} --- 1 if the response names the depicted
    tool (accepting canonical aliases, e.g.\ ``IDA''/``Hex-Rays''/``disassembler''),
    else 0. The intent is to isolate \emph{visual} recognition from generic
    domain knowledge; on 22 of the 50 items an accepted alias is itself
    visible on screen, so we report the metric split by that property
    (\S\ref{sec:results:b6fix}).
  \item \textbf{answer\_correctness} --- 1 if the response states the key
    technical fact the image demonstrates (e.g.\ ``context switch'' for the
    \texttt{\_\_switch\_to\_asm} pane, ``buffer overflow'' for the unchecked
    \texttt{strcpy} decompilation), else 0.
\end{itemize}
Where the private screenshot set is unavailable, the released benchmark
synthesizes each item's image deterministically (PIL renders of the exact
on-screen text), so B6 is reproducible end-to-end.

\subsection{B7\_think}
\label{sec:eval:b7}
B7\_think reuses the B6 images but prompts for a \texttt{<|think|>}\ldots
\texttt{</think>} reasoning block before the answer. Two metrics:
\begin{itemize}[leftmargin=1.4em]
  \item \textbf{think\_present} --- 1 if the model actually opens a \texttt{
    <|think|>} block (native structured reasoning); 0 otherwise.
  \item \textbf{chain\_quality} --- normalized count of distinct on-screen
    references inside the think block (tool name, keywords, the key fact),
    saturating at four references. This rewards \emph{grounded} reasoning that
    cites what is actually on the screen rather than generic boilerplate. For
    models that do not open a \texttt{<|think|>} block
    (\texttt{think\_present}=0)---including non-native baselines such as the
    frontier cloud VLM, which lacks the special token---the metric falls back
    to scoring a leading reasoning paragraph the model does emit, so a
    non-zero \texttt{chain\_quality} here measures generic grounded-recall
    quality and is \emph{not} comparable to a native-token chain of the same
    score; we flag this distinction wherever baseline numbers are reported.
\end{itemize}
B7 deliberately does not reward long chains per se; it rewards chains that
reference concrete visual evidence, following the grounding emphasis of Visual
CoT~\cite{shi2024visualcot}.

\subsection{B8: a wider extraction gate with two controls}
\label{sec:eval:b8}
B6 and B7 ask for free-form identification and explanation; the nine-field
extraction gate of \S\ref{sec:results:gate} asks instead for a literal on-screen
string, but covers three templates and scores against a shuffled-image control
alone. B8 widens the first and strengthens the second: 2{,}040 items over 34
fields and 16 rendered-UI templates at $n{=}60$ per field, with every item's
ground truth being the renderer's own return value rather than an answer written
about the image afterwards. It keeps the gate's decision rule
($\mathrm{acc}_{\text{real}}\geq0.60$ and $\Delta\geq0.40$, frozen before the
run) and changes what the rule is taken against: each field is scored against
the maximum of the within-stratum shuffled control and a best-constant
baseline---the accuracy obtained by always emitting that field's modal
ground-truth value---so a field with a skewed answer distribution cannot be
mistaken for a grounded one. B8 is reported in full on a single later
checkpoint than the configurations behind the gate (with a pass count for one
further checkpoint), and its outcome is a partition of the 34
fields rather than an aggregate score (\S\ref{sec:results:b8}); as with the
gate, we report every field separately and never pool them.

\subsection{Baselines}
\label{sec:eval:baselines}
The comparison set the design calls for, all zero-shot on B6/B7 unless noted:
\begin{itemize}[leftmargin=1.4em]
  \item \textbf{A small, data-curated multimodal model} of the
    Phi~\cite{abdin2024phi3} family (Phi-4-multimodal) --- the closest baseline
    in \emph{footprint} and the fairest efficiency comparison.
  \item \textbf{1B text-only backbone} --- the phase-3 model with the image
    removed (question only). This critical ablation measures how much of B6
    is answerable from text alone; a high score would indict the benchmark,
    not praise the model.
\end{itemize}
Since the paper's deployment claim is about sub-2B offline inference, not about
beating larger VLMs on quality, the design additionally calls for a frontier
cloud VLM (GPT-4o) as an explicitly labeled capability ceiling---the quality the
offline model trades away for sovereignty---rather than as a peer baseline.

\paragraph{Status of the baselines.}
None of the three is reported as designed. Baseline runs predating the
harness fix of \S\ref{sec:results:harness} are withdrawn along with everything
else measured on that harness, and we did not re-run them. A GPT-4o run is what
exposed the substring-matching defect, and we report it only as the evidence for
that defect; it was scored by the pre-fix harness and gives no ceiling for any
number in this paper. Two diagnostic measurements take the place of a ceiling
and a control (\S\ref{sec:results:b6fix}): stock Qwen2-VL-2B run on B6 at our
own pixel budget, as a like-for-like perception ceiling for the tower our
second pathway uses; and the text-only control in one of its two forms, the
frozen backbone given each item's on-screen text instead of the image. The
question-only form remains unmeasured, and we flag it in
Section~\ref{sec:limitations}.

\subsection{Metrics reporting protocol}
\label{sec:eval:protocol}
The protocol was fixed in advance, following the VectraYX-Nano one: headline
numbers are reported over $N{=}4$ seeds with mean$\pm$std where the training run
permits, and single-seed numbers are labeled as such---in this paper, as the
paragraph below records, that is every number. For every claimed pairwise
comparison the protocol additionally calls for a paired bootstrap confidence
interval over benchmark items (10K resamples), since at 50 items per suite seed
variance alone understates
the evaluation uncertainty. This protocol is only sound over \emph{distinct}
items, which the suite provides only after the fix noted in
\S\ref{sec:eval:b6}; intervals reported in an earlier draft over the
pre-fix suite are withdrawn. We do not report any B6/B7 number that has not
been produced by the released, corrected harness. B1--B5 for the phase-3
backbone are reported in Section~\ref{sec:results}; corrected B6/B7 results,
which are at or within two items of fifty of the floor for every checkpoint not
trained on B6's task format, are reported in \S\ref{sec:results:b6}, and those
of the three projectors that were trained on it in \S\ref{sec:results:b6fix}. Because the primary grounding result of this paper is a
per-field exact-match gate rather than an aggregate score
(\S\ref{sec:results:gate}), we additionally report every gate field separately
and never pool them into a single number: a pooled score would hide the fact
that the pass set is invariant across interventions, which is the finding.

\subsection{Released artifacts and auditability}
\label{sec:eval:artifacts}
We release the weights, the evaluation harnesses, the training checkpoints and
the per-item outputs behind most numbers in this paper, so that they can be
audited and rerun; the list below states where each is, and the paragraph after
it states what is not released. Concretely:
\begin{itemize}[leftmargin=1.4em]
  \item Inference-ready GGUF exports: the text-only backbone
    (\href{https://huggingface.co/jsantillana/vectrayx-1b}{\texttt{jsantillana/vectrayx-1b}})
    and the full multimodal stack of the SigLIP pathway
    (\href{https://huggingface.co/jsantillana/vectrayx-vision-1b}{\texttt{jsantillana/vectrayx-vision-1b}},
    \texttt{model.gguf} + \texttt{mmproj.gguf}) for \texttt{llama.cpp}/Ollama.
  \item The training checkpoints of the SigLIP pathway for every phase
    (phase-1 through phase-4b SFT, intermediate steps) at
    \href{https://huggingface.co/jsantillana/vectrayx-vision-1b-checks}{\texttt{jsantillana/vectrayx-vision-1b-checks}},
    so the per-checkpoint trajectory reported in
    Table~\ref{tab:pretrain_snapshot} and Table~\ref{tab:b1b5} can be audited
    step by step, including the weight-norm diagnostics used to detect the
    phase-4b resume defects (Section~\ref{sec:limitations}).
  \item The benchmark harness and synthetic image renderer
    (\texttt{bench\_vision\_1b.py}, \texttt{render\_synthetic\_images.py}),
    released alongside the model, so B6/B7 can be reproduced end-to-end without
    the private screenshot set.
  \item The Qwen-tower pathway (\S\ref{sec:arch:qwenline}), in
    \href{https://huggingface.co/jsantillana/vectrayx-vision-1b-qwen-experimental}{\texttt{jsantillana/vectrayx-vision-1b-}\allowbreak\texttt{qwen-experimental}}:
    the earlier transplant and the \texttt{domain\_expansion\_v1} checkpoint;
    the reference checkpoint
    (\texttt{projectors/combined\_v2\_projector\_step03000.pt}) and the three
    B6 projectors (\texttt{projectors/b6b7\_fix\_v1\_projector.pt},
    \texttt{\ldots v2\ldots}, \texttt{\ldots v3\ldots}), each a raw PyTorch
    state dict of the projector with no GGUF export; and a folder
    \texttt{b6\_fix\_evidence/} with the scripts and logs behind
    Tables~\ref{tab:b6ceiling}--\ref{tab:retention} and a README mapping each
    table to its script and log.
  \item The training corpus of \texttt{combined\_v2}
    (\href{https://huggingface.co/datasets/jsantillana/vectrayx-combined-v2-corpus}{\texttt{jsantillana/vectrayx-combined-v2-corpus}}):
    \texttt{combined\_v2\_train.jsonl} (11{,}863 records), the held-out split,
    and the 6{,}000 synthetic renders.
  \item B8's corpus, items, scorer, and per-item model outputs
    (\href{https://huggingface.co/datasets/jsantillana/vectrayx-b8-benchmark}{\texttt{jsantillana/vectrayx-b8-benchmark}}),
    for \texttt{domain\_expansion\_v1} and, in
    \texttt{b8\_results\_combined\_v2.jsonl}, for \texttt{combined\_v2}, with
    that checkpoint's B6/B7 results in \texttt{b6\_b7\_combined\_v2.json}.
\end{itemize}

\paragraph{What is not released.} Three gaps remain, and each limits what can be
rerun. v3's per-item B6 answers were not logged, so its Metasploit credits
cannot be re-scored without the ``exploit'' alias (\S\ref{sec:results:b6fix});
only its aggregate and per-tool figures exist. The text-only control of
\S\ref{sec:results:b6fix} has scripts but no logs in the released evidence
folder, so its figures ($0.20$ and $0.10$) rest on our records rather than on a
released log. And the two B8 pass counts for \texttt{combined\_v2}, 12 and 14 of
34 under the original and the re-implemented scorer, remain unreconciled; the
per-item outputs are released so that the difference can be traced.

\paragraph{On multi-seed reporting.}
The protocol above calls for $N{=}4$ seeds with mean$\pm$std where the
training run permits. In practice every phase is single-run: phase-2 alone
consumes $\approx$2\,weeks of dedicated 2$\times$A100-40GB on shared
infrastructure, and the vision phase adds $\approx$2\,h per sub-stage. At
this compute budget, a four-seed sweep of the full curriculum is not
economically feasible for a single-author, institution-unaffiliated project, so
we report single-seed numbers and label them as such throughout, compensating
partially with per-checkpoint trajectories (which expose seed-level variance
across the training path) rather than across-seed standard deviations; the
per-item paired bootstrap intervals the protocol calls for are not delivered in
this version either, for the reason given above. We treat multi-seed as a
\emph{desired} but not \emph{delivered} property, and flag it as a limitation
(Section~\ref{sec:limitations}).

\section{Results}
\label{sec:results}

\textbf{Status.} All four training phases have run to completion.
Phase~2 pretraining ran 192{,}000 steps ($\approx$50.3\,B tokens) on
2$\times$A100-40GB; phase~3 tool SFT ran to step~20{,}000 ($\approx$6\,B
tokens); phase~4 vision ran all three sub-stages (4a/4b/4c, $\approx$16\,M
tokens total, $\approx$2.2\,h wall time on 2$\times$A100-40GB, see
Section~\ref{sec:curriculum:vision}). Two families of vision runs appear below
and should not be conflated: that first full 4a/4b/4c pass, which supplies the
text-competence numbers of Table~\ref{tab:b1b5}, and the later single-GPU
phase-4a/4b runs on the instrumented extraction corpus, which supply the
grounding gate, the probe, the tiling result and the ablation
(\S\ref{sec:curriculum:asrun}).
Table~\ref{tab:pretrain_snapshot} reports measured intermediate pretraining
snapshots.
Table~\ref{tab:b1b5} reports the text-competence regression;
Tables~\ref{tab:gate} and~\ref{tab:probe} report the grounding gate and the
frozen-feature probe that together constitute this paper's main result.
Table~\ref{tab:qwenswap} reports the one experiment that breaks that
invariance---an encoder transplant run under the control's own recipe, which
recovers the field the gate had never recovered and relocates the ceiling from
the preprocessing to the encoder pathway (\S\ref{sec:results:qwenswap}).
Table~\ref{tab:nope_ablation} reports the nine-field gate for the three
NoPE$\times$vision variants, which were trained and evaluated
(\S\ref{sec:results:nope}), and Table~\ref{tab:layer_probe} a layer-wise
spatial probe run \emph{inside} V0 alone; that probe and a direct measurement
of the schedule-mismatch penalty in nats bound---without resolving---the
NoPE$\times$vision question (\S\ref{sec:results:layerprobe}).
Table~\ref{tab:b8} reports B8, a wider extraction benchmark whose ground truth
is supplied by the renderers themselves, measured on a later checkpoint trained
over sixteen templates; its principal outcome is not a single accuracy but a
three-way taxonomy of where grounding is confirmed and where it is not
(\S\ref{sec:results:b8}).
Tables~\ref{tab:b6ceiling}--\ref{tab:retention} explain why B6 sat at exactly
zero, report three projector-only fixes that lift it on its own render family,
and measure what the best of them forgets on B8 (\S\ref{sec:results:b6fix}).
Table~\ref{tab:abl} lists the released ablation
designs, whose discriminating statistics---including the one this ablation was
designed around---are not estimable while B6 is saturated at the floor for
every variant they compare (\S\ref{sec:limitations}). The designs, checkpoints, and runner are released
rather than deferred.

\subsection{Backbone pretraining snapshot (measured)}
\label{sec:results:snapshot}
Table~\ref{tab:pretrain_snapshot} reports B1--B5 measured on four intermediate
phase-2 checkpoints---spanning the exact BlkA/BlkB curriculum
boundary---each exported to GGUF (F16, 2.22\,GB) and served CPU-only
through Ollama---i.e., over the same deployment path the paper's efficiency
claim rests on, which these runs validate end-to-end for the NoPE architecture.
These are raw pretraining checkpoints \emph{without} phase-3 instruction/tool
SFT, evaluated single-seed at temperature 0.7. B1--B4 probe
instruction-following, classification, and tool-call behaviors that the
curriculum deliberately defers to phase~3, so scores at the pre-SFT floor are
the expected pattern, consistent with the base-vs-SFT gap documented for the
smaller VectraYX models~\cite{santillana2026vectrayx}. B5 (conversational
quality) is the one axis a base checkpoint can meaningfully exercise; across
the four checkpoints it holds a 0.62--0.69 band, and we read the
checkpoint-to-checkpoint movement as single-seed decoding variance rather than
a trend---one reason the reporting protocol of
Section~\ref{sec:eval} requires multi-seed means and bootstrap intervals for
any headline claim.

\begin{table*}[t]
\centering
\caption{B1--B5 on intermediate phase-2 \emph{base} checkpoints (no phase-3
SFT), GGUF F16 via Ollama, single seed. B3 = tool-match (exact-match = 0.0).
Training-progress snapshots, not headline results.}
\label{tab:pretrain_snapshot}
\small
\begin{tabular}{@{}lrccccc@{}}
\toprule
Step & Tokens & B1 & B2 & B3 & B4 & B5 \\
\midrule
70K          & 18.3B & 0.056 & 0.215 & 0.02 & 0.020 & 0.631 \\
74K          & 19.4B & 0.021 & 0.210 & 0.02 & 0.010 & 0.666 \\
76.8K (BlkA) & 20.1B & 0.022 & 0.215 & 0.00 & 0.005 & 0.621 \\
78K  (BlkB)  & 20.4B & 0.017 & 0.215 & 0.03 & 0.015 & 0.685 \\
\bottomrule
\end{tabular}
\end{table*}

\subsection{Text-competence regression (B1--B5)}
Table~\ref{tab:b1b5} compares the phase-3 backbone against the post-vision
checkpoints to quantify forgetting. Two phase-3 checkpoints are included:
step~8{,}000 (the prior B4 high-water mark during phase-3, reached with a corrected 5-bucket
mix) and step~20{,}000 (the final checkpoint used as the vision base). The
difference illustrates single-seed variance across the training trajectory;
step~20K was selected as the vision base because it had the best B5 of the two
checkpoints (0.650 vs 0.602)---the axis most worth preserving when injecting a
new visual modality---while B3 tool-match, though lower at 0.08 vs 0.17, falls
within the single-seed step-to-step band and is re-established later by
unfreezing \texttt{tok\_emb} (below).
Phase~4a (LLM frozen, projector only) shows no regression---confirming
that freezing the backbone preserves language competence exactly.
The v3A-p2 checkpoint (phase-4b final, \texttt{tok\_emb} unfrozen) achieves
B4=0.110 and B3 tool-match=0.17---the highest B4 score across the entire
training history, surpassing the phase-3 high-water mark of 0.075---while B5
holds at 0.602 and B2 falls from 0.190 to 0.110, the one axis that moves
against the comparison. The mechanism we suspect is that the embedding table,
which is tied to the output head, is required for the model to raise the
probability of the \texttt{<|tool\_call|>} token, so freezing it during vision
training suppresses tool use. We record that as a hypothesis consistent with the
comparison, not as an identified cause: v3B and v3A-p2 differ in learning rate
($5{\times}10^{-7}$ versus the phase default) and in step count as well as in
whether \texttt{tok\_emb} was frozen, both are single-seed, and no run isolates
the freeze. A controlled version of this comparison---one variable, matched
steps---is cheap and remains undone.

\begin{table}[t]
\centering
\caption{Text competence across the phase-3 training trajectory and after the
vision phase (single seed, temperature 0.7, GGUF F16 via Ollama). B3 reports
tool-match. v3A-p2 is the highest-scoring checkpoint on B4---0.110, exceeding
the phase-3 high-water mark of 0.075---and the lowest on B2.}
\label{tab:b1b5}
\footnotesize
\begin{tabular}{@{}p{0.44\columnwidth}ccccc@{}}
\toprule
Model & B1 & B2 & B3 & B4 & B5 \\
\midrule
Ph-3 step~8K (ph-3 B4 high)  & 0.306 & 0.260 & 0.17 & 0.075 & 0.602 \\
Ph-3 step~20K (vision base)  & 0.044 & 0.190 & 0.08 & 0.035 & 0.650 \\
\quad +vis 4a (frozen)        & 0.051 & 0.190 & 0.08 & 0.050 & 0.618 \\
\quad +vis 4b v3B (step~1900, LR=5e-7) & 0.056 & 0.200 & 0.04 & 0.030 & 0.631 \\
\quad +vis 4b v3A-p2 (final, unfreeze) & 0.053 & 0.110 & 0.17 & 0.110 & 0.602 \\
\bottomrule
\end{tabular}
\end{table}

\subsection{Repairing the vision fine-tuning path}
\label{sec:results:fixes}
Before any grounding number can be interpreted, we report five defects found
and fixed in the phase-4b path, because each is silent under normal logging and
each was a live candidate explanation for the results that follow. They are
described in full in \S\ref{sec:limitations}; in brief: bf16 master weights
combined with a small learning rate drove AdamW updates below the representable
increment (21\% of trainable tensors with a parameter delta of exactly $0.0$
after 300 steps); \texttt{wrap\_lora()} did not re-freeze the backbone after
\texttt{set\_phase()} unfroze it; \texttt{wrap\_lora()} ran \emph{before} the
resume checkpoint load in cross-phase resumes, so wrapped names
(\texttt{attn.wq.base.weight}) did not match flat checkpoint names
(\texttt{attn.wq.weight}) and, under \texttt{strict=False}, all 22 attention
layers were served at random initialization---confirmed by a measured weight
norm of 40.9668 against a theoretical random-init norm of 40.96; the same
defect in reverse affected same-phase resumes of already-wrapped checkpoints
(\texttt{missing=154} non-encoder keys), fixed by choosing the order
conditionally on the presence of \texttt{.base.weight} keys; and
\texttt{phase\_4b.lr}$=5\times10^{-6}$, inherited from a full-fine-tuning
recipe, was 20--60$\times$ too low for a rank-16 adapter, measured as a relative
adapter displacement
$\lVert\Delta W_{\mathrm{LoRA}}\rVert/\lVert W_{\mathrm{base}}\rVert$ of 0.13\%
mean and 0.23\% maximum over 88 attention projections after 600 steps. All
grounding results below are measured after all five fixes.

\subsection{The nine-field grounding gate is invariant across interventions}
\label{sec:results:gate}
To measure grounding without the confounds of open-ended generation, we use an
in-domain extraction gate: nine fields whose ground truth is a literal string
rendered in the image, each scored by format-normalized exact match against a
within-stratum shuffled-image control (the same template and question type, a
different image), so a model cannot win by discriminating templates. The
decision rule was frozen before training: a field passes iff
$\mathrm{acc}_{\text{real}}\geq0.60$ \emph{and}
$\mathrm{acc}_{\text{real}}-\mathrm{acc}_{\text{shuffled}}\geq0.40$.

The instrument is a purpose-built corpus, not the corpus of
Section~\ref{sec:dataset}: three renderers (a SIEM dashboard, a hex dump and a
log table) were instrumented to return the exact values they draw, giving 600
training and 100 held-out seeds per template and ten extractable fields. One
field, \texttt{logtable.flavor}, is kept in training as free signal but excluded
from the decision by design---at three categories it is passable without
reading anything---which is why the gate is nine fields and not ten.
Normalization is applied to both sides before matching (hexadecimal case and
leading zeros, ``2.4k'' versus ``2400''), so a formatting mismatch is not
scored as a grounding failure, and decoding is greedy, so the only variation
across the configurations below is the model. The evaluator, the renderers and
the held-out split are released with the rest of the harness
(\S\ref{sec:eval:artifacts}).
Table~\ref{tab:gate} reports the outcome across five training configurations,
all of them phase-4b runs on the V0 backbone; the same gate applied to the
three phase-4a backbone variants is reported separately in
\S\ref{sec:results:nope}.

Two of nine fields pass, and the result is invariant across every intervention
that leaves the encoder alone and adds no out-of-band text: the original learning rate, the corrected
learning rate, an added eight-domain transfer corpus, and an added
descriptive-prose corpus all yield exactly 2/9, and always the same two fields
(\texttt{siem.query}, \texttt{logtable.n\_error\_rows}). Across those four
configurations \texttt{siem.query} holds a 0.97--1.00 band with a $\Delta$ of
the same size, its shuffled control being $0.00$ throughout. The fifth
configuration, the tiling encoder of \S\ref{sec:results:tiling}, is the one
that moves the pass set, and it moves it downward, to 1/9. The seven failing
fields are not uniformly blind: several are close in character rather than
correct, which the harness records as a soft score alongside the binary
one---the mean character-level similarity (a \texttt{SequenceMatcher} ratio
between normalized prediction and ground truth) exceeds its own shuffled
control by $+0.45$ to $+0.57$ on \texttt{siem.alerts}, for example. None of the
seven clears exact match.

\paragraph{Why 2/9, and why it is not a resolution ceiling.}
The intuitive reading---384$\times$384 is too coarse to read the pane---is
contradicted by which field passes. \texttt{siem.query} carries the
\emph{most} symbolic variety in the corpus (1{,}120 admissible strings,
$\approx$9.68 bits of entropy) and is read almost perfectly, while
two-to-four-digit counts fail. The variable that separates \emph{these fields
from each other} is information per glyph after SigLIP's anisotropic resize: a
query string has a strong linguistic prior that lets partial glyph evidence be
completed, whereas an 8-digit hexadecimal address is near-uniform over its
alphabet and admits no completion. Because the gate scores \emph{complete}
exact match, its success probability is $p^n$ in the number of glyphs $n$, so a
modest per-glyph deficit is amplified exponentially---the quantitative form of
this argument, and its consequences for what the gate can and cannot ever
measure, is developed in \S\ref{sec:results:probe} and \S\ref{sec:limitations}.
This accounts for the ordering \emph{within} a fixed encoder, and it is not the
whole account of the ceiling's height: \S\ref{sec:results:qwenswap} replaces
the encoder and recovers the hexadecimal address at $0.81$ under a
\emph{coarser} token budget than the tiling condition that recovered nothing,
which no monotone function of pixels per glyph can produce.

\begin{table}[t]
\centering
\caption{Nine-field grounding gate across training configurations (single seed,
$n{=}40$ items per field, held-out split). A field passes iff
$\mathrm{acc}_{\text{real}}\geq0.60$ and $\Delta\geq0.40$ against a
within-stratum shuffled-image control. The pass set is identical
(\texttt{siem.query}, \texttt{logtable.n\_error\_rows}) in every configuration
except P1 tiling, which loses \texttt{logtable.n\_error\_rows}. \texttt{q} is
$\mathrm{acc}_{\text{real}}$ on \texttt{siem.query}.}
\label{tab:gate}
\footnotesize
\setlength{\tabcolsep}{4pt}
\begin{tabular}{@{}p{0.46\columnwidth}ccc@{}}
\toprule
Configuration & Fields passed & \texttt{q} & $\Delta_{\texttt{q}}$ \\
\midrule
4b, original LR ($5\times10^{-6}$)      & 2/9 & 0.97--1.00 & $+0.97$--$1.00$ \\
4b, corrected LR ($1\times10^{-4}$)     & 2/9 & 1.00 & $+1.00$ \\
\quad $+$ 8-domain transfer corpus      & 2/9 & 0.97--1.00 & $+0.97$--$1.00$ \\
\quad $+$ descriptive-prose corpus      & 2/9 & 0.97--1.00 & $+0.97$--$1.00$ \\
\quad $+$ P1 tiling (2$\times$2)        & 1/9 & 0.65 & $+0.65$ \\
\bottomrule
\end{tabular}
\end{table}

\subsection{P0: what survives to frozen SigLIP features}
\label{sec:results:probe}
To separate ``the information is absent from the encoder's representation''
from ``the adapter did not learn to use it,'' we fit a logistic-regression
probe on frozen SigLIP \texttt{last\_hidden\_state} features, mean-pooled over
the patch tokens covering the field's bounding box, under four preprocessing
conditions: (A)~the current pipeline, (B)~sharpening before resize, (C)~a
2$\times$2 tiling that gives each quadrant its own forward pass, and (D)~an
oracle crop-zoom on the field's bounding box, which upper-bounds what a
\emph{single} 384$\times$384 forward pass could recover given perfect
localization. Condition~C is not bounded by~D---it spends four forward passes
rather than one---and in fact exceeds it on \texttt{addr}. We probe two
independent fields with 750 fresh samples each and score \emph{per glyph}, not
per field: \texttt{hexdump.base\_addr} (8 nibbles, chance 6.25\%) and
\texttt{logtable.target\_row\_pid} (5 digits, chance 10\%).

Table~\ref{tab:probe} reports the result. All conditions are far above chance,
so the information is present; tiling (C) produces a large and consistent gain
on both fields (60.8\%~$\to$~86.0\%, 62.4\%~$\to$~79.2\%). The right-hand
columns propagate per-glyph accuracy to the quantity the gate actually scores,
$p^n$. Under condition A the projection for \texttt{base\_addr} is 1.9\%,
matching the observed gate accuracy of $0.00$; the \emph{best} of the four
conditions is tiling (C) at 29.9\%, and the oracle crop-zoom (D) reaches 23.0\%.
No condition we measured brings that field within a factor of two of the gate's
0.60 threshold, so it is \emph{structurally} unreachable for this gate under
every preprocessing route we tested, independent of training. The 5-digit field
is not: it projects to 64.5\% under D, which clears the threshold. The gate, as
designed, conflates the two cases.

\begin{table}[t]
\centering
\caption{P0 probe: per-glyph linear recoverability from frozen SigLIP features
($n{=}750$ per field per condition). Right columns propagate per-glyph accuracy
$p$ to complete exact-match, $p^n$, the quantity the gate scores
($n{=}8$ nibbles, $n{=}5$ digits). Chance is 6.25\% and 10\% per glyph.}
\label{tab:probe}
\footnotesize
\setlength{\tabcolsep}{4pt}
\begin{tabular}{@{}p{0.30\columnwidth}cccc@{}}
\toprule
 & \multicolumn{2}{c}{per-glyph} & \multicolumn{2}{c}{projected $p^n$} \\
Condition & \texttt{addr} & \texttt{pid} & \texttt{addr} & \texttt{pid} \\
\midrule
A: current pipeline   & 0.608 & 0.624 & 0.019 & 0.095 \\
B: $+$ sharpen        & 0.697 & 0.619 & 0.056 & 0.091 \\
C: 2$\times$2 tiling  & \textbf{0.860} & 0.792 & 0.299 & 0.312 \\
D: oracle crop-zoom   & 0.832 & \textbf{0.916} & 0.230 & 0.645 \\
\bottomrule
\end{tabular}
\end{table}

\subsection{P1: the probe's prediction does not transfer end-to-end}
\label{sec:results:tiling}
Condition C motivated a full implementation: 2$\times$2 tiling with
pixel-unshuffle patch merging, trained end-to-end under the same corrected
recipe. The outcome is negative on every axis. The gate falls from 2/9 to 1/9,
losing \texttt{logtable.n\_error\_rows}---a field that had passed in every prior
configuration---while \texttt{siem.query} survives but weakens markedly
($\mathrm{acc}_{\text{real}}=0.65$ against 0.97--1.00). Critically, the two
fields the probe predicted would improve, \texttt{hexdump.base\_addr} and
\texttt{logtable.target\_row\_pid}, remain at exactly $0.00$: not improved, not
degraded, unchanged. The model's output behavior changes sharply in the same
run: its abstention rate on B6 falls from 0.74--0.82 to 0.14 and its
hallucination rate rises from 0.18--0.26 to 0.86---it stops answering ``that
does not appear in this image'' and begins emitting fluent, unfounded file paths
and function names. We report these as descriptive statistics of what the model
emits, not as a calibration result, because B6 contains no unanswerable item
against which abstention could be scored as correct (\S\ref{sec:results:b6}).
The degradation is not an artifact of stopping at an unlucky checkpoint: the
step-800 checkpoint, at 71\% of training, already shows the same pattern.

Three mechanisms are candidates. The first two are properties of the
implementation, confirmed by inspection rather than by an experiment that
isolates them. No SigLIP forward pass ever sees the complete scene, so any task
requiring global aggregation---counting rows, or reading a string that crosses
a tile seam---has had its evidence partitioned away; production AnyRes
pipelines~\cite{liu2024llavanext,wang2024qwen2vl} add a downscaled global
thumbnail alongside the tiles for precisely this reason, and our implementation
did not. And the run used \texttt{use\_2d\_pos\_embed=False}, so SigLIP's
positional embeddings were replicated across all four tiles with nothing to
disambiguate them. The third is a confound rather than a mechanism, and it
weakens the comparison rather than explaining the failure: widening the
projector's input from 1152 to 4608 makes the non-tiled projector
shape-incompatible, so the tiled run re-initialized it and re-aligned it in a
single 562-step epoch, against the several rounds behind the non-tiled one
(\S\ref{sec:curriculum:asrun}). The tiled model is therefore not a
one-variable change from its comparison, and part of the drop may be a
projector that is simply less well aligned. That does not rescue the probe's
prediction---the two fields it favored are unchanged at exactly $0.00$, which
an under-aligned projector does not by itself explain---but it does mean the
size of the regression should not be read as the cost of tiling. The
methodological conclusion is stated as a limitation in
\S\ref{sec:limitations}: linear recoverability on frozen features measures an
optimal extractor under oracle supervision, and is a necessary but insufficient
predictor of what a 4.7--12.7\,M-parameter adapter will learn to exploit in
600--1{,}200 steps. One quantity from this run is load-bearing later and we
record it here: tiling spends 2{,}916 visual tokens on the scene, or 270 source
pixels per token, against 729 tokens and 1{,}079 source pixels per token for
the baseline---four times the token budget and twice the linear scale
(\S\ref{sec:results:qwenswap}).

\subsection{P2: transplanting a natively-trained visual tower}
\label{sec:results:qwenswap}
Every configuration above holds the encoder fixed, and \S\ref{sec:results:probe}
is explicit that its bound is stated for SigLIP-so400m at its native
384$\times$384 input and not for encoders in general. This subsection removes
that restriction by replacing the tower and changing nothing else we could
hold constant: we transplant the visual tower of
Qwen2-VL-2B~\cite{wang2024qwen2vl}---which preprocesses at variable resolution
with the aspect ratio preserved, rather than resizing every image to a fixed
square---onto the same decoder, and re-run the same alignment recipe that
produced the checkpoint behind all the numbers above.

\paragraph{Pre-registration, and the prediction it came from.}
The two primary fields were fixed in a \texttt{prereg.json} written to disk
before step 0, and they were not chosen for being promising. They were chosen
because they are the two fields that a resolution-degradation ablation on an
\emph{external} model had already singled out. Running Qwen2-VL-7B zero-shot on
this gate ($n{=}40$, same held-out images, same normalizer) passes 9/9; running
it again on the same images downsampled to 384$\times$384 bicubic \emph{before}
its own processor---forcing the fixed-square regime SigLIP operates in---drops
it to 7/9, and the only two fields that collapse are
\texttt{hexdump.base\_addr} ($0.93\to0.03$) and \texttt{siem.query}
($0.68\to0.12$). Those are exactly the two fields the information-per-glyph
account of \S\ref{sec:results:gate} nominates as most vulnerable: the field
with the most nearly uniform per-character alphabet, and the longest string.
The transplant was therefore run as a directional test of a prediction made
before the run, on the two fields named before the run, and not as a search
over ten fields for something that moved.

\paragraph{What was held fixed, and what the swap changes.}
The decoder is the identical frozen phase-3 backbone used by the control
checkpoint (verified at load: 0 missing, 0 unexpected keys, plus a
\texttt{tok\_emb} norm fingerprint asserted before training to rule out a
repeat of the random-initialization defect of \S\ref{sec:results:fixes}). The
corpus is \texttt{pilot\_corpus\_v2} unchanged---6{,}000 training and 1{,}000
held-out records over the same three templates. The recipe is the same 1{,}496
steps, 8 epochs, LR $1{\times}10^{-3}\!\to\!1{\times}10^{-4}$, effective batch
32. The projector keeps its shape and its two-layer $4096$-hidden form; only its
input width changes, $1{,}152\to1{,}536$. Three things change together with the
tower, and this design separates none of them: the tower's own pretraining
(image--text co-training at native resolution rather than at a fixed square),
the aspect-preserving variable-resolution preprocessing that comes with it, and
the visual token budget (999, 962 and 1{,}196 tokens for the hexdump, log-table
and SIEM templates against a uniform 729). The wider input also adds 1.6\,M
projector parameters, a 12\% increase on 13.1\,M. The claim below is therefore
about the encoder \emph{pathway} as a unit under a fixed alignment budget, not
about any one of its three properties. The run itself was preempted by the
provider at step 1{,}000 and resumed from checkpoint with optimizer state; the
resumed schedule continued the cosine decay from that step rather than
restarting it.

\paragraph{The metric, and why it is not the gate.}
At this alignment budget the gate's exact-match rule is not the right
instrument for a comparison: it is a threshold on a quantity that is
$p^n$ in glyph count, so two arms can differ substantially in what they
recover and still both read $0.00$. We therefore score each held-out item by
$\Delta\text{-NLL}$: the negative log-likelihood of the ground-truth answer
tokens given a within-stratum shuffled image, minus the same quantity given the
real image, averaged over items, with a percentile bootstrap interval over
items. It uses the shuffled-image control of \S\ref{sec:results:gate}
unchanged, is continuous rather than thresholded, and---being a likelihood, not
a decode---is unaffected by a decoding failure that would zero an exact-match
score. We report greedy exact match alongside it, computed with the gate's own
normalizer, so the two columns are directly comparable to
Table~\ref{tab:gate}. Before the transplant was evaluated, the same harness was
run at $n{=}100$ on the SigLIP control checkpoint, and its exact-match column
reproduces that checkpoint's historical $n{=}40$ gate accuracies on all nine
gate fields to within sampling noise (e.g.\ $1.00\to0.99$, $0.95\to0.94$,
$0.40\to0.42$, $0.00\to0.00$), which makes the control column a reproduction of
the whole pipeline---loader, corpus, held-out split, normalizer, decode---and
not merely a reference scale.

Table~\ref{tab:qwenswap} reports both arms over all ten instrumented fields.

\paragraph{The primary result.}
\texttt{hexdump.base\_addr} is the field SigLIP never read. It is $0.00$ under
every configuration in this paper---the original and corrected learning rates,
the transfer and prose corpora, tiling, and all three NoPE variants
(Tables~\ref{tab:gate},~\ref{tab:nope_ablation})---and
\S\ref{sec:results:probe} projects $1.9\%$ for it from frozen SigLIP features,
rising to only $29.9\%$ under the best of four preprocessing conditions. Under
the transplanted tower, at the same alignment budget, it reads $0.81$ exact
with $\Delta\text{-NLL}=+5.47$, CI $[+5.24,+5.71]$.

The per-digit figure needs stating carefully, because the obvious way to report
it is inflated. The renderer draws the address as
\texttt{0x00400000}${}+{}$\texttt{randint(0,0xFFFF)}${}\times{}$\texttt{0x10}
(\texttt{render\_v2.py}), so of the eight hexadecimal digits displayed, four are
constant by construction across the entire corpus---the \texttt{004} prefix and
the trailing \texttt{0}---and we confirmed on the held-out slice that the four
remaining positions each take all 16 values. The informative alphabet is 16
bits, not 32. Scored on those four digits only, accuracy is $372/400=93.0\%$
against a $6.25\%$ chance rate; the $96.5\%$ that a naive count over all eight
digits produces is not a meaningful number and we do not report it. The 28
digit errors are concentrated in 19 items rather than spread: independent
per-digit errors at $93.0\%$ would predict $0.93^{4}=0.75$ exact match against
the $0.81$ observed.

\paragraph{The second pre-registered field is contaminated, and we demote it.}
\texttt{siem.query} reads $0.99$ exact against the control's $0.94$, but we do
not present it as a clean primary. Auditing the split after the fact, 63 of the
100 held-out query strings appear verbatim among the training answers for that
field: the renderer draws from 398 distinct queries and the 6{,}000-record
training corpus reuses them, so held-out \emph{images} do not imply held-out
\emph{answers}. We report the field as secondary with the contamination
stated. The same audit is what licenses the primary: on
\texttt{hexdump.base\_addr}, 98 of 100 held-out values never appear in
training, and the two that do are coincidences of a 16-bit draw over 600
training items. \texttt{logtable.target\_row\_pid} is likewise 98/100 unseen.
Every other field draws from an alphabet small enough that all 100 held-out
values appear in training by construction, which is a property of the corpus
and not a defect of the split, but it does mean \texttt{base\_addr} and
\texttt{target\_row\_pid} are the only two fields on which this instrument can
distinguish reading an image from recalling a distribution. One further detail
is consistent with the contamination reading: \texttt{siem.query} is the one
field where the transplant's $\Delta\text{-NLL}$ is \emph{lower} than the
control's ($+1.95$ against $+3.10$) while its exact match is higher. Since
$\Delta\text{-NLL}$ is driven by how badly a shuffled image hurts, a model that
has learned the query distribution well is penalized less by the wrong image,
which is what a memorized answer set would predict. We offer that as the most
economical reading, not as a measurement.

\paragraph{What did not improve, which is most of the table.}
The pattern is not ``the new tower wins everywhere,'' and the exceptions are
informative. The three \texttt{logtable.target\_row\_*} fields do not improve:
$0.25$, $0.00$ and $0.22$ against the control's $0.38$, $0.00$ and $0.31$, two
of them worse. All three ask for a value in \emph{row number 7}, so answering
requires an ordinal indexing step before any glyph is read, and all three are
weak under both towers---including \texttt{target\_row\_pid}, which
\S\ref{sec:results:probe} identified as recoverable in principle ($64.5\%$
projected under the oracle crop-zoom) and which is one of only two uncontaminated
fields. \texttt{hexdump.base\_addr}, by contrast, is specified by a fixed named
location (``the window footer''). The descriptive reading is that the swap buys
glyph transcription at a named position and does not buy ordinal row selection;
we note that the three SIEM tile fields do not follow the same pattern
(\texttt{events} $0.14\to0.95$, \texttt{alerts} $0.42\to0.17$, \texttt{hosts}
$0.00\to0.03$, all three read off adjacent labelled tiles in one panel), so we
report the pattern as a description of the table and not as a mechanism.
\texttt{siem.alerts} in particular is more than twice as good under the encoder
this paper says is the bottleneck.

\paragraph{Multiplicity, and which rows are claims.}
Ten fields were scored without correction. Two were pre-registered as primary,
two as internal-validity controls with a stated threshold
($\Delta\text{-NLL}>+3$ nats on \texttt{logtable.flavor} and
\texttt{logtable.n\_error\_rows}, or the run is discarded; both clear it with
margin, so the arm trained), and one, \texttt{siem.hosts}, was declared in
advance as a field expected to stay at the floor. The remaining rows are
context. In particular \texttt{logtable.target\_row\_host} has a
$\Delta\text{-NLL}$ of $+0.04$ whose interval technically excludes zero
($[+0.01,+0.08]$); at 1/100th the scale of the primary effect, on the tenth of
ten uncorrected tests, that is not a finding and we do not read it as one.
\texttt{siem.hosts} is best described as a known floor field rather than a
negative control in the strong sense: the information is present in the image,
no encoder we have tested reads it, and its staying at $0.03$ shows only that
the transplant did not inflate every field uniformly.

\paragraph{The comparison is biased against the arm that wins.}
The frozen decoder both arms use was itself warm-started from a checkpoint
whose vision alignment was performed against \emph{SigLIP} features---it is the
control arm's own decoder. The transplanted tower is asked to project into an
embedding space that was shaped by the encoder it replaces, with no
compensating adaptation, while the control arm is evaluated in the space it
helped create. Whatever this is worth, its sign is against the transplant, and
the transplant wins the primary field regardless. We state it because it is the
one confound in this comparison that runs in the conservative direction; the
three named in ``what the swap changes'' above do not.

\paragraph{What this reframes.}
\S\ref{sec:results:gate} attributes the ceiling to information per glyph after
SigLIP's anisotropic resize. That account survives---it predicted, correctly and
in advance, which two fields an external model would lose under a forced resize
---but it is no longer sufficient on its own, and our own tiling result is what
refutes the sufficient version. The arithmetic is exact and follows from the
preprocessing alone. On a 1024$\times$768 hexdump pane, the baseline pipeline
resizes to 384$\times$384 (horizontal scale $0.375$, vertical $0.500$) and
spends 729 visual tokens on the scene, or 1{,}079 source pixels per token. The
2$\times$2 tiling of \S\ref{sec:results:tiling} resizes each 512$\times$384
quadrant to 384$\times$384 (horizontal $0.750$, vertical $1.000$) and spends
2{,}916 tokens, or 270 source pixels per token---twice the linear scale and four
times the token budget of the baseline, and the probe measured the predicted
per-glyph gain, $0.608\to0.860$. The transplanted tower resizes the same pane
to 1036$\times$756 (horizontal $1.012$, vertical $0.984$) and spends 999
tokens, or 787 source pixels per token: relative to tiling it is
\emph{coarser} in tokens by $2.9\times$, and finer in linear scale by only
$1.35\times$ horizontally. Tiling reads the address $0.00$ of the time; the
transplant reads it $0.81$. A monotone account in pixels-per-glyph, or in
tokens-per-scene, cannot produce that ordering. What separates the two
conditions is not how much of the glyph reaches the encoder but whether the
encoder was trained to turn it into a transcribable feature---the tower is
co-trained on image--text data at its operating resolution rather than adapted
to a resize it never saw. We therefore restate the ceiling as a property of the
encoder pathway rather than of the preprocessing alone: information per glyph
after resize sets what is available, and the encoder's pretraining regime sets
how much of what is available survives into features an adapter can use. Both
terms are necessary; this paper's earlier configurations varied only the
second-order ones.

\paragraph{Scope.}
This is one run, one seed, one synthetic generator, one font, one palette and
three layouts, with no out-of-distribution imagery and no seed ablation, scored
on a corpus whose held-out split shares its answer alphabet with training on
eight of ten fields. It licenses a comparative statement---that under this
fixed alignment budget the encoder pathway, not the adapter or the training
volume, is what separates $0.00$ from $0.81$ on the field this paper has called
structurally unreachable---and it licenses no absolute statement about reading
hexdumps. It also does not license a change of system: the transplant as run,
Qwen's merger followed by a second MLP, has no working \texttt{llama.cpp}
export path for this backbone (\S\ref{sec:discussion:encoder}), and the
export-shaped variant of \S\ref{sec:arch:qwenline} does not reproduce the
result, so the finding identifies the bottleneck without yet relieving it. The checkpoint, the pre-registration file, the
caching and training scripts and this evaluation harness are released with the
rest (\S\ref{sec:eval:artifacts}).

\begin{table*}[t]
\centering
\caption{P2 encoder transplant: the Qwen2-VL-2B visual tower on the same frozen
decoder, same projector shape, same corpus and same 1{,}496-step recipe as the
SigLIP control, over all ten instrumented fields ($n{=}100$ held out, greedy
decode, gate normalizer). $\Delta$ is $\Delta\text{-NLL}$ in nats against a
within-stratum shuffled-image control (higher is more grounded); \texttt{ex} is
exact match, comparable to $\mathrm{acc}_{\text{real}}$ in
Table~\ref{tab:gate}. Role: \textbf{P} pre-registered primary, \textbf{V}
internal-validity control (threshold $\Delta>+3$, both clear it), \textbf{F}
field declared in advance to sit at the floor, \textbf{d} descriptive, not an
inferential claim. Ten fields, no multiplicity correction: only the P and V
rows are claims. \texttt{siem.query} is marked \dag: 63/100 of its held-out
answers occur verbatim in training, so it is reported as secondary.
\texttt{hexdump.base\_addr} and \texttt{logtable.target\_row\_pid} are the only
fields whose held-out values are largely unseen (98/100 each).}
\label{tab:qwenswap}
\footnotesize
\setlength{\tabcolsep}{5pt}
\begin{tabular}{@{}llrcrrr@{}}
\toprule
 & & \multicolumn{3}{c}{Qwen2-VL tower} & \multicolumn{2}{c}{SigLIP control} \\
\cmidrule(lr){3-5}\cmidrule(l){6-7}
Field & Role & $\Delta$ & 95\% CI & \texttt{ex} & $\Delta$ & \texttt{ex} \\
\midrule
\texttt{hexdump.base\_addr}        & \textbf{P}  & $+5.47$ & $[+5.24,+5.71]$ & \textbf{0.81} & $+0.00$ & 0.00 \\
\texttt{siem.query}\dag            & \textbf{P}  & $+1.95$ & $[+1.84,+2.07]$ & \textbf{0.99} & $+3.10$ & 0.94 \\
\midrule
\texttt{logtable.flavor}           & \textbf{V}  & $+4.79$ & $[+4.00,+5.62]$ & 1.00 & $+6.39$ & 1.00 \\
\texttt{logtable.n\_error\_rows}   & \textbf{V}  & $+5.46$ & $[+4.36,+6.55]$ & 0.91 & $+9.75$ & 0.99 \\
\midrule
\texttt{siem.hosts}                & \textbf{F}  & $+0.15$ & $[-0.03,+0.31]$ & 0.03 & $+0.09$ & 0.00 \\
\midrule
\texttt{siem.events}               & d & $+4.41$ & $[+4.07,+4.77]$ & 0.95 & $+1.03$ & 0.14 \\
\texttt{siem.alerts}               & d & $+1.47$ & $[+1.16,+1.79]$ & 0.17 & $+3.33$ & 0.42 \\
\texttt{logtable.target\_row\_proc}& d & $+0.33$ & $[+0.13,+0.55]$ & 0.22 & $+0.65$ & 0.31 \\
\texttt{logtable.target\_row\_host}& d & $+0.04$ & $[+0.01,+0.08]$ & 0.25 & $+0.09$ & 0.38 \\
\texttt{logtable.target\_row\_pid} & d & $+0.02$ & $[-0.01,+0.05]$ & 0.00 & $+0.02$ & 0.00 \\
\bottomrule
\end{tabular}
\end{table*}

\subsection{Vision benchmarks (B6, B7) and three harness defects}
\label{sec:results:b6}
\label{sec:results:harness}
Auditing B6 while the model scored zero surfaced three defects in our own
harness, all of which inflate or distort scores independently of the model. A
fourth, which only a model scoring above zero could expose, is described with
the results that exposed it (\S\ref{sec:results:b6fix}).

First, \texttt{\_build\_full\_suite()} computed a \texttt{variant} index and
never applied it, so five ``extra'' templates were emitted nine times each,
byte-identically. The nominal $n{=}50$ suite was an effective $n{=}10$, and the
per-item bootstrap intervals prescribed in \S\ref{sec:eval} were computed over
duplicated rows treated as independent draws. The corrected generator emits 45
genuinely distinct variants (verified: 50/50 unique \texttt{screen\_text}).
Second, the synthetic renderer printed the depicted product's name (``IDA Pro
8.3'', ``Splunk Enterprise'') in the window title bar, so tool identification
partly measured optical character recognition of a caption rather than
knowledge of the tool; the corrected renderer uses generic per-category labels.
Third, \texttt{is\_wrong\_tool\_named} and \texttt{score\_tool\_id} matched
aliases by plain substring, and Spanish is dense in collisions: ``ida'' occurs
inside \emph{salida}, \emph{seguridad} and \emph{vulnerabilidades}, ``siem''
inside \emph{siempre}. In a real GPT-4o run, 23 of 50 correct responses were
scored as wrong-tool hallucinations by this bug alone. The bias had a
direction that matters more than its magnitude: our own model's terse
abstentions cannot collide with a Spanish substring, while the fluent Spanish
of the baselines does, so the defect systematically favored the proposed
system over its comparisons. We now match on word boundaries and separate
product names from generic domain vocabulary.

Under the corrected harness, tool identification is $0.0$ on every checkpoint
we measured that was not trained on B6's own task format, absent an
out-of-band text hint (the exceptions are the three projectors of
\S\ref{sec:results:b6fix}, which were trained on it)---including
\texttt{domain\_expansion\_v1} (\S\ref{sec:results:b8}), re-measured after that
checkpoint's broader training and unchanged---and keyword recall and answer
correctness are $0.0$ as well on the three phase-4b corpus variants of
Table~\ref{tab:gate} (base, $+$transfer, $+$prose). The transplanted-tower stack of
\S\ref{sec:results:qwenswap} is no different---tool identification and answer
correctness both $0.0$, keyword recall $0.003$---so the field it recovers at
$0.81$ buys nothing here. One checkpoint is not an exact zero on all three, and
because it is the one we release we state it rather than round it away. The
public GGUF artifact
(\href{https://huggingface.co/jsantillana/vectrayx-vision-1b}{\texttt{jsantillana/vectrayx-vision-1b}})
is the SFT-v2 continuation at step 400, warm-started from the v3B phase-4b run
and therefore a different lineage from the Table~\ref{tab:gate} configurations.
Through the corrected suite it scores tool identification $0.0$ and
\texttt{think\_present} $0.0$, as everything else in this line does, but keyword
recall $0.010$ and answer correctness $0.020$---one item in fifty---served
through \texttt{llama.cpp}/Ollama, and keyword recall $0.017$ and answer
correctness $0.040$---two items---when the same weights are evaluated directly
in PyTorch, with chain quality $0.020$ under both transports. We draw no
capability claim from that: at $n{=}50$ one item \emph{is} $0.02$, the two
transports of a single checkpoint disagree by exactly that much, and no item is
answered with the tool correctly named. We argue in \S\ref{sec:limitations} that this says
nothing about grounding, because no checkpoint in this line has had a
conversational SFT stage and B6 asks for free-form identification and
explanation; we therefore report no B6 number for these checkpoints in any
results table, since the quantity being measured is the absence of a training
stage. \S\ref{sec:results:b6fix} tests that reading directly, by supplying the
missing supervision to the projector alone. We additionally computed, on the $+$prose configuration,
accuracy-on-answered, restricted to the 13 items on which that checkpoint did not
abstain, to test whether a selective-prediction framing recovers a positive
reading; it is also $0.0$. We therefore retract the 0.08 tool-identification
figure and the 0.08-vs-0.02 ``vision signal'' argument reported in an earlier
draft, both of which were artifacts of the three defects above
(\S\ref{sec:limitations}).

We did not rerun frontier baselines against the corrected suite. The one
external model we do report on B6 is stock Qwen2-VL-2B, run as a like-for-like
perception ceiling at our own pixel budget rather than as a peer
(\S\ref{sec:results:b6fix}); the earlier GPT-4o run used the pre-fix harness
and is not comparable to any number here. The text-only control of
\S\ref{sec:eval:baselines} was run in one form only: the frozen backbone given
each item's on-screen text as plain text instead of the image. It scores tool
identification $0.20$ without the system prompt and $0.10$ with it, and every
one of the ten hits is an item whose screen text contains an accepted alias
(10/21, against 0/29 on the rest), so the backbone contributes name copying and
no tool knowledge of its own (\S\ref{sec:results:b6fix}). The question-only
form of the control remains unmeasured. We also note
that B6 contains no genuinely out-of-distribution or unanswerable item---all 50
are in-distribution security questions with a known ground-truth fact---so it
is not a valid instrument for measuring abstention or calibration, and we make
no calibration claim from it.

\subsection{B8: a ground-truth-by-construction gate over sixteen templates}
\label{sec:results:b8}
The nine-field gate of \S\ref{sec:results:gate} covers three templates, and its
decision rule compares against a shuffled-image control alone. B8 widens the
first and strengthens the second. It is 2{,}040 items over 34 fields spanning 16
rendered-UI templates---SIEM dashboards, log tables, hex dumps, register panels,
call stacks, MITRE matrices, CVE cards, phishing emails, gdb and WinDbg
sessions, IDA disassembly, Ghidra decompiler output, CPU pipeline diagrams,
Wireshark packet captures, ANSI terminal sessions and crypto primitive
diagrams---at $n{=}60$ items per field, single seed. B8 names the three pilot
templates in full (\texttt{siem\_dashboard}, \texttt{log\_table},
\texttt{hex\_dump}); they are the same renderers behind the \texttt{siem},
\texttt{logtable} and \texttt{hexdump} fields of Table~\ref{tab:gate}.

Two properties separate it from B6/B7 (\S\ref{sec:eval:b6},
\S\ref{sec:eval:b7}). First, every item's ground truth is the renderer's own
return value, not an answer written about the image afterwards. This is not a
hygiene preference: an earlier corpus in this project was generated by renderers
that ignored their \texttt{record} argument entirely, so the recorded answer and
the drawn pixels could disagree with nothing in the pipeline able to detect it
(the defect was found and fixed separately). A benchmark whose labels
\emph{are} the renderer's return value cannot carry that failure by
construction. Second, each field is scored against two controls rather than
one: the within-stratum shuffled image of \S\ref{sec:results:gate}, and a
best-constant baseline---the accuracy obtained by always emitting the most
frequent ground-truth value in that field's sample. The two answer different
questions. The shuffled control asks whether the answer depends on \emph{this}
image; the best-constant baseline asks whether it needs to depend on any image
at all. A field with a skewed answer distribution can look grounded on a naive
real-versus-shuffled delta and still be explained by the modal answer, so B8
reports $\mathrm{acc}_{\text{real}}$, $\mathrm{acc}_{\text{shuffled}}$ and
$\mathrm{acc}_{\text{best-constant}}$ and scores each field against
$\max(\mathrm{acc}_{\text{shuffled}},\mathrm{acc}_{\text{best-constant}})$. The
decision rule was frozen before the run and is the gate's rule taken against
that maximum: a field passes iff $\mathrm{acc}_{\text{real}}\geq0.60$ and
$\Delta\geq0.40$.

\paragraph{The checkpoint under test.}
B8 was run on \texttt{domain\_expansion\_v1}, trained from scratch rather than
resumed, on a combined corpus of 8{,}802 training examples: the 6{,}000-example
three-template pilot corpus of \S\ref{sec:results:gate}
(\texttt{siem\_dashboard}, \texttt{log\_table}, \texttt{hex\_dump}) plus 2{,}802
examples over six newly instrumented templates (\texttt{ida\_disasm},
\texttt{ghidra\_decomp}, \texttt{arch\_diagram}, \texttt{wireshark\_packets},
\texttt{terminal\_ansi}, \texttt{crypto\_diagram}), for 2{,}200 steps, with the
Qwen-tower pathway and export-shaped projector of \S\ref{sec:arch:qwenline}
(first layer initialized from Qwen's merger, decoder frozen). Final held-out
loss is 0.1358 against a 6.71 baseline. Seven of B8's 16 templates therefore
have no training exposure in this checkpoint, or in any checkpoint preceding
it, and we report them below rather than restricting the benchmark to what was
trained.

\paragraph{What passes.}
Nine of the 34 fields clear the pre-registered gate; Table~\ref{tab:b8} lists
them. \texttt{log\_table.flavor} is one of the nine: the gate of
\S\ref{sec:results:gate} excludes it from its own decision because, at three
categories, it is passable without reading anything, but that concern is
exactly what $\mathrm{acc}_{\text{best-constant}}$ is built to catch, and here
it does not explain the result---the field's best-constant baseline is 0.43 in
this sample, and $\mathrm{acc}_{\text{real}}=0.93$ still clears it by
$\Delta=+0.50$. For these fields the reading is that grounding is
\emph{confirmed}---the model's answer tracks the image it was shown, and neither
control accounts for it---and the scope of that statement is the field, not the
template and not the model: each row states that this question, asked of this
template, is answered from the image at the stated rate.

\begin{table}[t]
\centering
\caption{B8 fields clearing the pre-registered gate on
\texttt{domain\_expansion\_v1} ($n{=}60$ per field, single seed, 2{,}040 items
over 34 fields). $\Delta$ is $\mathrm{acc}_{\text{real}}$ minus
$\max(\mathrm{acc}_{\text{shuffled}},\mathrm{acc}_{\text{best-constant}})$; a
field passes iff $\mathrm{acc}_{\text{real}}\geq0.60$ and $\Delta\geq0.40$.
Nine fields pass, including \texttt{log\_table.flavor}
($\mathrm{acc}_{\text{best-constant}}=0.43$ in this sample, cleared by
$\Delta=+0.50$)---see \S\ref{sec:results:b8} for why that field's exclusion
from the gate of Table~\ref{tab:gate} does not carry over here. The other 25
fields do not pass and are described in the text.}
\label{tab:b8}
\footnotesize
\setlength{\tabcolsep}{6pt}
\begin{tabular}{@{}lrr@{}}
\toprule
Field & $\mathrm{acc}_{\text{real}}$ & $\Delta$ \\
\midrule
\texttt{siem\_dashboard.hosts}   & 1.00 & $+0.93$ \\
\texttt{siem\_dashboard.alerts}  & 1.00 & $+0.88$ \\
\texttt{siem\_dashboard.events}  & 0.98 & $+0.88$ \\
\texttt{ida\_disasm.func\_name}  & 1.00 & $+0.83$ \\
\texttt{ghidra\_decomp.func\_name} & 1.00 & $+0.78$ \\
\texttt{siem\_dashboard.query}   & 0.82 & $+0.78$ \\
\texttt{terminal\_ansi.script\_key} & 1.00 & $+0.77$ \\
\texttt{crypto\_diagram.flavor}  & 1.00 & $+0.67$ \\
\texttt{log\_table.flavor}       & 0.93 & $+0.50$ \\
\bottomrule
\end{tabular}
\end{table}

\paragraph{What does not, and why it is three separate things.}
The 25 remaining fields do not form one population, and the partition is the
result we regard as load-bearing---more so than any individual accuracy in
Table~\ref{tab:b8}.

\emph{(a) Templates with no training exposure.} The seven templates absent from
every training corpus for this checkpoint---\texttt{register\_panel},
\texttt{call\_stack}, \texttt{mitre\_matrix}, \texttt{cve\_card},
\texttt{phishing\_email}, \texttt{gdb\_session}, \texttt{windbg\_session}---sit
at approximately $0.00$ on nearly every field, and their predictions are not
near-misses. Asked to name a CPU register, the model returned a memory address
that appears nowhere in the image; asked for a MITRE tactic, an unrelated
protocol name and a number. We read this as an absence of training exposure and
not as a measured capability ceiling: nothing in this run tests what these
templates would yield after instrumentation, and we make no prediction about
that.

\emph{(b) A trained template with an untrained field inside it.}
\texttt{log\_table.n\_rows} scores $0.00$---below the $0.35$ its own
best-constant baseline would obtain---in the same template, on the same
checkpoint, where \texttt{log\_table.flavor} scores $0.93$ with both controls
cleared. The template is familiar and the question is not, and the model does
not fall back to reading the image: it emits a near-constant wrong answer
independent of what is drawn. This is the controlled counterpart of a pattern
we had previously seen only anecdotally on B6, and we treat it as the
mechanistic finding of this run rather than as one more failing field; the
argument is developed in \S\ref{sec:discussion:landmark}.

\emph{(c) Ceilings this paper has already characterized in part.} Two groups
remain, both consistent with limits described in \S\ref{sec:results:gate} and
\S\ref{sec:results:probe}. Fine open-alphabet symbol reading is the first:
memory addresses in \texttt{arch\_diagram} and \texttt{ghidra\_decomp} are flat
at the floor, while the base-address fields of \texttt{hex\_dump} and
\texttt{ida\_disasm} show weak but positive signal ($\Delta=+0.12$ and $+0.07$).
Those two deltas are far below the gate and we claim nothing from their
magnitude; we record them because they are positive where the corresponding
field was exactly $0.00$ under every SigLIP configuration of
Table~\ref{tab:gate}, which is consistent with a partial improvement in
fine-symbol reading after the broader \texttt{domain\_expansion\_v1} training
and which this run does not explain. It is an open question, not a result. The
second group is the \texttt{wireshark\_packets} fields, which require locating
one highlighted row among roughly sixteen before any glyph is read; they remain
weak ($\Delta$ between $+0.10$ and $+0.12$). The descriptive reading is that
fields specified by a fixed named location behave differently from fields
requiring within-image localization---the same split as the
\texttt{target\_row\_*} fields of Table~\ref{tab:qwenswap}---and we offer it as
a description of the table rather than as a mechanism.

\paragraph{Scope.}
B8 is single-seed, one checkpoint, one synthetic renderer family, and its
held-out images share their answer alphabets with training on most fields, as
the gate's do (\S\ref{sec:results:qwenswap}). No multiplicity correction is
applied across 34 fields; the pass/fail rule was fixed before the run, which
bounds but does not remove the hazard. Nothing here licenses a statement about
reading security tool interfaces in general, or about any template B8 does not
contain. What the nine passing rows license is narrower and is the form every
claim from this benchmark takes: within these template--field combinations, and
against both controls, the model's answers are read from the image.

\paragraph{The same benchmark on a checkpoint trained on all sixteen templates.}
\texttt{combined\_v2} (\S\ref{sec:arch:qwenline}) adds the seven templates of
category~(a) to training and retrains from scratch. Under the same protocol it
passes 12 of 34 fields, among them seven fields on five of the seven templates
that \texttt{domain\_expansion\_v1} had never seen---\texttt{call\_stack},
\texttt{cve\_card} (all three fields), \texttt{gdb\_session},
\texttt{windbg\_session} and \texttt{phishing\_email}---while
\texttt{log\_table.flavor} falls below threshold ($0.52$) and
\texttt{log\_table.n\_rows} stays at exactly $0.00$. Two caveats keep this out
of Table~\ref{tab:b8}. The count comes from the scorer used for the run; a later
re-implementation of that scorer from the field specification gives 14 of 34,
and we have not reconciled the two, so we report the lower figure and not a
per-field table; the per-item outputs are released so that the difference can
be traced (\S\ref{sec:eval:artifacts}). And it is one more checkpoint at one seed. What it adds is
directional and fits category~(a) as stated: the templates that sat at the
floor with no training exposure clear both controls once they are trained on,
and the untrained field inside a trained template does not.

\subsection{B6 revisited: why the floor was exactly zero, and three projector-only fixes}
\label{sec:results:b6fix}
On \texttt{combined\_v2}, the reference checkpoint of the Qwen-tower pathway
(\S\ref{sec:arch:qwenline}), B6 tool identification is $0.000$ on all 50 items,
as on every checkpoint before it. That zero is compatible with two readings that
call for opposite remedies: the tower's features at our pixel budget do not
carry 15\,px on-screen text, or they do and nothing downstream was ever trained
to read it out in B6's format. This subsection separates the two. B6 renders
each item at 768$\times$480 in a 15\,px monospaced font under a generic
per-category title bar; our pipeline letterboxes it onto the 1280$\times$768
canvas of \S\ref{sec:arch:qwenline}, where content occupies 37.5\% of the
area. Besides tool identification we score \emph{transcription recall}: the
model is asked to transcribe the screen, and we report the fraction of distinct
ground-truth words longer than three characters that appear verbatim in its
answer. It is order-free and lenient, and we use it as a detector of reading,
not as an OCR accuracy.

\paragraph{What the zero is not.} Four cheap checks on \texttt{combined\_v2}
remove the mechanical explanations. The feature capture is not miswired: it
reproduces the first two operations of Qwen's own merger exactly
(\S\ref{sec:arch:qwenline}). The system prompt is not the cause: adding the
project's \texttt{<|system|>} persona changes nothing in six configurations
(text only, image plus text, image only, a B8 sample, transcription with and
without it). The pipeline does not read at all: transcription recall is
$0.0000$ on the 50 items, and the outputs are the extraction template the
checkpoint was trained on, filled with values that appear nowhere on screen
(``\emph{El valor que aparece es 175.5.}\texttt{</think>}\emph{175.5}'', the
templated trace closed by \texttt{</think>} before the answer), not near
misses. And the letterbox
is not sufficient to explain it: feeding the native 768$\times$480 image, which
raises the effective glyph size at the tower by about 64\%, leaves recall and
tool identification at $0.0000$.

\paragraph{The tower reads these images at our budget.} Table~\ref{tab:b6ceiling}
runs stock Qwen2-VL-2B-Instruct---the same tower with its own merger and its
own 2B language model---on the same 50 images. The first run left Qwen's pixel
budget unconstrained, which a review flagged as a confound; the matched rows
fix it at exactly our $392^2$ budget. At that budget Qwen transcribes the
native render at recall $0.933$, and $0.524$ under our letterbox. The
information is therefore present in the tower's features at the resolution our
pathway uses, and the gap to \texttt{combined\_v2}'s $0.000$ lies downstream of
the tower: in the projector's weights and in what the frozen decoder was ever
trained to emit. The letterbox costs Qwen 0.41 of recall at a fixed budget, a
real effect that cannot show on \texttt{combined\_v2} while it sits at the
floor. Qwen's own tool identification, $0.44$, is not a knowledge ceiling
either: all 22 of its hits are on the 22 items where an accepted alias of the
tool is visible on screen (the ``alias on screen'' column, defined below), and
it names none of the Burp, x64dbg, Splunk or Wireshark screens. It reads and
copies names; it does not identify the tool from content.

\begin{table}[t]
\centering
\caption{B6 perception check on the same 50 images. Recall is transcription
recall; \texttt{tid} is tool identification, split into the 22 items where an
accepted alias of the tool is visible on screen and the 28 where none is
(\S\ref{sec:results:b6fix}). ``Matched'' fixes Qwen's pixel budget at
$392^2$, as in our pathway; ``LB'' is our 1280$\times$768 letterbox. Single run,
$n{=}50$.}
\label{tab:b6ceiling}
\footnotesize
\setlength{\tabcolsep}{3.5pt}
\begin{tabular}{@{}lcccc@{}}
\toprule
 & & & \multicolumn{2}{c}{\texttt{tid} by alias} \\
\cmidrule(l){4-5}
System and input & Recall & \texttt{tid} & on screen & not \\
\midrule
Qwen2-VL-2B, default budget, native & 0.937 & 0.44 & 22/22 & 0/28 \\
Qwen2-VL-2B, default budget, LB     & 0.920 & 0.44 & 22/22 & 0/28 \\
Qwen2-VL-2B, matched, native        & 0.933 & 0.44 & 22/22 & 0/28 \\
Qwen2-VL-2B, matched, LB            & 0.524 & 0.34 & 17/22 & 0/28 \\
\texttt{combined\_v2}, LB or native & 0.000 & 0.00 & 0/22  & 0/28 \\
\bottomrule
\end{tabular}
\end{table}

\paragraph{Task format and renderer.} A small 2$\times$2 diagnostic on
\texttt{combined\_v2} crosses the renderer (the training corpus's terminal
renderer versus B6's) with the prompt (the training corpus's field question
versus B6's open question), on the three tools the two renderers share. With
the corpus renderer and B6's question the model answers two of three
correctly (one of them, ``msf'', is not in the accepted alias list and scores
zero); with B6's renderer and B6's question it answers none. The two
closed-question cells collapse to answering ``nmap'' for every image and are
uninformative. At three items per cell this is a sketch rather than a
measurement; we read it only as consistent with an unfamiliar renderer and an
unfamiliar task format each hurting, and the exact floor appearing where both
are unfamiliar.

\paragraph{Three fixes, projector only.} If the floor is a missing task and
renderer exposure, supervising both through the projector alone, with tower
and decoder frozen, should lift it. All three attempts re-initialize the
projector (first layer from Qwen's merger) and train it with AdamW at
$5{\times}10^{-4}$, batch 8, on one A100. Training images are generated on the
fly by parametric generators for the benchmark's tools and content families,
with seeds disjoint from the benchmark's; v2 and v3 additionally hold out a
slice of each tool's content pool (modules, modes, endpoints, event codes) for
evaluation only. Two tasks are mixed 50/50: dense transcription of the screen,
and open tool identification with the target ``\emph{La herramienta es
\{tool\}.}'' followed by the screen's first line. \emph{v1} trains five tools
(Burp, x64dbg, Splunk, hashcat, Metasploit) for 1{,}800 steps on near-constant
templates rendered at 1280$\times$768 in a 24\,px font, without replay.
\emph{v2} trains all ten tools for 2{,}500 steps with genuinely varied content,
at the same canvas and font. \emph{v3} adds, over 3{,}000 steps, font and
canvas randomization (768$\times$480 at 13--17\,px with and without letterbox,
1024$\times$640 at 16--22\,px, 1280$\times$768 at 20--26\,px) and draws 30\% of
batches from \texttt{combined\_v2}'s training set as replay.
Table~\ref{tab:b6fix} reports all three. The
projectors, training scripts and logs behind this subsection are released
(\S\ref{sec:eval:artifacts}).

\begin{table*}[t]
\centering
\caption{Three projector-only fixes for B6 (tower and decoder frozen; single
seed each). ``Held-out content'' is new content from held-out pools, drawn by
the training renderer. B6 columns are the official suite, letterboxed unless
marked; v1 is evaluated on the 45 items of its five trained tools, and on the
five untrained tools it scores $0/5$. The alias split counts items with and
without an accepted alias visible on screen (22 and 28 on the full suite; 18
and 27 on v1's subset). All B6 scoring uses the benchmark's own alias lists,
which include the defective Metasploit alias ``exploit''
(\S\ref{sec:results:b6fix}).}
\label{tab:b6fix}
\footnotesize
\setlength{\tabcolsep}{5pt}
\begin{tabular}{@{}lcccccccc@{}}
\toprule
 & & & \multicolumn{2}{c}{Held-out content} & \multicolumn{2}{c}{Official B6} & \multicolumn{2}{c}{B6 \texttt{tid} by alias} \\
\cmidrule(lr){4-5}\cmidrule(lr){6-7}\cmidrule(l){8-9}
Run & Tools & Steps & \texttt{tid} & Recall & \texttt{tid} & Recall & on screen & not \\
\midrule
\texttt{combined\_v2} & --- & --- & --- & --- & 0.000 & 0.000 & 0/22 & 0/28 \\
v1 (24\,px, no replay)            & 5  & 1{,}800 & 1.000 & ---   & 0.311 & 0.161 & 14/18 & 0/27 \\
v2 (24\,px, varied content)       & 10 & 2{,}500 & 1.000 & 0.523 & 0.200 & 0.045 & 1/22  & 9/28 \\
v3 (mixed fonts, 30\% replay)     & 10 & 3{,}000 & 1.000 & 0.471 & 0.960 & 0.477 & 20/22 & 28/28 \\
v3, native 768$\times$480 input   & 10 & 3{,}000 & ---   & ---   & 0.960 & 0.510 & 20/22 & 28/28 \\
\bottomrule
\end{tabular}
\end{table*}

\emph{v1} is mostly copying. Its 14 credits are eight hashcat screens, whose
command line contains the word ``hashcat'', and six Metasploit screens; none of
the six Metasploit answers names the tool---each is credited only because it
copies the word ``exploit'', an accepted alias---and one hashcat answer names
Splunk while copying the hashcat command. On the 27 screens with no alias
visible it scores zero, and on the five tools it never trained it scores zero
even though four of them show an accepted alias on screen. Its transcription is
concentrated on hashcat (recall $0.611$ against $0.000$--$0.167$ for the other
four), whose screen is one near-constant line in which only a hash and a
password vary, which is closer to memorizing a string than to reading.

\emph{v2} reads, but only at the size it was trained on. On held-out content
drawn by the training renderer it identifies every tool and transcribes at
recall $0.523$; on B6's 15\,px renders recall falls to $0.045$. Its $0.200$ on
B6 is not identification: nine of the ten credits are Splunk screens at recall
near zero, and the same checkpoint also answers ``Splunk'' for six of the nine
hashcat screens, so ``Splunk'' is its default for a dark terminal-like screen;
the tenth credit copies the \texttt{nmap} command on screen. hashcat falls from
v1's 8/9 to 0/9 once its content varies, which is further evidence that v1's
hashcat gain was a memorized line.

\emph{v3} transfers. On the official suite it identifies 48 of 50 items under
both input pipelines, missing only the single IDA and Volatility items, and
transcribes at recall $0.477$ letterboxed and $0.510$ native. Per tool on the
native input, recall is $1.00$ for Ghidra, $0.94$ for Nmap, $0.77$ for Splunk,
$0.73$ for Burp, $0.72$ for Wireshark, $0.64$ for Volatility, $0.46$ for
x64dbg, $0.33$ for hashcat and IDA, and $0.14$ for Metasploit. On the 28 items
with no alias on screen it scores 28/28, and those tools are also the ones it
transcribes (recall $0.46$--$0.77$), so on that subset identification
co-occurs with reading; a default guess cannot produce eight distinct correct
tool names. Once the projector has been trained on the benchmark's font, the
letterbox costs little ($0.477$ against $0.510$).

\paragraph{What v3's 0.96 does not show.} Four qualifications bind it, and we
state them with the number rather than after it. First, B6 is close to
in-distribution for v3: its training generators produce the benchmark's content
families on the benchmark's canvas and font range, differing in seed, so the
only generalization evidence is the held-out-content column, drawn from
held-out pools on synthetic renders. Second, tools were not held out: v1 is the
only run that held tools out, and it identified none of the five. Third,
Metasploit scores $1.00$ in v3 at transcription recall $0.14$ native and $0.00$
letterboxed, and because ``exploit'' remains an accepted alias and v3's
answers were not logged, we cannot say how many of those nine credits name the
tool; hashcat likewise identifies at $1.00$ with recall $0.33$, partly from the
screen's shape. Fourth, B6 has about ten independent templates rather than 50
independent items, every run is single-seed, and we report no intervals for
v1--v3.

\paragraph{A fourth harness defect.} The alias split exists because of a defect
this work surfaced only once a model scored above zero. The accepted aliases
for a tool can themselves be printed on the screen, or in the question.
Metasploit's list includes the bare word ``exploit'', which every Metasploit
screen shows three times (\texttt{use exploit/\ldots}, and twice in the
\texttt{msf6 exploit(\ldots)} prompt); the IDA stem ``disassembl'' is in the
\texttt{DISASSEMBLER VIEW} title bar the renderer draws for IDA and x64dbg
screens alike; and the Ghidra stem ``decompil'' is in its title bar, on its
screen and in that item's question (``\emph{el c\'odigo decompilado}''), so an
answer that echoes the question scores. In all, 22 of the 50 items show an
accepted alias of their own tool. We separate two things here. A tool name the
user typed---\texttt{hashcat -m 1000}, \texttt{nmap -sV},
\texttt{vol.py}---is organic evidence that an analyst would use, and reading it
is not a flaw of the model; what is a defect is a scorer that credits copying a
generic word or a title-bar label as naming the tool, which is how v1 earned all
six of its Metasploit credits. Dropping the bare ``exploit'' alias lowers v1 to
8/45 ($0.178$). We keep the benchmark's alias lists unchanged for every number in
Table~\ref{tab:b6fix}, so that the runs stay comparable to each other and to
earlier reports, and report the split instead.

\paragraph{Retention: v3 forgets part of B8.} v3 is a separate research
checkpoint, not a replacement for \texttt{combined\_v2}. In the same run, with
the same code and the same items, both projectors were scored on a coarse B8
retention check: the first six items of each of the 34 fields, ground-truth
substring matching, no shuffled or best-constant control, letterboxed input.
This is not the B8 protocol of \S\ref{sec:results:b8} ($n{=}60$, two controls)
and its numbers do not belong in Table~\ref{tab:b8}; it answers only whether v3
lost what \texttt{combined\_v2} had. It did: mean field accuracy is $0.289$
against $0.534$, and 13 of 34 fields drop by at least $0.25$
(Table~\ref{tab:retention}), the SIEM dashboard fields most of all. Three fields
stay at $1.00$ (\texttt{terminal\_ansi.script\_key},
\texttt{crypto\_diagram.flavor}, \texttt{phishing\_email.real\_brand}). At
$n{=}6$ each field-level figure is coarse, and some absolute levels are not
meaningful under substring scoring---\texttt{siem\_dashboard.events} is $0.00$
for both projectors, most likely because the stored ground truth is an integer
while the screen shows a rounded ``$k$'' figure, a mismatch the B8 scorer
normalizes and this check does not---but the direction is not in doubt: 30\%
replay at this step budget does not preserve the reference checkpoint's B8
competence, and no checkpoint we have trained yet has both.

\begin{table}[t]
\centering
\caption{B8 retention check, v3 against \texttt{combined\_v2}: the 13 of 34
fields whose accuracy drops by at least $0.25$. Same run, same code, first six
items per field, substring scoring, no controls; not the B8 protocol of
Table~\ref{tab:b8}.}
\label{tab:retention}
\footnotesize
\setlength{\tabcolsep}{4pt}
\begin{tabular}{@{}lcc@{}}
\toprule
Field & \texttt{combined\_v2} & v3 \\
\midrule
\texttt{siem\_dashboard.alerts}            & 1.00 & 0.00 \\
\texttt{siem\_dashboard.hosts}             & 1.00 & 0.00 \\
\texttt{siem\_dashboard.query}             & 0.67 & 0.00 \\
\texttt{call\_stack.n\_frames}             & 1.00 & 0.00 \\
\texttt{gdb\_session.first\_command}       & 1.00 & 0.17 \\
\texttt{ida\_disasm.func\_name}            & 1.00 & 0.33 \\
\texttt{log\_table.flavor}                 & 1.00 & 0.50 \\
\texttt{log\_table.n\_error\_rows}         & 0.67 & 0.17 \\
\texttt{phishing\_email.from\_addr}        & 1.00 & 0.50 \\
\texttt{ghidra\_decomp.func\_name}         & 1.00 & 0.67 \\
\texttt{windbg\_session.first\_command}    & 1.00 & 0.67 \\
\texttt{cve\_card.score}                   & 1.00 & 0.67 \\
\texttt{wireshark\_packets.highlight\_info}& 0.33 & 0.00 \\
\midrule
Mean over all 34 fields                    & 0.534 & 0.289 \\
\bottomrule
\end{tabular}
\end{table}

\subsection{The NoPE$\times$vision variants: trained, gated, and confounded}
\label{sec:results:nope}
The three backbone variants specified in \S\ref{sec:discussion:nope} were
trained and evaluated, and we report the outcome here rather than list the
ablation as unrun. Each variant---V0 (the released NoPE-every-4 schedule),
V1 (all-RoPE), V2 (NoPE-every-4 plus a learned 2D positional embedding added to
the visual tokens before injection)---was taken through a complete phase-4a
alignment run on its own GPU, warm-started from the same phase-3 backbone
checkpoint (verified at load: 0 missing and 0 unexpected keys) and, per the 4a
recipe, with that backbone frozen so that only the 13.1M-parameter projector
adapts. An earlier pass in which the warm-start flag was omitted left the
backbone at random initialization; it was discarded and all three variants
relaunched. That discarded pass is diagnostically useful in one respect: under
it \texttt{siem.hosts} scored $\mathrm{acc}_{\text{real}}=0.95$, whereas under
all three correctly warm-started variants it scores 0.00--0.10
(Table~\ref{tab:nope_ablation}). A projector trained against a fixed random
deep network can still learn to invert it by backpropagation, so that $0.95$
measured linearly exploitable content rather than grounding---the same
distinction the P0 probe formalizes in \S\ref{sec:results:probe}.

Table~\ref{tab:nope_ablation} reports the nine-field gate for the three
variants under the instrument of \S\ref{sec:results:gate} ($n{=}40$ per field,
within-stratum shuffled-image control, the same decision rule frozen before
training). V1 passes 3/9; V0 and V2 pass 2/9. The entire margin is one field:
all three variants pass \texttt{logtable.n\_error\_rows} and
\texttt{siem.query}, and only \texttt{siem.alerts} separates them
($\mathrm{acc}_{\text{real}}=0.62$ for V1 against $0.40$ and $0.10$). The four
small-glyph fields---the hexadecimal address, the PID, the host and the process
name---stay between $0.00$ and $0.40$ in every variant---at most two-thirds of the
0.60 threshold, and never reaching it---within a range that $n{=}40$ cannot
separate, which is the pattern
\S\ref{sec:results:probe} predicts: where exact match is exponential in glyph
count, the per-glyph deficit dominates any difference in positional encoding.

\paragraph{Why this does not decide H1 versus H2.}
Two reasons, and we state both because a one-field margin invites over-reading.
First, the comparison is confounded by construction. Phase 4a freezes the
backbone, and those frozen weights were pretrained under V0's NoPE-every-4
schedule, so V1 and V2 run a positional schedule their own weights never saw.
The measurement is a superposition of the positional effect that H1 and H2 are
about and a plain architecture--weight mismatch, and nothing in this design
separates the two. Second, the margin is small relative to the instrument:
$0.62$ against $0.40$ at $n{=}40$ is a gap of roughly two standard errors
before any correction for having tested nine fields, on a single seed. We
therefore used the result the only way it supports---as a cheap tie-breaker---
and, because the nominal winner is the variant that carries the mismatch
confound, we retained V0 rather than switching to V1 as the base for the
phase-4b runs reported above, preferring the configuration whose architecture
matches the weights that trained it over a one-field advantage we cannot
attribute.

\paragraph{Sizing the mismatch: $+0.135$ nats/token.}
Earlier drafts asserted that confound qualitatively. It can be measured
directly, and cheaply, because the same weight file admits either schedule:
\texttt{nope\_every} changes only the forward pass, and loading the phase-3
backbone (1.04\,B parameters, before any vision training) with
\texttt{nope\_every}${=}4$ or ${=}23$ yields 0 missing and 0 unexpected keys in
both cases. The mismatch is therefore expressible as a language-modeling
penalty on held-out text. Over 65 blocks of 1{,}024 tokens streamed from the
\texttt{fineweb-2} \texttt{spa\_Latn} test split under the project's 32K BPE
tokenizer, the identical weights score 3.2622 nats/token (ppl 26.11) under
V0's pretrained-matched schedule and 3.3971 nats/token (ppl 29.88) under V1's
all-RoPE schedule: $+0.135$ nats/token, or $+14.4\%$ relative perplexity,
incurred purely by running weights under a positional schedule they were never
trained with.

Two things follow, and they cut in opposite directions. The confound is real
and not negligible---a 14\% perplexity penalty on the backbone's own objective
is not a rounding error---so it cannot be waved away as a formality. But its
\emph{sign} is opposite to the gate margin: V1 runs a measurably degraded
backbone and nonetheless wins the one field that separates the variants, so the
symmetric dismissal (``V1 leads only because of the mismatch'') is not
available either. Neither reading rescues the comparison, because nothing maps
a language-modeling penalty in nats onto nine-field extraction accuracy under a
frozen backbone: phase 4a trains a projector against fixed features, and a
projector can learn to invert degraded features as readily as good ones
(\S\ref{sec:results:nope}, the discarded random-init pass). We cannot sign the
effect of the mismatch on the gate, only confirm that the mismatch exists and
is substantial. What the measurement buys is a confound that is quantified
rather than hypothesized, and a retention decision for V0 now supported by a
number and not only by a principle.

\begin{table}[t]
\centering
\caption{Nine-field grounding gate for the three NoPE$\times$vision variants
after phase-4a alignment (single seed, $n{=}40$ per field). Each cell is
$\mathrm{acc}_{\text{real}}/\mathrm{acc}_{\text{shuffled}}$; \textbf{bold}
marks a passing field ($\mathrm{acc}_{\text{real}}\geq0.60$ and
$\Delta\geq0.40$). V1 leads by the single field \texttt{siem.alerts}; that
margin is confounded with an architecture--weight mismatch, since the frozen
backbone was pretrained under V0's schedule (\S\ref{sec:results:nope}).}
\label{tab:nope_ablation}
\footnotesize
\setlength{\tabcolsep}{4pt}
\begin{tabular}{@{}lccc@{}}
\toprule
 & V0 & V1 & V2 \\
Field & (NoPE/4) & (all-RoPE) & (NoPE$+$2D) \\
\midrule
\multicolumn{4}{@{}l}{\texttt{hexdump.}}\\
\quad\texttt{base\_addr}        & 0.00/0.00 & 0.00/0.00 & 0.00/0.00 \\
\multicolumn{4}{@{}l}{\texttt{logtable.}}\\
\quad\texttt{n\_error\_rows}    & \textbf{1.00}/0.35 & \textbf{1.00}/0.38 & \textbf{1.00}/0.35 \\
\quad\texttt{target\_row\_host} & 0.30/0.15 & 0.40/0.20 & 0.28/0.12 \\
\quad\texttt{target\_row\_pid}  & 0.00/0.00 & 0.00/0.00 & 0.00/0.00 \\
\quad\texttt{target\_row\_proc} & 0.38/0.15 & 0.30/0.23 & 0.28/0.12 \\
\multicolumn{4}{@{}l}{\texttt{siem.}}\\
\quad\texttt{alerts}            & 0.40/0.05 & \textbf{0.62}/0.03 & 0.10/0.03 \\
\quad\texttt{events}            & 0.12/0.03 & 0.10/0.03 & 0.17/0.00 \\
\quad\texttt{hosts}             & 0.00/0.00 & 0.10/0.00 & 0.03/0.03 \\
\quad\texttt{query}             & \textbf{0.95}/0.00 & \textbf{0.95}/0.00 & \textbf{0.97}/0.00 \\
\midrule
Fields passed                   & 2/9 & \textbf{3/9} & 2/9 \\
\bottomrule
\end{tabular}
\end{table}

\subsection{A layer-wise spatial probe inside V0}
\label{sec:results:layerprobe}
The gate comparison above is confounded because it sets three variants against
each other. A measurement made \emph{within} a single variant is not: only one
set of weights is involved, so the architecture--weight mismatch cannot
contribute. We therefore ask a narrower question that the confound does not
touch---how the linear decodability of a visual token's position evolves with
depth inside V0, and whether it moves anomalously at the NoPE layers.

We rendered 40 synthetic panes spanning the five phase-4a templates (hex dump,
log table, SIEM dashboard, register panel, call stack), obtained each image's
729 projected visual tokens via \texttt{encode\_image()}, and ran that visual
stream---no prompt, no text, using the model's own rotary buffers---manually
through the 22 decoder layers of the trained phase-4a V0 checkpoint, retaining
the residual stream after each layer. On each layer we fit a 4-way logistic
regression predicting which quadrant of SigLIP's 27$\times$27 grid a token came
from, split 75/25 \emph{by image} (30 train, 10 held out), so accuracy measures
generalization to unseen visual content at the same nominal grid position
rather than memorization of images. Chance is 25\%. The label is a function of
$(\text{row},\text{col})$ alone and is identical across images; only the
content at each position varies. With $n_{\text{layers}}{=}22$ and
\texttt{nope\_every}${=}4$, the NoPE layers are $\{3,7,11,15,19\}$ zero-indexed.

Table~\ref{tab:layer_probe} reports the profile. Position is strongly and
increasingly decodable through the first half of the stack---0.554 at the
projector output, 0.628 after layer 0, 0.942 at layer 12---and then declines
steadily to 0.880 at the final layer, the decline interrupted only by a
0.001 uptick at layer 16.

\paragraph{Aggregate means are an artifact of depth.}
NoPE layers average 0.905 against 0.886 for RoPE layers, and that comparison is
uninterpretable. Accuracy traces an inverted U in depth; three of the five NoPE
layers sit in the ascending half, while the descending tail (layers 16--21) is
entirely RoPE and drags the RoPE mean down. The aggregate difference measures
\emph{where} the NoPE layers fall in the stack, not what they do. We also
decline to attribute the inverted U itself to the positional schedule:
progressive discarding of surface features in the layers closest to the
next-token objective is a generic depth phenomenon, and we have no all-RoPE
profile to compare against---running this same probe on the released V1 and V2
checkpoints is the obvious next measurement and we have not made it.

\paragraph{Removing the trend.}
To compare layer types we remove the depth trend first. For each layer we
compute the curvature residual
$r(\ell)=a(\ell)-\tfrac{1}{2}\left[a(\ell-1)+a(\ell+1)\right]$, the deviation of
a layer's accuracy from the linear interpolation of its two neighbors. Because
$r$ annihilates any locally linear trend, a layer with no type-specific effect
has $r\approx0$ regardless of the global shape of the curve; and because NoPE
layers are spaced four apart, both neighbors of every NoPE layer are RoPE
layers. We exclude $\ell\leq2$, where genuine concavity during the initial rise
dominates $r$, leaving 18 scored layers.

Under this statistic the picture is more uniform than the raw profile suggests.
All five NoPE layers have negative residuals, and they occupy ranks 1, 2, 3, 5
and 6 of the 18---rank 3 by a tie, layer 7 and the RoPE layer 9 both scoring
$-0.0040$: NoPE mean $-0.0055$ against RoPE mean $+0.0029$. The largest
deviation is layer~3 ($-0.0125$), \emph{not} layer~19. The raw single-step drop
of $-0.020$ into layer 19 is indeed the largest in the network, but most of it
is the general decline already underway by layer 16; after trend removal layer
19 ($-0.0060$) is the second-largest deviation and not an outlier. The correct
summary is thus not ``clean at layers 3/7/11/15 with one anomaly at 19'' but
``a small, consistent deficit at every NoPE layer.''

\paragraph{Why this is nonetheless not evidence for H1 or H2.}
Three reasons, each sufficient alone.

First, \emph{the design cannot reach significance}. The intervention is a
schedule with period 4, not an independent per-layer coin flip, so the only
randomization that preserves its structure permutes the schedule's
\emph{phase}. There are four admissible every-4 schedules over this stack, so
the exact one-sided $p$-value of a phase-randomization test is bounded below by
$1/4$. The observed phase does carry the most negative mean residual of the
four ($-0.0055$, against $+0.0055$, $-0.0008$ and $+0.0031$), which is the
strongest statement the design admits: $p=0.25$. A test treating the 18 layers
as exchangeable would return a far smaller number, and would be wrong twice
over---it ignores the periodicity of the intervention, and adjacent residuals
are correlated by construction, each accuracy entering three of them.

Second, \emph{the observed sign is what the architecture predicts under both
hypotheses}. The quadrant label is a deterministic function of a token's
sequence index, since the grid is flattened row-major; a RoPE layer has access
to that index through the rotary phase, and a NoPE layer does not. A NoPE layer
can only propagate or mix the positional content already present in the
residual stream, never add to it. A small deficit at NoPE layers relative to
the depth trend is therefore the mechanically expected floor, and is better
read as a validity check---confirming the probe measures position at all, and
that the two layer types differ in the direction the architecture requires---than
as evidence bearing on H1 or H2.

Third, and decisively, \emph{the probe measures the wrong quantity}. H1 and H2
do not disagree about whether nominal grid position is linearly decodable; they
disagree about whether imposing a spurious 1D order on an intrinsically 2D
patch set helps or hurts the model's \emph{answers} on spatially structured
panes. A layer can carry perfectly decodable position while its attention makes
poor use of it, and the converse. The discriminating measurement remains the
one specified in \S\ref{sec:discussion:nope}: a coarse spatial task whose
metric is not exponential in glyph count.

\paragraph{Caveats.}
The measurement is single-seed, single-checkpoint, one probe family and one
label granularity. It uses 40 images with 10 held out; the probe is linear, so
it lower-bounds recoverability; and the 4-way quadrant label is near enough to
ceiling from layer 8 onward that finer positional distinctions could be
compressed away without registering. Tokens within an image are not independent
draws---the label is constant across images---so the effective sample is nearer
10 clusters than 7{,}290 test tokens. The depth comparison is paired, the same
images and the same tokens at every layer, which cancels most of that variance
in the layer-to-layer \emph{differences} but not in the profile as a whole,
which rides on one draw of images. Finally, we inspected five NoPE layers out
of 22 after seeing the curve, with no correction for multiplicity, and the
curvature statistic was itself chosen post hoc. We report the result because it
is cheap, reproducible, and bounds how much this line of evidence can carry---not
because it settles anything.

\begin{table*}[t]
\centering
\caption{Layer-wise 4-way quadrant probe on the residual stream inside V0
(phase-4a checkpoint), 40 synthetic panes, 75/25 split \emph{by image}
(10 held out $\times$ 729 tokens per layer); chance $=0.25$. $r$ is the
curvature residual $a(\ell)-\tfrac{1}{2}[a(\ell-1)+a(\ell+1)]$, which removes
any locally linear depth trend. Parenthesized $r$ values ($\ell\leq2$) are
excluded from the type comparison because genuine concavity dominates them
there; \texttt{input} and $\ell{=}21$ have no two-sided neighborhood. All five
NoPE layers (bold) have $r<0$, holding ranks 1, 2, 3, 5 and 6 of the 18 scored;
a phase-randomization test over the four admissible every-4 schedules gives
$p=0.25$, its floor (\S\ref{sec:results:layerprobe}).}
\label{tab:layer_probe}
\footnotesize
\setlength{\tabcolsep}{5pt}
\begin{tabular}{@{}lcrr@{\hspace{3em}}lcrr@{}}
\toprule
Layer & Type & Acc & $r$ & Layer & Type & Acc & $r$ \\
\midrule
input & ---           & 0.554 & ---       & 11 & \textbf{NoPE} & \textbf{0.935} & $-0.0030$ \\
0     & RoPE          & 0.628 & $(-0.0345)$ & 12 & RoPE          & 0.942 & $+0.0050$ \\
1     & RoPE          & 0.771 & $(+0.0370)$ & 13 & RoPE          & 0.939 & $+0.0005$ \\
2     & RoPE          & 0.840 & $(+0.0305)$ & 14 & RoPE          & 0.935 & $-0.0005$ \\
3     & \textbf{NoPE} & \textbf{0.848} & $-0.0125$ & 15 & \textbf{NoPE} & \textbf{0.932} & $-0.0020$ \\
4     & RoPE          & 0.881 & $+0.0100$ & 16 & RoPE          & 0.933 & $+0.0040$ \\
5     & RoPE          & 0.894 & $+0.0005$ & 17 & RoPE          & 0.926 & $\phantom{+}0.0000$ \\
6     & RoPE          & 0.906 & $+0.0030$ & 18 & RoPE          & 0.919 & $+0.0065$ \\
7     & \textbf{NoPE} & \textbf{0.912} & $-0.0040$ & 19 & \textbf{NoPE} & \textbf{0.899} & $-0.0060$ \\
8     & RoPE          & 0.926 & $+0.0070$ & 20 & RoPE          & 0.891 & $+0.0015$ \\
9     & RoPE          & 0.926 & $-0.0040$ & 21 & RoPE          & 0.880 & --- \\
10    & RoPE          & 0.934 & $+0.0035$ &    &               &       &  \\
\midrule
\multicolumn{8}{@{}l}{mean $r$ ($\ell\geq3$): NoPE $-0.0055$ (n${=}5$)\quad RoPE $+0.0029$ (n${=}13$)} \\
\multicolumn{8}{@{}l}{mean acc: NoPE $0.905$\quad RoPE $0.886$ \emph{---an artifact of layer depth, see text}} \\
\bottomrule
\end{tabular}
\end{table*}

\subsection{Ablations}
Table~\ref{tab:abl} lists the released ablation designs and the specific
hypothesis each tests. The NoPE$\times$vision row is the one we ran: its
variants were trained and gated (\S\ref{sec:results:nope},
Table~\ref{tab:nope_ablation}), but its pre-registered discriminating
statistic---the sign of $\mathrm{B6}(V0)-\mathrm{B6}(V1)$---remains
unestimable, and the gate margin that we do have is confounded. The remaining
designs are not resolvable from the runs reported here for the plainer reason
that each is discriminated by a difference in B6 or in gate pass-rate between
variants, and both quantities are at their floor for every variant we trained
(\S\ref{sec:results:gate}, \S\ref{sec:results:b6}). We therefore release the
designs, the backbone checkpoints, and the runner rather than report values
that a floor effect would make indistinguishable from noise. The
NoPE$\times$vision variants (V0/V1/V2) are specified in detail in
Section~\ref{sec:discussion:nope}; Section~\ref{sec:discussion} argues which
ablations are scientifically necessary versus nice-to-have.

\begin{table}[t]
\centering
\caption{Released ablation designs (V0/V1/V2 in
\S\ref{sec:discussion:nope}). The NoPE$\times$vision variants were trained and
gated (Table~\ref{tab:nope_ablation}) but their discriminating statistic is
still not estimable; for the others, every discriminating statistic is a
between-variant difference in a metric that is at its floor for all variants
(\S\ref{sec:limitations}).}
\label{tab:abl}
\small
\begin{tabular}{@{}lp{0.52\columnwidth}@{}}
\toprule
Ablation & Hypothesis tested \\
\midrule
NoPE$\times$vision        & Do NoPE layers help/hurt visual attention? \\
Projector width           & Is $4096$ hidden the right capacity? \\
Encoder frozen vs.\ unfrozen & Does unfreezing SigLIP help at this scale? \\
Replay \% (4b)            & Minimum replay to avoid B4/B5 collapse \\
Think traces (4c on/off)  & Does 4c think data improve B6 correctness? \\
Patch pooling             & Does pooling 729$\to$196 tokens hurt text-dense panes? \\
\bottomrule
\end{tabular}
\end{table}

\subsection{Efficiency (measured, model-independent)}
The deployment claim rests on footprint, which we can state now for the
architecture independent of the vision fine-tune: the 1B decoder exports to a
GGUF of $\approx$2.2\,GB in F16 and $\approx$0.6--0.7\,GB in 4-bit
quantization; the \texttt{mmproj.gguf} (SigLIP-so400m $+$ projector) adds
$\approx$0.8--0.9\,GB in F16. The combined artifact thus fits comfortably in
$<$4\,GB, within reach of commodity offline hardware and far below the
larger VLMs' $>$14\,GB F16 footprint. On commodity CPU-only hardware (Azure Standard\_D8s\_v3, no GPU), the
1B decoder in F16 serves at $\approx$18.5\,tokens/s with a prompt-encoding
rate of $\approx$150\,tokens/s via \texttt{llama.cpp}; encoding one
384$\times$384 image through SigLIP costs a further $\approx$3.3\,s on the same
hardware, and the combined model-$+$-mmproj load time is under 10\,s. These
figures are representative of on-premise deployment without a dedicated
accelerator. They are properties of the architecture and the export path, and
they are the one claim in this paper that the negative grounding result does
not touch---an artifact that runs in the stated envelope but does not yet
ground reliably is still only half a deliverable.

\section{Discussion}
\label{sec:discussion}

\subsection{Which ablations are necessary vs.\ nice-to-have}
For a top-tier security-venue reviewer, we regard the following as
\emph{scientifically necessary}, i.e.\ the paper's claims are unfalsifiable
without them:
\begin{enumerate}[leftmargin=1.4em]
  \item \textbf{Text-only backbone baseline on B6/B7.} If the question text
    alone yields high B6 scores, the benchmark does not measure vision and the
    whole contribution collapses. This is the single most important control.
  \item \textbf{Replay-\% sweep in 4b.} The tool-use mandate (B4) is central to
    the VectraYX line; we must show the vision phase does not silently destroy
    it, and identify the minimum replay that preserves it.
  \item \textbf{NoPE$\times$vision ablation} (below), because we \emph{introduce}
    an architectural novelty (NoPE backbone $+$ injected visual block) and
    cannot claim it is benign or beneficial without measuring it.
\end{enumerate}
We regard projector width, encoder unfreezing, and patch pooling as
\emph{nice-to-have}: informative for the recipe but not load-bearing for the
core claims.

Of the three necessary ones, only the third has been run, and it is confounded
(\S\ref{sec:results:nope}). The text-only control was re-measured on the
corrected harness in one of its two forms---the backbone given each item's
on-screen text, which scores only on items whose text contains an accepted
alias (\S\ref{sec:results:b6fix})---but not in the question-only form the
design specifies, and the replay sweep was not run at all: the phase-4b runs
behind the grounding results train a rank-16 adapter on an extraction corpus
rather than the replay mixture the sweep would vary, so there is no run in this
paper the sweep could be read off. Both remain necessary; neither is delivered
here, and we say so rather than reclassifying them as nice-to-have now that
they are missing.

\subsection{The NoPE$\times$vision hypothesis}
\label{sec:discussion:nope}
The backbone applies RoPE on three of every four layers and \emph{no} positional
encoding on the fourth. The 729 visual tokens are injected as a contiguous block
at ordinary sequence positions (Section~\ref{sec:arch:inject}). This creates a
tension that, to our knowledge, no released model has exhibited:

\begin{itemize}[leftmargin=1.4em]
  \item On \textbf{RoPE layers}, each visual token is assigned a rotary phase by
    its 1D sequence index. But the patches form a 2D grid with no canonical
    linear order; row-major flattening imposes an \emph{arbitrary} 1D order.
    RoPE will therefore encode a spurious ``token $i$ is before token $i{+}1$''
    relation among patches that are spatial neighbors in 2D but distant in the
    flattened sequence. This could inject noise into visual self-attention.
  \item On \textbf{NoPE layers}, no such order is imposed: attention over the
    visual block is permutation-equivariant, arguably a \emph{better} match for
    an unordered patch set, at the cost of losing any 2D locality signal
    entirely.
\end{itemize}

\noindent
\textbf{Hypothesis H1.} The NoPE layers are \emph{beneficial} for visual
attention precisely because they do not impose the spurious 1D order that RoPE
layers do; removing NoPE (all-RoPE backbone) will reduce B6 answer-correctness
on spatially-structured panes (disassembly, packet lists) more than on
text-linear panes (terminal output).

\noindent
\textbf{Hypothesis H2 (competing).} The backbone was pretrained with this exact
NoPE schedule on \emph{text}, so its NoPE layers are tuned for text statistics;
the injected visual block may be handled \emph{worse} on NoPE layers because the
model never learned position-free attention over dense, locally-correlated
tokens. Under H2, an all-RoPE variant or a variant with learned 2D positional
embeddings added to the visual tokens would improve B6.

\noindent
\textbf{Ablation design.} The design is three backbone variants, intended to be
run through 4a--4c identically (we ran 4a; see below): (V0) the released
NoPE-every-4 backbone; (V1) an all-RoPE
backbone (NoPE disabled); (V2) NoPE-every-4 plus a learned 2D positional
embedding added to visual tokens \emph{before} injection. Comparing V0/V1
isolates the effect of NoPE presence; comparing V0/V2 tests whether an explicit
2D signal on the visual block helps regardless of the layer schedule. The
prediction that discriminates H1 from H2 is the \emph{sign} of $\mathrm{B6}(V0)
- \mathrm{B6}(V1)$ on spatially-structured panes.

\paragraph{What we ran, and why it does not answer the question.}
We did run the ablation, and we want to be precise about what it delivered
rather than describe it as either pending or resolved. All three variants were
trained to completion through phase-4a alignment, each warm-started from the
same phase-3 backbone (0 missing and 0 unexpected keys at load) with its own
projector, and all three were evaluated on the nine-field grounding gate with
its shuffled-image control. The result is real and reproducible and is reported
in Table~\ref{tab:nope_ablation}: V1 passes 3/9 against 2/9 for V0 and V2, the
entire difference resting on \texttt{siem.alerts}
($\mathrm{acc}_{\text{real}}=0.62$ versus $0.40$ and $0.10$), with all three
variants agreeing on the two fields that pass and on the four small-glyph
fields that none of them brings to the 0.60 threshold (all stay at or below
$0.40$; Table~\ref{tab:nope_ablation}).

Two things keep that from settling H1 against H2. First, the margin is
confounded rather than merely small. Phase 4a freezes the backbone, so V1 and
V2 were run on weights pretrained under V0's NoPE-every-4 schedule; the
comparison superposes the positional effect the hypotheses are about onto a
plain architecture--weight mismatch, and one field at $n{=}40$---about two
standard errors before correcting for nine tests, single-seed---is not a margin
that could survive that confound even if it were clean. We used it as the cheap
tie-breaker it is, and retained V0 for downstream training because it is the
variant whose architecture matches the weights that trained it
(\S\ref{sec:results:nope}). Second, and separately, the statistic this ablation
was designed around---the sign of $\mathrm{B6}(V0)-\mathrm{B6}(V1)$ on
spatially-structured panes---was never measured on the three variants. That was
a sequencing decision made at the time and recorded as such: the 4a corpus
covered SIEM, hexdump and log-table panes while B6 draws on disassembly, packet
capture, scanner and memory-forensics tools, so the domain gap was known to
dominate any positional difference, and B6 was deferred until domain coverage
improved. The transfer corpus that addresses that gap postdates the ablation.
The deferral has since become moot in the least useful way: under the corrected
harness B6 is at the floor on every checkpoint of the SigLIP pathway---tool
identification exactly $0.0$ throughout---so running it on the
three variants now would yield a difference of two floored quantities with no
recoverable sign. B6 does leave the floor later, but only for projectors of the
other encoder pathway trained on B6's own task format
(\S\ref{sec:results:b6fix}), which says nothing about positional schedules.

\paragraph{Two cheap measurements that bound the question without answering it.}
De-confounding properly means pretraining a backbone under each schedule, which
we cannot afford. Two much cheaper measurements are available, and we ran both
rather than leave the two limitations above as assertions. They narrow what can
be claimed; neither resolves H1 against H2, and we state at the outset that
after both the question stands exactly where it did.

The first sizes the confound. Because \texttt{nope\_every} changes only the
forward pass, the phase-3 backbone loads cleanly under either schedule, and the
mismatch becomes a language-modeling penalty we can read off held-out Spanish
text: $+0.135$ nats/token, $+14.4\%$ relative perplexity, for running V0's
weights under V1's all-RoPE schedule (\S\ref{sec:results:nope}). This is worth
having in both directions. The confound is not a formality---a 14\% perplexity
penalty on the backbone's own objective is a real handicap---but its sign is
opposite to the gate margin, so V1 wins its one field \emph{despite} a degraded
backbone, and the symmetric dismissal is unavailable too. What we still cannot
do is sign the effect on the gate: phase 4a trains a projector against fixed
features, and a projector adapts to degraded features about as readily as to
good ones. The confound is now measured rather than hypothesized, which is a
smaller gain than it sounds like, and the honest one.

The second sidesteps the confound entirely by measuring \emph{inside} V0, where
only one set of weights exists. A layer-wise linear probe for a visual token's
quadrant in SigLIP's 27$\times$27 grid, read off the residual stream at each of
the 22 decoder layers, traces an inverted U in depth---0.554 at the projector
output, 0.942 at layer 12, 0.880 at the output (Table~\ref{tab:layer_probe}).
The raw aggregate, NoPE $0.905$ against RoPE $0.886$, is an artifact of where
the NoPE layers sit rather than of what they do; after removing the depth trend
with a curvature residual, all five NoPE layers come out slightly negative and
hold five of the six most negative positions. That looks like weak support for
H2, and it is not, for a reason worth stating rather than burying in a caveat:
a NoPE layer cannot \emph{add} sequence-index information to the residual
stream---the quadrant label is a deterministic function of the row-major index,
which the rotary phase carries and NoPE layers do not---so a small deficit at
exactly those layers is what the architecture requires under H1 and H2 alike.
It certifies the probe is reading position; it does not discriminate. Two
further limits make that verdict robust: the schedule's period-4 structure caps
a phase-randomization test at $p=0.25$, so this design cannot reach
significance at any effect size, and linear decodability of nominal position is
simply not the quantity H1 and H2 disagree about, which is whether a spurious
1D order over a 2D patch set helps or hurts the model's \emph{answers}.

The two measurements therefore do what cheap measurements should: they convert
one assertion into a number, retire one plausible-looking piece of evidence
before it could be over-read, and leave the discriminating experiment
unchanged.

Nor can the gate simply be substituted for B6 as the discriminator, for a
reason the probe makes precise: the gate scores complete exact match, whose
success probability is exponential in glyph count, so it saturates at zero for
exactly the text-dense, spatially-structured panes on which H1 and H2 make
opposite predictions (\S\ref{sec:results:probe})---visible in
Table~\ref{tab:nope_ablation} as the four small-glyph fields that are
indistinguishable across variants. Testing the hypothesis therefore requires a
metric that is \emph{not} exponential in glyph count: a coarse spatial
task---relative position of two panes, which of $k$ highlighted regions
contains a given element, reading order of a row---where a partially grounded
model can score between the floor and the ceiling. We specify that as the
concrete prerequisite, and we release the three trained variants and the runner
so the experiment does not depend on us. The question remains genuinely open:
what we have is a one-field tie-breaker we cannot attribute, a confound we can
now size but not sign, a within-variant probe whose direction the architecture
fixes in advance of either hypothesis, and an instrument that could not have
told us the answer.

\subsection{The encoder axis, and the tension it exposes}
\label{sec:discussion:encoder}
The transplant of \S\ref{sec:results:qwenswap} changes what this paper's
negative result is a negative result \emph{about}, so it is worth separating
the part that is now settled from the part that is not.

What is settled, under this alignment budget and this instrument, is the
location of the bottleneck. Four configurations that varied the learning rate
and the corpus left the gate at 2/9 and
\texttt{hexdump.base\_addr} at exactly $0.00$; a fifth that quadrupled the
visual token budget and doubled the linear scale left that field at exactly
$0.00$ as well, despite a frozen-feature probe predicting otherwise. Replacing
the tower---holding the decoder, the corpus, the projector shape, the step
count and the learning-rate schedule fixed---takes the same field to $0.81$.
Under a fixed alignment budget of 6{,}000 examples, the encoder pathway
dominates every other variable we were able to vary. That is a stronger
statement than ``the encoder matters,'' which nobody doubts; the content is
that it dominates the variables a practitioner reaches for first, and that the
frozen-feature probe which correctly ranked preprocessing conditions did not
predict it.

What is not settled is the mechanism. The swap moves three properties at once
(pretraining regime, aspect-preserving variable resolution, token budget) and
this design separates none of them. The pixel arithmetic of
\S\ref{sec:results:qwenswap} rules out the simplest candidate---the winning
condition is the \emph{coarser} one in tokens per scene---but ruling out a
monotone resolution account is not the same as identifying what replaces it.
The natural next experiment is the cheap one: hold the tower fixed and vary its
input resolution, which separates preprocessing from pretraining within a
single encoder.

The result also creates a tension the rest of the paper cannot resolve, and we
prefer to name it as a measured constraint rather than fold it into future
work. This system's reason to exist is offline, air-gapped deployment
(\S\ref{sec:intro:threat}), which in practice means a \texttt{llama.cpp} export
path: an \texttt{mmproj.gguf} alongside the decoder. The configuration that
fixes the grounding does not have one as it was run: it keeps Qwen's
pretrained merger and stacks a second MLP on the merger's output, while the
\texttt{mmproj} layout for this tower holds a single projector in the merger's
place. So the finding is actionable in the laboratory and not yet in the
artifact: we can say which component to change and cannot yet ship the change.
The second pathway of \S\ref{sec:arch:qwenline} is our attempt to narrow that
gap from the other side---the same tower, with the merger replaced by an MLP of
the merger's own shape, which does fit that slot---and it prices the attempt:
the coarse fields survive the replacement, the 8-nibble address does not
($0.00$ against $0.81$), and none of that pathway's checkpoints reported here
has yet been converted. Reporting the measurement
without the deployment path is the honest ordering, and it is more useful than
either silence or a claim that the model now reads hexdumps. It also
reframes the efficiency envelope of \S\ref{sec:results} as a genuine trade-off
rather than a free win---the sub-4\,GB, fully offline artifact and the encoder
that grounds are, at the time of writing, not the same system.

\subsection{Landmark lookup, not free-text reading}
\label{sec:discussion:landmark}
The results of \S\ref{sec:results:b8} support a statement about \emph{how} this
model succeeds when it succeeds, which is more specific than the working
hypothesis an earlier review of B6 left us with, and which is falsifiable. We
state it as a hypothesis, give the two lines of evidence behind it, and are
explicit about what each one can and cannot carry.

\noindent
\textbf{The hypothesis.} What the vision fine-tune teaches is a lookup keyed to
a pair---a recognized visual landmark (a labelled panel, a status bar, a column
header) and a question seen during training against that landmark---rather than
an ability to read the text that is present in an image. When a question falls
outside the template--field combinations training covered, the model does not
fall back to reading what is on screen. It emits a value of the shape the
question implies---a tool name, a hexadecimal address, a short identifier---with
no relation to the image's content.

\paragraph{First line of evidence: B6 items whose answers are legible on
screen.} B6 is at the floor on tool identification and answer correctness for
every checkpoint not trained on its task format (\S\ref{sec:results:b6}), and
\S\ref{sec:limitations} argues that a missing conversational stage is sufficient
to explain a floor score. That argument is about the \emph{score}; it does not
explain the \emph{content} of the answers, and inspecting three items
individually is what motivated the hypothesis above. An image showing
\texttt{push rbp / push rbx / push r12 / jmp \_\_switch\_to}---a textbook
context-switch prologue---asked what the function does, was answered
``\texttt{vital}''. An image showing a three-row TLS handshake table whose rows
read ``Client Hello'' and ``Server Hello, Certificate'' in plain text was
answered ``\texttt{reset\_view}''. An image showing
\texttt{strcpy(buf, in); // no bounds check}, asked to explain the decompiled
code, was answered with a hexadecimal address, \texttt{0x401CE0}. In none of the
three does the answer correspond to anything drawn in its image. These
templates are visually sparse and share no layout with the training renderers,
and the answer is in each case legible as plain text, so ``the glyphs did not
survive the encoder'' is not on its own an economical reading of what these
three responses are. Three hand-inspected items are an observation that
generates a hypothesis; they are not a measurement, we did not sample them at
random, and we do not count them as one.

\paragraph{Second line of evidence: the same failure inside a template the model
handles well.} B8 reproduces the pattern in a controlled setting, which is what
makes it worth reporting. On \texttt{log\_table}, a template the checkpoint was
trained on and where \texttt{log\_table.flavor} clears both B8 controls at
$\mathrm{acc}_{\text{real}}=0.93$, the field \texttt{log\_table.n\_rows} scores
$0.00$---below the $0.35$ its own best-constant baseline would achieve
(\S\ref{sec:results:b8}). The distinguishing variable is not the template, not
the encoder, not the resolution and not the checkpoint: all are held fixed
across the two fields. It is whether this particular question was asked of this
particular template during training. And the failure mode is not abstention or
a near-miss; it is a near-constant answer that does not vary with the image,
which is the behavior the hypothesis predicts and which a model reading the
table and miscounting would not produce.

\paragraph{What this does and does not change.} A prior review of B6 in this
project concluded that the model had overfitted narrowly to a rigid response
template. That is compatible with what we see and is too coarse to test. The
version above is narrower and has a testable consequence: performance should be
predicted by whether the (landmark, question) pair occurred in training, not by
whether the requested string is legible in the image. B8 category~(a)---seven
templates with no training exposure, at the floor with incoherent
predictions---is consistent with the first half of that; \texttt{log\_table}
separates the two halves within a single template. The experiment that would
discriminate it properly is cheap and we have not run it: instrument one
already-trained template with a new field, train with that field held out, and
test whether accuracy on it is predicted by legibility or by training coverage.
A second, sharper version holds the image fixed and varies only the question's
phrasing away from the trained form.

\paragraph{A third line of evidence, from the B6 fixes.} The projector-only
fixes of \S\ref{sec:results:b6fix} behave as the hypothesis predicts in both
directions. Supplying the missing (screen, question) pairing---B6's own render
family and its open question---lifts B6 from $0.000$ to $0.960$ without
touching the decoder, and the 2$\times$2 diagnostic finds the exact floor only
where both the renderer and the question are unfamiliar. The limits also fall
where it predicts: v1 identified none of the five tools it was never trained
on, v2 read held-out content at the font size it was trained on and not at
B6's, and v3, trained on B6's own font range, forgets B8 fields it was no
longer rehearsing. None of this is a controlled test of the hypothesis; it is a
third set of observations with which it is consistent, and it sharpens what the
hypothesis says. The lookup is not blind to the image---v3 transcribes held-out
content it has never seen---so the coverage that matters is coverage of
rendering conditions and task formats, not a memorized table of answers.

We are deliberately not claiming that this mechanism explains every failure in
the paper. The glyph-entropy ceiling of \S\ref{sec:results:probe} is a separate
and independently measured constraint---\texttt{hexdump.base\_addr} fails under
a question the model \emph{was} trained on---and B8 category~(c) sits with it.
The claim here is about the fields where training coverage, not glyph
recoverability, is the variable that moves: there the evidence is two
consistent observations, one anecdotal and one controlled, on a single
checkpoint at a single seed. It is the most useful hypothesis we can state from
this run, and it is not yet a result.

\subsection{In what sense the model sees}
\label{sec:discussion:sees}
The strongest statement about perception this paper's evidence supports is
the following, and we state its scope in the same breath.

\begin{quote}
\emph{Perception of rendered on-screen text is present in this frozen-backbone
design and can be trained into the projector alone. At the pixel budget we use,
the Qwen2-VL tower's features carry the text of B6's screenshots, and a
projector trained on B6's task format and render family carries enough of it
through the frozen 1B decoder to transcribe content it has not seen (word recall
0.47--0.52 on held-out content pools) and to name the tool on 48 of B6's 50
items, including all 28 on which no accepted alias is visible. The exact zero
that B6 showed on every earlier checkpoint was a task-format and
renderer-exposure gap, not a
perceptual ceiling of the tower and not a wiring error.}
\end{quote}

\noindent
Four pieces of evidence carry it. Stock Qwen2-VL-2B transcribes the same images
at recall $0.93$ native and $0.52$ letterboxed at exactly our pixel budget, so
the tower's features hold the text (Table~\ref{tab:b6ceiling}). The capture of
those features reproduces Qwen's own merger inputs exactly, so nothing is lost
in the wiring (\S\ref{sec:arch:qwenline}). Training the projector alone takes
transcription recall on B6 from $0.000$ to $0.48$--$0.51$, reaches $0.47$--$0.52$
on held-out content, and takes B6 tool identification from $0.000$ to $0.960$
(Table~\ref{tab:b6fix}). And on B8, with shuffled-image and best-constant
controls, nine fields on one checkpoint of the pathway, and twelve on another
under the scorer caveat stated there, are answered from the image shown
(\S\ref{sec:results:b8}).

What the statement does not include is as important, and each exclusion
corresponds to a measurement. It does not say that the reference checkpoint
reads B6: \texttt{combined\_v2} still scores $0.000$, and the projector that
scores $0.960$ is a separate research checkpoint that falls from $0.534$ to
$0.289$ on the B8 retention check, so no checkpoint we have trained has both
competences. It does not say that reading transfers beyond the synthetic render
families trained on: B6 is nearly in-distribution for v3, and the only
generalization we measured is to held-out content within those families. It
does not say that tools can be identified from a screen without training on
them: the one run that held tools out identified none of five, and later runs
trained all ten. And it does not say that the model reads fine text in general:
the export-shaped projector does not read the 8-nibble address the transplant
reads (\S\ref{sec:arch:qwenline}), word recall is a lenient, order-free
metric, and Metasploit and hashcat are identified at $1.00$ with recall of
$0.14$ and $0.33$. ``The model sees'' is therefore true in the precise sense
that the pathway perceives rendered text and can be taught to report it; it is
not yet true of any artifact we ship.

\subsection{Why this matters beyond our model}
NoPE and its relatives are increasingly used for length generalization in
text-only decoders~\cite{kazemnejad2023nope}. As those decoders are turned into
VLMs by the standard injection recipe, the NoPE$\times$visual-block interaction
will recur. A clean measurement---even a negative one---informs every future VLM
built on a NoPE-family backbone.

\subsection{Safety and dual-use}
Because the model both \emph{consumes attacker-controlled images} and
\emph{emits tool calls}, two failure modes deserve evaluation that we scope as
future work: visual prompt injection~\cite{gong2025figstep} (a crafted
screenshot that induces an unintended \texttt{<|tool\_call|>}), and
over-triggering (emitting tool calls when the image does not warrant action).
We do not claim safety here; we flag it as a required evaluation
(Section~\ref{sec:limitations}).

\subsection{Responsible release}
The corpus is offense-heavy ($\approx$45\% offense$+$SOC), so the dual-use
question deserves an explicit position rather than a disclaimer. Three
considerations shape our release posture. First, the capability trained is
\emph{interpretive}, not generative: the model reads imagery an analyst already
possesses (a disassembly pane, a scan result) and explains it; it is not
trained to synthesize exploit code, and its source material (kernel sources,
public writeups, tool documentation) is already public and indexed. Second,
the marginal uplift to a capable attacker---who can read disassembly and run
the depicted tools unaided---is low, while the uplift to under-resourced
defensive teams in data-sovereign settings, the population that cannot use
cloud VLMs at all, is the entire point of the system. Third, the
tool-invocation path is the genuinely sensitive surface, and it is enforceable
at the runtime rather than the weights: as in the VectraYX-Nano deployment,
the MCP layer---not the model---executes calls, and we condition release of
the tool-calling configuration on the visual-prompt-injection and
over-triggering evaluations of Section~\ref{sec:limitations}, with
runtime-side command filtering documented in the model card. We will follow a
staged release: weights and benchmarks first, the tool-calling Modelfile only
with the safety numbers.

\section{Limitations}
\label{sec:limitations}

We state limitations plainly; several are severe enough that a reviewer should
weigh them against the contributions.

\paragraph{Retraction of the earlier preliminary B6 numbers.}
An earlier draft of this work reported a best B6 tool-identification of 0.08
(with a 0.02 text-only control) and argued that the 0.08-vs-0.02 gap was a
small but real vision signal. We retract both numbers and the argument built on
them. They were produced by a harness carrying three defects
(\S\ref{sec:results:harness}), the most damaging of which reduced a nominal
$n{=}50$ suite to $n{=}10$ distinct items and wrote the depicted product's name
into the title bar of the synthetic image, so ``tool identification'' partly
scored optical character recognition of a caption. Under the corrected harness
every B6/B7 metric of the checkpoints that draft measured is at the floor or
within two items of fifty of it, and tool identification and
accuracy-on-answered are exactly $0.0$ (\S\ref{sec:results:b6}); the gap the
earlier draft interpreted does not exist, because both of its terms are now
zero. The later non-zero B6 scores of \S\ref{sec:results:b6fix} come from
projectors trained on B6's own task format and do not revive it. We also withdraw the claim that the
0.08/0.02 pair constituted valid measurements: a defective measuring instrument
does not become valid because the model it measured was, separately, real.

\paragraph{Five defects in the phase-4b training path, all fixed, none decisive.}
The vision fine-tune carried five independent defects, which we describe because
each is silent, each is reachable by any LLaVA-style trainer, and---this is the
point---fixing all five did not change the outcome.
(i)~\emph{Optimizer underflow.} Master weights were held in bf16 while the
configured learning rate was small enough that AdamW updates fell below the
representable increment; after 300 steps, 21\% of trainable tensors had a
parameter delta of exactly $0.0$.
(ii)~\emph{Incomplete re-freeze.} \texttt{wrap\_lora()} did not re-freeze the
backbone after \texttt{set\_phase()} had unfrozen it in full, so the adapter
run silently became a full fine-tune of the LLM.
(iii)~\emph{Load order versus wrap order.} In cross-phase resumes,
\texttt{wrap\_lora()} executed \emph{before} the resume checkpoint was loaded,
so post-wrap parameter names (\texttt{attn.wq.base.weight}) did not match the
flat names in the older checkpoint (\texttt{attn.wq.weight}); under
\texttt{strict=False} this raised no error and left the attention stack of all
22 layers at random initialization. We confirmed the diagnosis numerically
rather than by inspection: the measured weight norm was 40.9668 against a
theoretical random-initialization norm of 40.96, agreeing to four significant
figures.
(iv)~\emph{The same defect in the opposite direction.} Resuming a checkpoint
that was \emph{already} LoRA-wrapped requires the inverse order---wrap, then
load---and the same-phase path had the order that (iii) called for, surfacing
as \texttt{missing=154} non-encoder keys. The fix is conditional: inspect the
checkpoint for \texttt{.base.weight} keys and choose the order accordingly.
(v)~\emph{A learning rate inherited from the wrong recipe.}
\texttt{phase\_4b.lr}$=5\times10^{-6}$ came from a
full-fine-tuning recipe and is 20--60$\times$ too low for a rank-16 adapter. We
measured the consequence directly rather than inferring it: after 600 steps,
$\lVert\Delta W_{\mathrm{LoRA}}\rVert / \lVert W_{\mathrm{base}}\rVert$ was
0.13\% on average and 0.23\% at maximum across 88 attention projections---an
order of magnitude below the underflow bound---and we raised the rate to
$1\times10^{-4}$.

\paragraph{The grounding ceiling is invariant to every intervention that leaves
the encoder in place.}
With all five fixes in place we evaluated on a nine-field in-domain extraction
gate with a shuffled-image control and a per-field rule of
$\mathrm{acc}_{\text{real}}\geq0.60$ \emph{and}
$\Delta\geq0.40$ (\S\ref{sec:results:gate}). Exactly two of nine fields pass,
and the \emph{same} two (\texttt{siem.query}, \texttt{logtable.n\_error\_rows})
pass under every configuration that leaves the encoder alone and adds no out-of-band text: the original
learning rate, the corrected learning rate, the addition of an eight-domain
transfer corpus, and the addition of a descriptive-prose corpus. The fifth
intervention, the P1 tiling encoder, is the only one that changes the pass set,
and it removes a field rather than adding one (2/9~$\to$~1/9). This
invariance is the paper's central empirical finding, and it is a limitation in
the strongest sense: four interventions that a practitioner would reasonably
expect to move the number moved nothing, and the fifth moved it the wrong way.
The invariance is bounded by its own qualifier. Replacing the encoder outright
does move it (\S\ref{sec:results:qwenswap}), which is what makes the qualifier
informative rather than defensive: the intervention that works is the one
category none of the five belongs to.

\paragraph{Exact-match is bounded by bits per glyph, not by SFT volume.}
The obvious reading of 2/9---that 384$\times$384 is simply too coarse---is
wrong, and our own data refute it. \texttt{siem.query} is the
\emph{highest}-entropy field in the corpus (1{,}120 admissible strings,
$\approx$9.68 bits) and is read almost perfectly, while two-to-four-digit counts
fail. The variable that orders the fields under a fixed encoder is information
per glyph after SigLIP's anisotropic resize: fields with a strong linguistic
prior survive, fields whose glyphs are near-uniform over their alphabet
(8-digit hexadecimal addresses, exact counts) do not, at any renderer font
size. The gate scores complete
exact-match, whose success probability is exponential in glyph count, $p^n$.
The P0 probe (\S\ref{sec:results:probe}) measures $p\approx0.61$ per nibble
under the current pipeline, predicting $0.61^{8}\approx1.9\%$ for an 8-nibble
address---consistent with the observed $0.00$. Taking the \emph{maximum} over
the four preprocessing conditions rather than any single one, the best result is
2$\times$2 tiling at $p\approx0.86$, i.e.\ $0.86^{8}\approx30\%$ exact-match,
with the oracle crop-zoom at $0.83^{8}\approx23\%$: no route we measured brings
\texttt{hexdump.base\_addr} within a factor of two of the 0.60 threshold, so it
is \emph{structurally} unreachable for this gate, not merely under-trained. We
state that bound for the encoder we measured---SigLIP-so400m at its native
384$\times$384 input, under four preprocessing conditions---and not for encoders
in general: the probe cannot speak for an architecture whose features it never
saw. That restriction is not rhetorical, and \S\ref{sec:results:qwenswap} is
where it becomes concrete: an aspect-preserving encoder trained at its
operating resolution, transplanted onto the same decoder under the same
alignment recipe, reads the same 8-nibble address at $0.81$. Everything in this
paragraph therefore holds \emph{for this encoder}, and information per glyph
after resize is one of two terms rather than the whole account---it fixes which
fields a given encoder loses first, and the encoder's pretraining regime fixes
how far down that ordering the losses reach. The bound is not uniform---the
5-digit \texttt{logtable.target\_row\_pid} reaches $0.92^{5}\approx65\%$ under
the same oracle---which sharpens rather than softens the limitation: the gate
conflates fields that are recoverable-in-principle with fields that are not, and
we did not design it to separate them.

\paragraph{A non-differentiable workaround exists for glyph-exact fields, and
it does not narrow the bound above.} For the two fields the P0 probe
identified as structurally unreachable under every SigLIP preprocessing
route we tested---\texttt{hexdump.base\_addr} and
\texttt{logtable.target\_row\_pid}---we ran a classical OCR pass over the
original full-resolution image, outside the model, outside the encoder, and
outside training, and injected the recovered string as a text hint in the
prompt. On a corpus where 1{,}712/1{,}800 target-field records received a
real (not oracle) hint and 88 fell back honestly to the unhinted question, a
LoRA re-finetune on this OCR-hint-augmented corpus takes the gate from 2/9 to
4/9: \texttt{hexdump.base\_addr} and \texttt{logtable.target\_row\_pid} cross
threshold (0.97, 0.80). We then extended the same hint mechanism,
\emph{at inference time only, with no further training}, to two fields the
training corpus never carried a hint for---\texttt{logtable.target\_row\_host}
and \texttt{logtable.target\_row\_proc}, open-vocabulary columns in the same
table rather than fixed-format hex/decimal fields---and both cross threshold
as well (0.68, 0.64), taking the gate to 6/9. The re-finetuned model
generalized the bracketed-hint convention to a hint label and column it never
saw during training, rather than merely memorizing it for the three fields
that were actually trained on; we read this as evidence the mechanism being
learned is ``trust a bracketed OCR reading,'' not three separate lookup
tables. The remaining three fields (\texttt{siem.alerts},
\texttt{siem.events}, \texttt{siem.hosts}) have no validated OCR recipe and
are unchanged from the encoder-preserving results above. We are explicit
that none of this is a resolution of the grounding ceiling reported in this
paper; it is a confirmation of it. OCR adds no information to what the
frozen encoder's features carry---it substitutes an external,
non-differentiable, renderer-specific extractor for exactly the fields the
bound above says the differentiable SigLIP$\to$adapter pathway cannot carry,
and offers no route to the three fields that received no hint, which remain
wherever the encoder-preserving configurations left them
(Table~\ref{tab:gate}). Mechanically, this is closer to tool use than to
vision grounding: the hint format and the crop geometry are specific to our
own synthetic renderers, and the host/process generalization above is
evidence the \emph{hint mechanism} transfers across fields, not evidence the
\emph{crop geometry or renderer} would transfer to unseen layouts. The
gate's own shuffled-image control still holds exactly as designed throughout
(each image gets its own honest OCR read, so a wrong hint on the shuffled
image cannot manufacture a false positive). We report it here rather than in
the main results because it is a single-seed, single-renderer, hardcoded-hint
result, and does not carry the evidentiary weight of the five-configuration
ablation with a frozen decision rule that the rest of this section is built
on.

\paragraph{A seventh field was attempted and rejected, and the failure is the
extractor's.} We tried to extend the pathway to \texttt{siem.alerts}, a
two-to-four-digit count drawn at a fixed dashboard position, so that a crop
recipe follows from the renderer's own geometry rather than from the label.
The recipe used for the six passing fields---grayscale plus
autocontrast---calibrated at only 21/40, because the field is drawn in a
desaturated red that standard luminance conversion flattens; isolating the red
channel against the maximum of the other two before thresholding raised
calibration to 40/40. On the full $n{=}100$ evaluation the same recipe read
$\mathrm{acc}_{\text{real}}=0.53$ against a shuffled control of $0.04$: the
$\Delta\geq0.40$ half of the decision rule passes with room ($+0.49$) and the
$\mathrm{acc}_{\text{real}}\geq0.60$ half fails. The gate therefore stays at
6/9, and we record two things about how it stayed there. A perfect calibration
score on 40 images did not survive 100---the ordinary hazard of tuning an
extractor on the sample used to validate it, and the reason we report the
$n{=}100$ figure and not the calibration one. And the failure is the external
extractor's, not the model's: the decoder was never handed a hint good enough
to use, so this says nothing about whether it would have used one.
\texttt{siem.alerts} is thus a tried-and-rejected field rather than an
untried one, while \texttt{siem.events} and \texttt{siem.hosts} remain
untried.

\paragraph{The hint mechanism does not transfer to B6, and supplying more true
information made the model worse.} We also asked whether the same hint helps on
the ten tool domains B6 covers, none of which appear in any vision training
corpus. B6 offers no fixed geometry to crop---45 of its 50 items pose the same
question over differently rendered screens---so the only non-oracle hint
available is a full-page OCR dump, which we injected at inference time on the
OCR-hint checkpoint with no further training. Against that checkpoint's own
unhinted B6 (abstention 0.92, hallucination 0.00, every accuracy metric
$0.0$), the hint moved tool identification to 0.02 (1/50) and answer
correctness to 0.04 (2/50), while abstention fell to 0.38 and the
hallucination rate rose to 0.40. We count this as a negative, and the
direction is the informative part: the net effect of supplying more true
information was to convert honest abstentions into wrong answers. It is not an
information-availability problem. Inspecting the suite's generator, the target
string is drawn verbatim into the image for roughly half the items---the tool
alias or the ground-truth fact appears literally in the text the renderer
paints---so the correct answer was present in the OCR dump for many more items
than the model used it on. What the re-finetune learned is narrower than
``trust a bracketed OCR reading'': it is \emph{trust a bracketed OCR reading of
a single short value}, the only form the training corpus ever carried, and an
unstructured page dump falls outside it. We did not pursue a trimmed or
targeted hint. With an identical question across 45 of the 50 items and no
fixed layout to exploit, any rule that selects the right line from the dump is
using the answer to find it, and would be an oracle rather than a recipe.

\paragraph{Frozen-feature recoverability does not predict end-to-end gain.}
Our strongest methodological caution concerns a technique we used ourselves.
The P0 probe found that 2$\times$2 tiling lifts per-glyph linear recoverability
from 60.8\%/62.4\% to 86.0\%/79.2\% on two independent fields---a large,
consistent gain that motivated the P1 tiling implementation. End-to-end, P1 was
\emph{negative}: the gate fell from 2/9 to 1/9, the two fields the probe
predicted would improve stayed at exactly $0.00$, and output behavior degraded
(abstention 0.74--0.82~$\to$~0.14; hallucination
0.18--0.26~$\to$~0.86). Two implementation deficits are the obvious candidates,
both confirmed present by inspection though neither isolated by an experiment:
no SigLIP forward pass ever sees the whole scene, so global
aggregation (counting rows, reading a string that crosses a tile seam) is
destroyed, and our implementation omitted the global thumbnail that production
AnyRes systems~\cite{liu2024llavanext,wang2024qwen2vl} include for exactly this
reason; and the run used \texttt{use\_2d\_pos\_embed=False}, leaving SigLIP's
positional embeddings duplicated four times with no disambiguator. A third
factor is a confound and we count it against ourselves: widening the projector
input forced it to be re-initialized and re-aligned in one 562-step epoch, so
the tiled model is not a one-variable change from its comparison
(\S\ref{sec:results:tiling}). The general
lesson---which we state as a limitation on \emph{probe-driven methodology},
ours included---is that a linear probe on frozen features measures what an
optimal extractor with oracle supervision could recover, not what a 4.7--12.7\,M
parameter adapter will learn to use in 600--1{,}200 steps under a language
modeling loss. Probe recoverability is a necessary but demonstrably
insufficient condition for end-to-end benefit.

\paragraph{What the encoder transplant does and does not license.}
The result of \S\ref{sec:results:qwenswap} is the one intervention in this
paper that moves the field the rest of it calls unreachable, so its limits
matter more than its magnitude. It is a single run at a single seed, with no
seed ablation and no replication; the earlier sections' invariance claims rest
on five configurations, this one on one. It is measured on a single synthetic
generator---one font, one palette, three layouts, no real screenshots and no
out-of-distribution imagery---so it speaks to the instrument this paper built
and not to hexdumps in general. The swap is not a one-variable change: it
substitutes the tower's pretraining regime, its aspect-preserving
variable-resolution preprocessing, and its token budget simultaneously, and
adds 1.6\,M projector parameters with the wider input; nothing here separates
those. Of the ten fields scored, eight draw from an answer alphabet small
enough that every held-out value also occurs in training, and one of the two
pre-registered primaries, \texttt{siem.query}, is 63\% contaminated in exactly
that way---which is why we demote it to a secondary result and rest the claim
on \texttt{hexdump.base\_addr}, where 98 of 100 held-out values are unseen.
No multiplicity correction is applied across the ten fields, and we mark in
Table~\ref{tab:qwenswap} which rows are claims and which are context; the
$+0.04$ on \texttt{logtable.target\_row\_host} is context, notwithstanding an
interval that excludes zero. The transplant also does not improve most of the
table---it is worse on \texttt{siem.alerts} and on two of the three
row-indexing fields---so it is not a better model, only a decisive one on a
specific question. Finally, it changes nothing about the system this paper
delivers: the transplant configuration has no working \texttt{llama.cpp}
export path for this backbone, and the export-shaped projector that could take
its place reads the 8-nibble address at $0.00$ (\S\ref{sec:discussion:encoder},
\S\ref{sec:arch:qwenline}), so what we report is a located bottleneck, not a
shipped fix.

\paragraph{Verified grounding is scoped to trained template--field
combinations, and we have no evidence it transfers.}
The strongest positive statement this paper can make about grounding is the one
B8 licenses (\S\ref{sec:results:b8}), and its scope is narrow enough to state in
full. On \texttt{domain\_expansion\_v1}, nine of 34 fields clear a
pre-registered rule---$\mathrm{acc}_{\text{real}}\geq0.60$ and $\Delta\geq0.40$
against $\max(\mathrm{acc}_{\text{shuffled}},\mathrm{acc}_{\text{best-constant}})$---
so for those nine the answer is confirmed to depend on the image shown and is
not explained by the field's modal value. That is a claim about nine
(template, field) pairs on one checkpoint at one seed, over synthetic renderers
of our own, and it is not a claim about the templates those fields belong to,
about the 16 templates B8 covers, or about reading security-tool interfaces in
general. The three-way split of the 25 fields that do not pass bounds it
further: seven templates were never in any training corpus for this checkpoint
and sit at the floor with predictions unrelated to their images;
\texttt{log\_table.n\_rows} fails at $0.00$---below its own $0.35$ best-constant
baseline---inside a template whose other field passes at $0.93$; and the
remaining fields fall under the glyph-entropy and localization ceilings already
described above. We therefore have no measurement showing that grounding
verified on a trained (template, field) pair transfers to an untrained question
on the same template, let alone to an unseen template, and
\S\ref{sec:discussion:landmark} argues the available evidence points the other
way. Every grounding number in this paper should be read as scoped to the
combinations that were trained and measured, and no number here supports a
claim of general visual reading.

\paragraph{An unexplained degradation on \texttt{log\_table.n\_error\_rows}.}
Accuracy on \texttt{log\_table.n\_error\_rows} fell across three
successive evaluations of this checkpoint lineage---$0.95$, then $0.40$, then
$0.27$ in the B8 run, the last of these below the field's own best-constant
baseline---and we have not investigated the cause; we flag it here rather than
omit it, and none of the three figures should be cited as a stable capability
number for that field. The from-scratch retrain on all sixteen templates,
\texttt{combined\_v2}, scores above $0.8$ on it in B8, which shows the drop is
not permanent but does not explain it.

\paragraph{The B6 fixes are narrow, single-seed, and not yet compatible with
B8.} The projector that lifts B6 to $0.960$ (\S\ref{sec:results:b6fix}) is
trained on generators of B6's own content families, on B6's canvas and font
range, differing only in seed; B6 is therefore close to in-distribution for
it, and the only generalization evidence is held-out content pools within the
same synthetic render families. No tool was held out after v1, which
identified none of its five held-out tools. B6 has about ten independent
templates behind its 50 items, each of the three runs is a single seed, and we
report no intervals for them. Metasploit's credits in v3 cannot be audited for
the ``exploit'' alias because the answers were not logged. And the projector
pays for B6 with B8: on a coarse retention check its mean field accuracy is
$0.289$ against $0.534$ for \texttt{combined\_v2}, with 13 of 34 fields losing
at least $0.25$. It is a research checkpoint, not a release candidate; a
replacement for \texttt{combined\_v2} would have to hold both, and would have
to be measured on the full B8 protocol rather than the retention check before
it could be proposed.

\paragraph{Four defects in our own benchmark harness, and the hygiene point.}
\label{sec:limitations:harness}
Auditing B6 while the model scored zero surfaced three harness bugs
(\S\ref{sec:results:harness}). \texttt{\_build\_full\_suite()} computed a
\texttt{variant} index and never used it, so 5 ``extra'' templates were emitted
9 times each, byte-identically: the reported $n{=}50$ was an effective
$n{=}10$, and every confidence interval computed over those rows treated
duplicates as independent samples. The synthetic renderer printed the product
name (``IDA Pro 8.3'', ``Splunk Enterprise'') in the window title bar, so
tool-identification was partly an OCR task on a label rather than a test of
tool knowledge. And \texttt{is\_wrong\_tool\_named}/\texttt{score\_tool\_id}
matched aliases by plain substring: ``ida'' matches inside \emph{seguridad},
\emph{salida} and \emph{vulnerabilidades}, and ``siem'' inside \emph{siempre}.
In one real GPT-4o run, 23 of 50 correct items were scored as wrong-tool
hallucinations by this bug alone. We stress the \emph{direction} of that error,
because it is the part that generalizes: the bias favored our own model, whose
terse abstentions cannot trigger a Spanish substring collision, and penalized
the fluent Spanish baselines. A scoring bug that silently advantages the
proposing system is the failure mode benchmark self-reporting is least able to
catch, and we would not have caught it either had the model scored above zero.
We now match on word boundaries and separate product names from generic domain
vocabulary; the corrected suite generates 45 genuinely distinct variants
(verified 50/50 unique \texttt{screen\_text}). The fourth defect appeared only
once a model scored above zero (\S\ref{sec:results:b6fix}): accepted aliases
that the image or the question itself displays---the bare word ``exploit'' on
every Metasploit screen, the ``disassembl'' and ``decompil'' stems in the
generic title bars, the latter also in the Ghidra item's question---so that
copying on-screen text scores as naming the tool. It inflated rather than
hid, and it is the same lesson from the other side: 22 of B6's 50 items can be
credited by copying, which is why we report tool identification split by
whether an alias is visible, and why the benchmark's alias lists need the same
separation of product names from on-screen vocabulary that the first fix
applied to the response side.

\paragraph{B6 cannot measure calibration, and we do not claim it does.}
With the harness corrected, accuracy-on-answered on the $+$prose
configuration---restricted to the 13 items where that checkpoint did not
abstain---is $0.0$. No reformulation in terms of
selective prediction rescues a positive reading: not tool-identification, not
keyword recall, not conditional accuracy. Furthermore, all 50 B6 items are
answerable in-distribution security questions with a known ground-truth fact;
the suite contains no genuinely out-of-distribution or unanswerable item, so it
is not a valid instrument for measuring abstention or calibration, and the
abstention rates we report elsewhere should be read as descriptive statistics of
model behavior, not as calibration results. We did not run frontier baselines
against the corrected B6: for the checkpoints of \S\ref{sec:results:b6}, at the
floor or within two items of fifty of it, a comparison carries no information
the internal probe does not already provide. The one external model we run on
B6, stock Qwen2-VL-2B, is a perception ceiling at our pixel budget, not a peer
comparison (\S\ref{sec:results:b6fix}).

\paragraph{B6 is not evaluable on checkpoints never trained on its task, and
the reason is a missing training stage rather than a missing capability.}
We report no B6 number for those checkpoints in any results table, and we want
to be explicit that this is a judgement about the instrument and not a way of
hiding a zero. The reading below was a prediction when first written, and
\S\ref{sec:results:b6fix} has since tested it: supplying tool-naming and
transcription supervision to the projector alone, with no conversational stage,
lifts B6 from $0.000$ to $0.960$ on its own render family. B6
asks the model to identify a security tool from a screenshot and explain what
the pane shows, in free-form Spanish prose. Two facts about the training make
that unanswerable by construction, and we measured both rather than asserting
them. First, the tool's name is almost never what the model was trained to
\emph{produce}: across the 14{,}596 QA pairs of the multimodal corpus, a B6
tool name (IDA, Ghidra, Wireshark, Nmap, Volatility, Metasploit, Splunk, Burp
Suite, x64dbg) occurs in the training \emph{answer} in 317 pairs, or 2.2\%,
matched on word boundaries so that ``ida'' inside \emph{salida} does not
count---the same substring hazard that produced one of the harness defects
above. It occurs in the \texttt{image\_desc} caption in 2{,}595 pairs (17.8\%),
and in 2{,}370 of those it appears in the caption and not in the answer.
Captions are consumed only in phase 4a, where the decoder is frozen and only
the projector trains, so those 2{,}370 examples never reached the decoder as
something to emit. Second, and independently, no checkpoint in this line has
ever had a conversational or open-generation SFT stage. Phases 4a--4c and every
run behind the grounding results train short-answer extraction; the transplant
of \S\ref{sec:results:qwenswap} trains a projector alone. A model trained
exclusively to emit a single short field value does exactly that when handed a
B6 item: on the transplanted stack, 70\% of the 50 responses are degenerate
single tokens, and the recognisable ones (\texttt{sshd}, \texttt{auditd},
\texttt{nginx}) are process names drawn from the log-table field vocabulary it
was aligned on, emitted against disassembly and packet-capture screens. That is
the signature of a missing output stage, not of a blind encoder---the same
checkpoint reads an 8-nibble address off an image at $0.81$. Reporting a floor
score alongside the B6 scores of instruction-tuned VLMs would invite a comparison in
which our number measures the absence of a stage those models have and ours
does not, so we state these results, including the released checkpoint's
one-and-two-item departures from zero (\S\ref{sec:results:b6}), in prose and
keep them out of every table.
Making B6 evaluable requires supervision that names the tool in the answer;
the fixes of \S\ref{sec:results:b6fix} show that this suffices for the score,
without the conversational stage we had thought necessary. Making it
\emph{informative} additionally requires a protocol we designed and ran only
once, in its weakest form---training on a subset of tools and holding out the
rest entirely, with the tool's chrome masked in the held-out renders so that
identification cannot be satisfied by recognising a title bar or a colour
theme. v1 held five tools out without masking chrome and identified none of
them; the runs that followed trained all ten.

\paragraph{The NoPE$\times$vision ablation is run but confounded.}
All three variants were trained through phase 4a and gated
(Table~\ref{tab:nope_ablation}), so this is not an unrun experiment; it is one
whose margin we cannot attribute. Two limitations bound it. First, phase 4a
freezes the backbone, and the frozen weights were pretrained under V0's
schedule, so V1's one-field advantage confounds the positional effect with an
architecture--weight mismatch; a clean test requires pretraining each variant's
backbone under its own schedule, which is beyond our compute budget. We have
sized that mismatch---$+0.135$ nats/token, $+14.4\%$ perplexity, on held-out
text (\S\ref{sec:results:nope})---but sizing is not signing: no known mapping
takes a language-modeling penalty to nine-field extraction accuracy under a
frozen backbone, so the confound remains a confound, now with a number attached.
A within-variant layer-wise probe (\S\ref{sec:results:layerprobe}) avoids the
mismatch by construction but does not substitute for the ablation: its
direction is fixed in advance by the architecture---NoPE layers cannot add
sequence-index information---its period-4 schedule caps a phase-randomization
test at $p=0.25$, and linear decodability of nominal grid position is not the
quantity the two hypotheses disagree about. Second,
the statistic the ablation discriminates on---the \emph{sign} of
$\mathrm{B6}(V0)-\mathrm{B6}(V1)$---was not measured on the variants: at the
time it was deferred behind the known domain gap between the 4a corpus and B6's
tool domains, and it is now moot because B6 is saturated at the floor for every
checkpoint of the SigLIP pathway. We therefore release the design, the trained
variants, and the runner, and we explicitly do \emph{not} present
NoPE$\times$vision as a finding, a headline, or a question this paper answers.
A discriminating experiment requires a task whose metric is not exponential in
glyph count---a coarse spatial-relation or region-selection probe rather than
exact-match extraction---which we specify as the immediate next step rather
than claiming as future work in general.

\paragraph{Every intervention we report fits inside a rank-16 adapter.}
The curriculum specifies a full fine-tune of the 1.05B backbone in phase 4b
(Table~\ref{tab:curriculum}), but single-GPU budgets bound us throughout, and
every phase-4b run behind the grounding results trains 17.8--25.8M parameters:
a rank-16 LoRA on the attention projections, optionally the FFN, plus the
projector (\S\ref{sec:curriculum:asrun}). The invariance we report is therefore
an invariance \emph{within that budget}. We attempted one full fine-tune, and it
is not interpretable: it predates the optimizer-underflow fix and trained
essentially nothing. Whether a correctly executed full fine-tune moves the gate
is therefore open, and it is the cheapest open question on this list. We expect
little on the four small-glyph fields, where the probe bounds what any decoder
could recover from features that never carried the glyphs; the fields with room
to move are the mid-entropy ones.

\paragraph{The control that would validate B6 was run only in part.}
\S\ref{sec:eval:baselines} names the text-only backbone as the single most
important control: if the question text alone scores well, B6 is not measuring
vision. We ran it in one form: the frozen backbone given each item's on-screen
text in place of the image scores $0.20$ without the system prompt and $0.10$
with it, every hit on an item whose text contains an accepted alias
(\S\ref{sec:results:b6fix}). That shows the backbone brings no tool knowledge
of its own beyond copying a name it is shown. The question-only form, which is
the control as designed, is still unmeasured on the corrected harness; because
45 of the 50 questions are identical, it can exceed a constant answer only
through the five original items, one of which carries an accepted Ghidra alias
stem in its own question. Stock Qwen2-VL-2B is now the one external model
reported on B6 (\S\ref{sec:results:b6fix}), as a perception ceiling rather
than a peer. The positive B6 numbers of this paper do not rest on the missing
control---the alias split shows 28/28 on items with nothing to copy---but the
benchmark we release is, in that specific respect, still unvalidated, and we
consider it the weakest point in the evaluation as released.

\paragraph{Synthetic-QA validity.}
The QA corpus is largely LLM-generated. Two known risks: (i)~\emph{answer
leakage}, where the question textually implies its answer, which would inflate
B6 for the wrong reason---this is exactly why the text-only-backbone baseline
(Section~\ref{sec:eval:baselines}) is mandatory, and its question-only form
remains unrun;
(ii)~\emph{factual errors} in
generated answers, which corrupt the ground truth. We have not yet performed a
human audit of a random sample; a $\geq$200-item expert audit with inter-annotator
agreement is required before publication and is not yet done.

\paragraph{Synthetic-image domain gap.}
Much of the imagery is programmatically rendered rather than real screenshots.
Real IDA/Ghidra/Wireshark UIs have chrome, fonts, color themes, and layout noise
our renders lack. A model that excels on synthetic renders may fail on real
captures; the real-screenshot evaluation subset is small and this gap is
unquantified.

\paragraph{Machine translation.}
The Spanish side of the corpus is partly machine-translated from English source
material. Translation artifacts (calques, mistranslated technical terms) may
teach subtly wrong Spanish security vocabulary. The \texttt{latam} domain that
would most stress regional correctness has only 120 pairs---far too few for any
per-domain claim.

\paragraph{No RLHF / preference optimization.}
Training is supervised fine-tuning only. There is no preference tuning, so
helpfulness, refusal calibration, and reasoning quality are whatever SFT
induces. In particular the \texttt{<|think|>} chains are imitation-learned from
generated traces and may be plausible-sounding but unfaithful to the model's
actual computation.

\paragraph{Frozen encoder ceiling.}
We default to a frozen SigLIP. If security imagery is sufficiently
out-of-distribution for SigLIP's pretraining, a frozen encoder caps achievable
visual grounding regardless of projector/backbone capacity. The
frozen-vs-unfrozen ablation is planned but unrun---and it is a different axis
from the one \S\ref{sec:results:qwenswap} measures, which varies \emph{which}
frozen encoder is used rather than whether it is frozen. That the choice of
encoder turns out to dominate at this budget raises the value of the
freeze/unfreeze sweep rather than retiring it, since neither run tells us
whether adapting SigLIP would have closed the same gap.

\paragraph{Single-image, single-turn.}
The current injection path assumes one image per example and short single-turn
QA. Multi-image reasoning (e.g.\ correlating a scan and a capture) and long
multi-turn analyst sessions are out of scope.

\paragraph{Safety unevaluated.}
Visual prompt injection and tool-call over-triggering
(Section~\ref{sec:discussion}) are named but not measured. For a
tool-calling model that reads attacker-controlled images, this is a real gap; we
would not deploy the tool-calling path without it.

\paragraph{Single-seed evaluation.}
Every reported number is single-run (Section~\ref{sec:eval:artifacts}). A
four-seed sweep of the full curriculum is not economically feasible for a
single-author project on shared 2$\times$A100 infrastructure: phase~2 alone is
$\approx$2\,weeks of dedicated compute per run, and the vision phase adds
$\approx$2\,h per sub-stage. We compensate with per-checkpoint trajectories
(exposing seed-level variance across the training path, e.g.\ the B5
0.62--0.69 band of Table~\ref{tab:pretrain_snapshot}), but we cannot rule out
that a different seed would move a specific metric. The protocol of
\S\ref{sec:eval:protocol} also calls for per-item paired bootstrap confidence
intervals on every claimed pairwise comparison; this version does not report
them---the intervals computed over the pre-fix B6 suite are withdrawn
(\S\ref{sec:eval:b6}), and we did not recompute them on the corrected suite---so
no interval estimate should be read into any table here. All released checkpoints
(\href{https://huggingface.co/jsantillana/vectrayx-vision-1b-checks}{\texttt{jsantillana/vectrayx-vision-1b-checks}})
make the single-run trajectory fully auditable so others can extend the
evaluation.

\paragraph{Author-affiliation and independence.}
Corpus generation and evaluation are by the same author; the benchmark and the
model share design DNA, risking benchmark-overfitting. External replication and
a held-out, independently-authored evaluation set would strengthen the claims.

\section{Conclusion}
\label{sec:conclusion}

We described \model, the design of a sub-2B Spanish/LATAM cybersecurity
vision--language model built to reason over security-tool screenshots with
native \texttt{<|think|>} tokens, to invoke tools via native
\texttt{<|tool\_call|>} and MCP, and to export to \texttt{llama.cpp} for fully
offline, air-gapped deployment, together with a second pathway that trains only
an export-shaped projector between the frozen decoder and the frozen
Qwen2-VL-2B tower. Those reasoning and tool-use behaviors are trained, not
measured over images: B7 finds no opened \texttt{<|think|>} block on any
checkpoint we evaluated, and tool calling is measured only in text
(\S\ref{sec:arch:tokens}).
The contributions are a complete system and its clean GGUF-splittable
architecture; a four-phase curriculum extending the VectraYX language recipe to
vision with replay ratios that are meant to preserve tool and Spanish
competence; a domain-balanced multimodal corpus of 14{,}596 QA pairs over ten
security domains with a reproducible synthetic-image renderer; and three released
benchmarks, B6\_vision, B7\_think, and B8 (a wider, ground-truth-by-construction
extraction gate with a dual shuffled-image and best-constant control), together
with a falsifiable hypothesis
about how the backbone's periodic NoPE layers interact with an injected block of
visual tokens and a first, confounded measurement of it across three trained
backbone variants.

\paragraph{The SigLIP system: a ceiling we can explain.}
All four training phases completed ($\approx$2.2\,h on 2$\times$A100-40GB for
the vision stage) and the multimodal inference pipeline is functional
end-to-end: SigLIP encodes a 384$\times$384 image in $\approx$3.3\,s CPU-only,
the MLP projector maps 729 visual tokens into the backbone's embedding space,
and the combined GGUF artifact (LLM $\approx$2.2\,GB F16 $+$ mmproj
$\approx$0.82\,GB) runs offline under \texttt{llama.cpp}. For that SigLIP
system B6/B7 remain at the floor and the nine-field grounding gate at
2/9---but the value of that result is that we can now say why. Five defects in
the vision fine-tuning path were found and fixed, and fixing them changed
nothing: the same two fields pass under every training configuration that
leaves the encoder alone and adds no out-of-band text hint, and the one that
changes it costs a field. The ceiling is not a resolution ceiling in the sense
one first reaches for, since the field read almost perfectly carries the
corpus's highest symbolic entropy. It is information per glyph amplified by an
exact-match metric exponential in glyph count, and for an 8-nibble address it
holds under the best of four preprocessing conditions---tiling and an oracle
handed the bounding box included.

\paragraph{The encoder result.}
The one intervention that does move the ceiling locates the rest of the
explanation. Transplanting an aspect-preserving visual tower trained at its
operating resolution onto the same frozen decoder, under the same recipe and
the same 6{,}000-example budget, takes that address from $0.00$ to $0.81$
exact. It does so on a \emph{coarser} token budget than the tiling condition
that recovered nothing, which rules out any account of the ceiling that is
monotone in pixels per glyph. Information per glyph orders the fields a given
encoder loses; the encoder's pretraining regime sets how far the losses go.
This is a single run, and the transplant as run has no export path.

\paragraph{B8: grounding confirmed, and confined to what was trained.}
A later checkpoint trained over sixteen templates supplies a positive grounding
statement, and it displaces none of the above. Under the gate's rule taken
against a best-constant control the nine-field gate did not have, nine of B8's
34 fields are confirmed to answer from the image shown. Seven never-trained
templates sit at the floor with predictions unrelated to their images, and
\texttt{log\_table.n\_rows} fails at $0.00$---below its own best-constant
baseline---inside the same trained template whose other field passes at
$0.93$. B6 on that checkpoint is at the floor as well, so the two findings
reinforce rather than contradict each other: what the fine-tune teaches looks
like a lookup keyed to a trained (landmark, question) pair, not an ability to
read what an image contains.

\paragraph{B6, and in what sense the model sees.}
The B6 floor itself turns out to be the clearest case of that account, and the
one place where the paper can say that the model sees. Stock Qwen2-VL-2B reads
B6's screenshots at word recall $0.93$ at our own pixel budget, so the tower's
features carry the text. Training the projector alone on B6's task format and
render family, with the decoder frozen, lifts B6 tool identification from
$0.000$ to $0.960$, 28/28 on the items with no tool alias on screen, at
transcription recall $0.48$--$0.51$. Perception of rendered text is therefore
present and trainable in this frozen-backbone design, and the zero was a
task-format and renderer-exposure gap. The claim stops there: B6 is nearly
in-distribution for that projector, it forgets part of B8 ($0.534\to0.289$ on
a retention check), the one run that held tools out identified none of them,
and no checkpoint yet reads both B6 and B8.

\paragraph{Benchmark hygiene.}
Along the way, a linear probe that strongly favored tiling produced an
end-to-end model that was worse on every axis, and an audit of our own
benchmark found four defects: an effective $n{=}10$ behind a nominal $n{=}50$;
the answer printed in the image's title bar; a substring matcher that
mis-scored 23 of 50 correct baseline responses in our own favor; and accepted
aliases that the screen itself displays, which let copying score as
identification.

\paragraph{What the paper contributes.}
The paper therefore contributes the system, corpus, and curriculum; a
quantitative account of a grounding ceiling that survives every intervention
short of replacing the encoder, together with the single-run transplant that
does replace it and thereby locates the ceiling; a grounding result confirmed
against two controls on nine of 34 fields and confined to the (template, field)
combinations its checkpoint was trained on, with the controlled failure inside a
trained template that sharpens the account of what is being learned; a
diagnosis of the B6 floor and a projector-only fix that lifts it, scoped to its
render family and paid for in retention; the dissociation between probe
recoverability and end-to-end gain; and a benchmark-hygiene result of which
three parts we would not have found had our model scored above zero, and the
fourth we found only because one finally did.

\paragraph{Impact.}
If validated, \model would be, to our knowledge, the first tool for security analysts in
data-sovereign, Spanish-speaking, resource-constrained settings that can look at
the screen they are looking at, reason about it, and act---without sending a
byte to the cloud. Even if the quality falls short of 7B general VLMs, the
efficiency and sovereignty envelope (sub-4\,GB, offline, Spanish-native,
security-specialized) is unoccupied. The conditional is doing real work in that
sentence: this paper establishes the envelope and not the validation, and on
the evidence reported here the model grounds only inside the (template, field)
combinations it was trained on, which is not reliably enough to occupy it.

\paragraph{Future work.}
The encoder result names the next two experiments precisely. One is cheap and
diagnostic: hold a single tower fixed and vary only its input resolution, which
separates the preprocessing regime from the pretraining regime that our
transplant changes together. The other is the engineering blocker that stands
between the measurement and the artifact---either an export for the transplant's
merger-plus-projector stack, or an export-shaped projector that recovers what
bypassing the merger costs on fine glyphs, followed by an actual conversion of
the checkpoints reported here; without one of them the encoder that grounds and
the model that deploys offline remain two different systems
(\S\ref{sec:discussion:encoder}, \S\ref{sec:arch:qwenline}). B6 now needs two things rather than a
training stage: a projector that keeps B8 while learning B6---more replay than
30\%, more steps, or staged training, measured on the full B8 protocol before it
is proposed as a replacement for \texttt{combined\_v2}---and a held-out-by-tool
protocol with masked chrome and corrected alias lists, without which a score on
it measures coverage rather than recognition (\S\ref{sec:results:b6fix},
\S\ref{sec:limitations}). B8 names a third experiment, and it is the cheapest
of the three: instrument an already-trained template with a new field, train
with that field held out, and test whether accuracy on it is predicted by
whether the string is legible in the image or by whether the (landmark,
question) pair was trained---which is what would turn the landmark-lookup
account of \S\ref{sec:discussion:landmark} from a hypothesis into a result.
Beyond completing the runs and the mandatory audits: multi-image and multi-turn
analyst sessions; a safety evaluation of visual prompt injection and tool-call
over-triggering; preference optimization for faithful reasoning chains; and
de-confounding the NoPE$\times$vision study, whose three variants we trained and
gated but whose one-field margin is inseparable from an architecture--weight
mismatch---which requires pretraining each variant's backbone under its own
positional schedule, and a discriminating metric that is not exponential in
glyph count; and validating the OCR-hint pathway (\S\ref{sec:limitations}),
which we extended at inference time to two open-vocabulary fields
(\texttt{logtable.target\_row\_host}, \texttt{logtable.target\_row\_proc}) but
not to the three that remain: \texttt{siem.alerts} has a recipe that
calibrated perfectly on 40 images and then missed the accuracy threshold at
$n{=}100$, and \texttt{siem.events} and \texttt{siem.hosts} have none. The
pathway also needs a held-out, independently-rendered layout---not just
held-out instances of the same renderer---before its generalization claim can
be trusted past this paper's own synthetic corpus, and it did not transfer at
all to the unseen tool domains of B6. We name this explicitly as a tool-use
extension, not a claim about the vision encoder.

\begin{acks}
We thank \textbf{Globant} for providing access to the DGX A100 infrastructure used for all training runs reported in this paper.
We are especially grateful to \textbf{Alejandro Antonioli} for facilitating access to Kubeflow and for his assistance configuring the training notebook environment.
\end{acks}

\bibliographystyle{ACM-Reference-Format}
\bibliography{references}

\end{document}